%% file: main.tex
\documentclass[10pt,twocolumn,letterpaper]{article}

\usepackage{iccv}              % To produce the CAMERA-READY version
\input{preamble}

\definecolor{iccvblue}{rgb}{0.21,0.49,0.74}
\usepackage[pagebackref,breaklinks,colorlinks,allcolors=iccvblue]{hyperref}

\def\paperID{3554} % *** Enter the Paper ID here
\def\confName{ICCV}
\def\confYear{2025}

\title{MOBIUS: Big-to-Mobile Universal Instance Segmentation via \\Multi-modal Bottleneck Fusion and Calibrated Decoder Pruning}

\author{
\hfill Mattia Segu$^{1,2}$, Marta Tintore Gazulla$^1$, Yongqin Xian$^1$, Luc Van Gool$^3$, Federico Tombari$^1$ \hfill \\
\hfill $^1$ Google \; \hfill $^2$ ETH Zurich \; $^3$ INSAIT, Sofia University, St. Kliment Ohridski\hfill\\
\hfill  \hfill 
}

\begin{document}
\maketitle
\input{sec/0_abstract}    
\input{sec/1_intro}

\input{sec/2_related_work}
\input{sec/3_method}

\input{sec/4_experiments}

\input{sec/5_conclusion}

\clearpage
{
    \small
    \bibliographystyle{ieeenat_fullname}
    \bibliography{main}
}

\clearpage
\input{sec/X_suppl}

\end{document}

%% file: preamble.tex
\usepackage{tikz}
\usepackage{pgfplots}
\pgfplotsset{compat=1.17}
\usepgfplotslibrary{groupplots}

\usepackage{caption}

\usepackage{graphicx}
\usepackage{amsmath}
\usepackage{amssymb}
\usepackage{booktabs}
\usepackage{makecell}
\usepackage{multirow}
\usepackage[accsupp]{axessibility}
\usepackage{colortbl}  %彩色表格需要加载的宏包
\usepackage{xcolor}
\usepackage{amssymb} % For checkmarks
\usepackage{adjustbox} 

\DeclareMathAlphabet\mathbfcal{OMS}{cmsy}{b}{n} % Define \mathbfcal
\newcommand{\Eqref}[1]{(Eq.~\ref{#1})}
\NewDocumentCommand{\EqrefRange}{m m}{%
(Eq.~\ref{#1}--\ref{#2})%
}

\newcommand{\methodNAME}{MOBIUS\xspace}

\newcommand{\myparagraph}[1]{\vspace{2pt}\noindent{\bf #1}}

\usepackage{pgfplotstable}
\definecolor{cr1}{RGB}{193,225,193}
\definecolor{cr2}{RGB}{177,156,217}
\definecolor{cr3}{RGB}{250,200,152}

%% file: sec/0_abstract.tex
\begin{abstract}
% Scaling up but not down
Scaling up model size and training data has advanced foundation models for instance-level perception, achieving state-of-the-art in-domain and zero-shot performance across object detection and segmentation. However, their high computational cost limits adoption on resource-constrained platforms.  
% Our proposal
We first examine the limitations of existing architectures in enabling efficient edge deployment without compromising performance. We then introduce MOBIUS, a family of foundation models for universal instance segmentation, designed for Pareto-optimal downscaling to support deployment across devices ranging from high-end accelerators to mobile hardware.
To reduce training and inference demands, we propose:  
(i) a bottleneck pixel decoder for efficient multi-scale and multi-modal fusion,  
(ii) a language-guided uncertainty calibration loss for adaptive decoder pruning, and  
(iii) a streamlined, unified training strategy.  
% Our results
Unlike efficient baselines that trade accuracy for reduced complexity, MOBIUS reduces pixel and transformer decoder FLOPs by up to 55\% and 75\%, respectively, while maintaining state-of-the-art performance in just a third of the training iterations. MOBIUS establishes a new benchmark for efficient segmentation on both high-performance computing platforms and mobile devices.
\end{abstract}

%% file: sec/1_intro.tex
\section{Introduction}
\label{sec:intro}
% The problem
Scaling up model size and training datasets has demonstrated remarkable in-domain accuracy and impressive zero-shot generalization for a variety of domains, including natural language processing (NLP)~\cite{Devlin2019BERTPO, brown2020language, chowdhery2023palm, radford2018improving}, computer vision~\cite{dosovitskiy2021an,kolesnikov2019big,radford2021learning,li2021align}, and reinforcement learning~\cite{silver2017mastering,vinyals2019grandmaster,schulman2017proximal}.
Advances in modern hardware accelerators and growing data availability have fueled the development of foundation models for instance-level perception, addressing tasks ranging from generic object detection and segmentation~\cite{chen2017deeplab,ren2015faster,ronneberger2017u,badrinarayanan2017segnet,detr} to interactive segmentation using visual prompts~\cite{zhang2023segment,lu2023interactive} or referring expressions~\cite{hu2016segmentation,wang2022refcoco}.

Recently, several generalist models~\cite{uniperceiverv2,UNINEXT,liu2024grounding,glee} have built on flexible multi-modal DETR-based architectures~\cite{detr,dino,maskdino} to simultaneously address multiple such tasks.
Their architecture is typically composed of a vision and a text encoder, a pixel decoder that fuses multi-scale vision features with the text modality, and a transformer decoder that refines a set of queries to be used for downstream detection and segmentation by attending to the multi-scale features enhanced by the pixel decoder.
While preliminary generalist models specialized only on a subset of instance-level tasks and domains, GLEE~\cite{glee} scaled up the dataset and model size, employing a multi-stage curriculum learning approach to handle incrementally more difficult tasks while avoiding instability.
%
%%%%%%%%%%%%%%%%%%%%%%
\input{new_tables/teaser}
%%%%%%%%%%%%%%%%%%%%%%
% By enabling more accurate and versatile perception capabilities, such models positively impact numerous real-world applications, including autonomous vehicles, medical diagnostics, and augmented reality,
%
Despite these advancements, the pursuit of ever-larger models has prioritized state-of-the-art performance over efficiency, limiting their adoption on resource-constrained platforms such as autonomous systems, mobile devices, and edge computing.
While scaling up has been widely explored, the challenge of scaling down - reducing model size, training time, and inference complexity while preserving strong in-domain performance and zero-shot generalization - remains unaddressed.
%
% Moreover, the multiple training stages involved result in overly-complex training procedures and extremely long training times.
%

% Our goal
In this paper, we first analyze existing architectures and their performance-efficiency trade-offs towards edge deployment, independently evaluating the pixel decoder, modality fusion, and transformer decoder components (\cref{fig:relative_flops}). 
Then, we introduce \methodNAME (\cref{fig:method}), a family of Big-to-\underline{\textbf{Mobi}}le models for \underline{\textbf{U}}niversal instance \underline{\textbf{S}}egmentation. MOBIUS is designed for Pareto-optimal downscaling, supporting state-of-the-art deployment across devices ranging from high-end accelerators to mobile hardware.
% we aim to provide the first foundation model for instance-level perception built with Pareto-optimal downscaling in mind and a streamlined unified training stage, while retaining strong performance.
% analysis
% \todo{this sentence may be unnecessary:} First, we analyze the architecture of previous generalist models and find that their pixel decoder and transformer decoder take up to 60\% of the total inference latency. We trace this back to the pixel decoder's inefficient multi-scale and multi-modal fusion and to the transformer decoder processing all feature scales. 
%
%
% Refer to \cref{fig:method} for an illustration of the proposed model.
% architectural improvements
To this end, we propose improvements to the model architecture and training strategy to reduce training and inference time while retaining competitive performance:
\begin{itemize}
% bottleneck encoder
\item We introduce a novel pixel decoder - namely the \emph{bottleneck encoder} - which fuses multi-scale and multi-modal information into a single informational bottleneck.
Unlike previous pixel decoders - such as MaskDINO’s transformer encoder~\cite{maskdino} (\cref{tab:mobile_results}, a) and RT-DETR’s hybrid design~\cite{zhao2024detrs} (\cref{tab:mobile_results}, c) - our bottleneck encoder achieves competitive open-vocabulary performance (\cref{tab:mobile_results}, d) while reducing pixel decoder FLOPs by 55\% (\cref{fig:relative_flops}, Pixel Decoder).
By compressing multi-scale and multi-modal features into a single, highly-expressive representational bottleneck, our approach eliminates the need for inefficient multi-scale feature processing in DETR-based transformer decoders~\cite{zhu2020deformable,maskdino}, further reducing decoder FLOPs by 50\% (\cref{fig:relative_flops}, Decoder).

% calibration and decoder pruning
\item We propose a \emph{language-guided uncertainty calibration loss} to calibrate the vision-language object classification scores, which enables our novel \emph{inference-time decoder pruning strategy} to prune irrelevant decoder queries according to their predictive confidence, effectively halving the transformer decoder FLOPs.

% unified / streamlined training stage
\item 	We propose a unified training strategy that stabilizes training across datasets and tasks in a single stage, achieving state-of-the-art performance in just one-third of GLEE’s training iterations.

\end{itemize}

% Experiments
We validate MOBIUS on diverse in- and out-of-domain datasets, demonstrating competitive or superior performance across big and mobile model sizes. Notably, MOBIUS runs in real-time, achieving 10 FPS on mobile devices and 25 FPS on high-end GPUs, making it the most Pareto-efficient universal instance segmentation model (\cref{fig:pareto}).

% Summary of contributions
% To summarize, we introduce a Pareto-optimal family of big-to-mobile models for instance-level perception, owing to:
% (i) a novel optimally-downscalable pixel decoder for scale and modality fusion, 
% (ii) an efficient inference-time decoder pruning strategy via language-guided uncertainty calibration,
% (iii) an streamlined training recipe to unify training stages and improve stability.

%% file: new_tables/teaser.tex
\begin{figure}[t] % or [H] for strict placement
    \centering
\begin{tikzpicture}
    \begin{axis}[
        width=0.95\linewidth,
        height=6cm,
        xlabel={FLOPs (G)},
        ylabel={Performance ($AP_{\text{mask}}$)},
        xmin=0, xmax=750,
        ymin=23, ymax=50,
        % Custom transformation for x axis:
        x coord trafo/.code={
            \pgfmathparse{(#1 < 100) ? 2*#1 : #1 + 100}
            \pgfmathresult
        },
        x coord inv trafo/.code={
            \pgfmathparse{(#1 < 200) ? #1/2 : #1 - 100}
            \pgfmathresult
        },
        xtick={0,50,100,200,300,400,500,600,700},
        xticklabels={0,50,100,200,300,400,500,600,700},
        legend style={at={(0.5, -0.2)}, anchor=north, legend columns=4, font=\tiny, row sep=-3pt, inner xsep=2pt, inner ysep=1pt},
        legend cell align={left},
        xtick distance=100,
        ytick distance=5,
        grid=major,
        major grid style={dashed},
        label style={font=\scriptsize},
        tick label style={font=\scriptsize},
        title=\textbf{Instance Segmentation on LVIS-val},
        title style={font=\small}
    ]
    
    % GLEE family - High resolution
    \addplot[blue, mark=square*, mark options={fill=blue}] coordinates {
        (239, 40.2) % GLEE-Lite highres
        (704, 47.4) % GLEE-Plus highres
    };
    \addlegendentry{$\text{GLEE}_{big}$};
    
    % GLEE family - Low resolution
    \addplot[blue, dashed, mark=square*, mark options={fill=none}] coordinates {
        (30.6, 26.1) % GLEE-Mini-CM lowres
        (119, 33.2) % GLEE-Mini-CM highres
    };
    \addlegendentry{$\text{GLEE}_{mobile}$};
    
    % MOBIUS family - High resolution
    \addplot[red, mark=triangle*, mark options={fill=red}] coordinates {
        (123, 40.3) % MOBIUS-0 highres
        (155, 42.1) % MOBIUS-1 highres
        (206, 43.1) % MOBIUS-2 highres
        (354, 46.7) % MOBIUS-3 highres
    };
    \addlegendentry{$\text{MOBIUS}_{big}$};
    
    % MOBIUS family - Low resolution
    \addplot[red, dashed, mark=triangle*, mark options={fill=none}] coordinates {
        (18, 25.8) % MOBIUS-Mini-CM lowres
        (65.3, 33.8) % MOBIUS-Mini-CM highres
    };
    \addlegendentry{$\text{MOBIUS}_{mobile}$};
    
    % Draw arrow from GLEE-Plus highres (704, 47.4) to MOBIUS-3 highres (354, 46.7)
    \draw[->, thick, green!50!black] (axis cs:690,47.4) -- (axis cs:360,47.4) node[midway, above, font=\scriptsize, green!50!black] {\textbf{-50\%}};
    \draw[->, thick, green!50!black] (axis cs:230,40.2) -- (axis cs:135,40.2) node[midway, above, font=\scriptsize, green!50!black] {\textbf{-50\%}};
    
    \node[anchor=south, font=\scriptsize] at (axis cs:690,47.5) {$\text{GLEE}$};
    \node[anchor=south, font=\scriptsize] at (axis cs:360,47.5) {$\text{MOBIUS}$};
    
    \end{axis}
\end{tikzpicture}
    \caption{\textbf{Pareto efficiency.} The \methodNAME family demonstrates Pareto-efficient downscaling of universal instance segmentation compared to state-of-the-art GLEE. We compare computational requirements (FLOPs) with performance ($AP_{\text{mask}}$) on LVIS-val for big and mobile model sizes. The text encoder fixed cost is omitted.}
    \label{fig:pareto}
\end{figure}
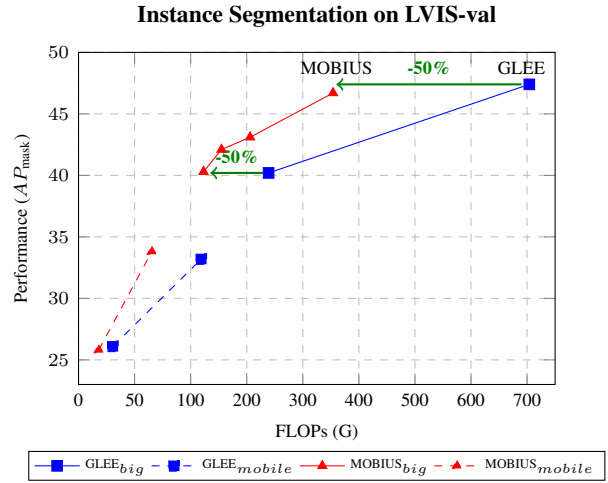

%% file: sec/2_related_work.tex
% \input{tables/flops}
\input{new_tables/component-efficiency-relative}

\section{Related Work}

\paragraph{Generalist Models for Instance Perception.}
Instance-level perception encompasses tasks like generic object detection and segmentation~\cite{chen2017deeplab,ren2015faster,ronneberger2017u,badrinarayanan2017segnet,detr}, segmentation from referring expressions~\cite{hu2016segmentation,wang2022refcoco}, and interactive segmentation from visual prompts~\cite{zhang2023segment,lu2023interactive}.
Generalist models unify these tasks into a single framework.
Early models framed instance perception as a sequence generation task, but suffered from inefficient autoregressive inference~\cite{uniperceiver,wang2022ofa,lu2022unified}. More recent models, like X-Decoder\cite{zou2023generalized} and SEEM~\cite{seem}, process vision, text and prompt modalities through a unified transformer decoder architecture. However, self-attention over many tokens incurs high computational cost, limiting deployment on edge devices. 
Building on DETR-based architectures\cite{detr,dino,maskdino}, Uni-Perceiver v2~\cite{uniperceiverv2}, Unicorn~\cite{unicorn} and UNINEXT~\cite{lin2023uninext} achieve strong in-domain performance but struggle with zero-shot generalization due to closed-set training. In contrast, GLIP\cite{glip1,glip2} and GroundingDINO\cite{liu2024grounding,ren2024grounding,ovdino} redefine multi-modal object detection as a phrase grounding task, and scale up training data to enhance generalization.
GLEE~\cite{glee} extends these models to a broader universal instance segmentation framework - addressing a larger set of instance-level perception tasks - but requires a multi-stage training process to address instability. These methods, however, scale up training data and model size at the expense of efficiency.
In this work, we introduce MOBIUS, the first Pareto-efficient family of generalist models for universal instance segmentation, scaling from high-end GPUs to mobile devices (\cref{fig:pareto}). MOBIUS also eliminates training instability, unifying training stages and achieving similar performance to GLEE in just one-third of the training iterations (\cref{ssec:method_unified_training}).

\paragraph{Efficient End-to-end Object Detectors.}
Following the success of DETR-based architectures~\cite{detr,zhu2020deformable,dino,maskdino}, various works attempt to mitigate DETR’s inefficiencies in the pixel decoder~\cite{yao2021efficient,roh2021sparse,li2023lite,zhao2024detrs} and transformer decoder~\cite{lv2024rt}.
EfficientDETR~\cite{yao2021efficient} reduces decoder layers while compensating with two-stage query selection.
SparseDETR~\cite{roh2021sparse} and FocusDETR~\cite{zheng2023less} sparsify the attention by focusing it on a reduced set of visual tokens.
LiteDETR~\cite{li2023lite} introduces layers of interleaved cross-attention between high- and low-level feature tokens for more efficient cross-scale aggregation.
RT-DETR~\cite{zhao2024detrs} proposes to combine intra-scale attention on high-level features with convolutional top-down and bottom-up cross-scale feature fusion~\cite{Wang_2019_ICCV}.
Due to its efficiency, the RT-DETR pixel decoder has been extended to multi-modal fusion in GroundingDINO 1.5 Edge~\cite{ren2024grounding}.
While RT-DETR improves efficiency in a closed-set vocabulary setting, we find that it struggles with open-vocabulary generalization (\cref{tab:mobile_results}, c), underperforming compared to the MaskDINO-based pixel decoder (\cref{tab:mobile_results}, a).
We propose a novel pixel decoder - the bottleneck encoder (\cref{ssec:method_pixel_decoder}) - that compresses multi-scale and multi-modal information into a single expressive representation. Unlike prior designs, our approach preserves open-vocabulary performance while achieving a 55\% FLOPs reduction over MaskDINO’s pixel decoder (\cref{fig:relative_flops}, Pixel Decoder).
By condensing multi-scale features into a single expressive representation, MOBIUS eliminates redundant multi-scale processing in the transformer decoder, a major inefficiency in DETR-based models. Our single-scale design cuts transformer decoder FLOPs by 50\% (\cref{fig:relative_flops}, Decoder). 
Finally, our language-guided uncertainty calibration loss refines query confidence, enabling adaptive decoder pruning and an additional 50\% FLOPs reduction in the transformer decoder (\cref{fig:pruning}).

%% file: new_tables/component-efficiency-relative.tex
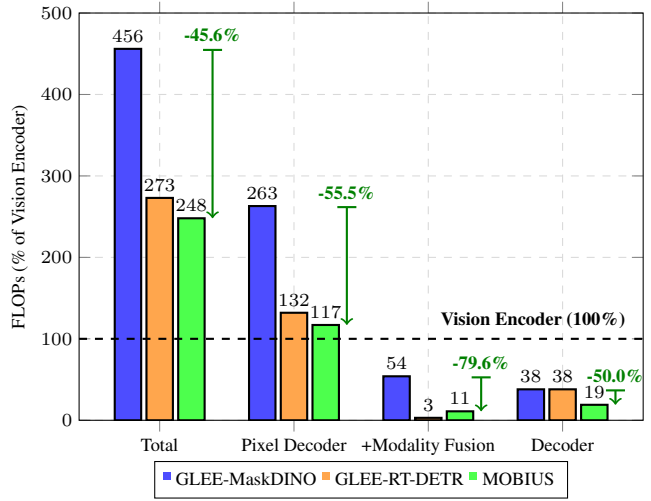
\begin{figure}[!t]
    \centering
    \scriptsize
    \begin{tikzpicture}
        \begin{axis}[
            ybar,
            bar width=10pt, % Keep reduced bar width
            width=0.52\textwidth, % Slightly wider to compensate for spacing
            height=7cm,
            ymin=0, ymax=500,
            enlarge x limits={0.2}, % Reduced from 0.6 to reduce side spacing
            symbolic x coords={Total, Pixel Decoder, +Modality Fusion, Decoder},
            xtick=data,
            grid=major,
            grid style={dashed, gray!30},
            ylabel={FLOPs (\% of Vision Encoder)},
            font=\scriptsize,
            legend style={
                at={(0.5,-0.1)},
                anchor=north,
                legend columns=-1,
                font=\scriptsize,
            },
            % Force perfect square markers in legend:
            legend image post style={scale=0.7}, 
            legend image code/.code={%
                \draw[draw=none, fill=#1] (0cm,0cm) rectangle (0.15cm,0.15cm);
            },
            every axis plot/.append style={thick},
            nodes near coords,
            nodes near coords align={vertical}
        ]
            % GLEE-MaskDINO
            \addplot [fill=blue!70] 
                coordinates {(Total,456) (Pixel Decoder,263) (+Modality Fusion,54) (Decoder,38)};
            % GLEE-RT-DETR
            \addplot [fill=orange!70] 
                coordinates {(Total,273) (Pixel Decoder,132) (+Modality Fusion,3) (Decoder,38)};
            % MOBIUS
            \addplot [fill=green!70] 
                coordinates {(Total,248) (Pixel Decoder,117) (+Modality Fusion,11) (Decoder,19)};

            % FIXED: Dashed horizontal line at 100%
            \draw [black, thick, dashed] (rel axis cs:0,0.2) -- (rel axis cs:1,0.2);

            % FIXED: Text label for the Vision Encoder line at the correct height
            \node[anchor=south east, font=\scriptsize] at (rel axis cs:0.98,0.21) {\textbf{Vision Encoder (100\%)}};

            % Delta arrows and labels (realigned to match green bars)
            \draw[|->, thick, green!50!black] 
                ([xshift=20pt]axis cs:Total,456) node[above, font=\scriptsize, green!50!black] {\textbf{-45.6\%}} -- ([xshift=20pt]axis cs:Total,248);
                
            \draw[|->, thick, green!50!black] 
                ([xshift=20pt]axis cs:Pixel Decoder,263) node[above, font=\scriptsize, green!50!black] {\textbf{-55.5\%}} -- ([xshift=20pt]axis cs:Pixel Decoder,117);
            
            \draw[|->, thick, green!50!black] 
                ([xshift=20pt]axis cs:+Modality Fusion,54) node[above, font=\scriptsize, green!50!black] {\textbf{-79.6\%}} -- ([xshift=20pt]axis cs:+Modality Fusion,11);
            
            \draw[|->, thick, green!50!black] 
                ([xshift=20pt]axis cs:Decoder,38) node[above, font=\scriptsize, green!50!black] {\textbf{-50.0\%}} -- ([xshift=20pt]axis cs:Decoder,19);
            
            % Restore the legend
            \legend{GLEE-MaskDINO, GLEE-RT-DETR, MOBIUS}

        \end{axis}
    \end{tikzpicture}
    % \caption{\textbf{Component-wise FLOPs Comparison.} We compare the computational cost of MOBIUS to GLEE~\cite{glee} with MaskDINO~\cite{li2023mask} and RT-DETR~\cite{zhao2024detrs} pixel decoders. FLOPs are expressed as a percentage of an R50 vision encoder (52.4G), excluding the text encoder. Models are profiled at 800×800 resolution. MOBIUS halves the total cost while keeping competitive performance.}
    \caption{\textbf{Component-wise FLOPs Comparison.} We compare MOBIUS to GLEE~\cite{glee} with MaskDINO~\cite{li2023mask} and RT-DETR~\cite{zhao2024detrs} pixel decoders. FLOPs are given as a percentage of an R50 vision encoder (52.4G), excluding the text encoder. Models are profiled at 800×800 resolution. MOBIUS halves all costs while retaining competitive performance wrt. the GLEE-MaskDINO baseline.}

    \label{fig:relative_flops}
\end{figure}

%% file: sec/3_method.tex
\input{figures/method}

\section{Method}
%
% \todo{in this section we mention the FLOPs improvements multiple times. add a reference to the FLOPs and latency analysis here}
%
We introduce MOBIUS, a Pareto-efficient family of big-to-mobile universal instance segmentation models, designed to scale seamlessly from high-end GPUs to mobile devices while maintaining state-of-the-art performance at a fraction of the computational cost.
First, we outline the overall architecture in \cref{ssec:method_architecture} and \cref{fig:method}.
Then, we propose a novel pixel decoder relying on a representational bottleneck to fuse multi-modal and multi-scale information (\cref{ssec:method_pixel_decoder}).
In \cref{ssec:method_decoder}, we introduce an inference-time query pruning strategy for the transformer decoder, enabled by our novel language-guided uncertainty calibration loss.
Finally, in \cref{ssec:method_unified_training} we describe our technical improvements to streamline the training procedure, enabling stable training in a single-stage across all datasets and tasks.

\subsection{Architecture} \label{ssec:method_architecture}
We aim to provide a foundation model for instance-level perception, capable of solving a variety of tasks ranging from generic object detection and segmentation to grounded segmentation through free-form text or visual prompts. 
% Thus, our architecture (\cref{fig:method}) follows a design similar to many DETR-based generalist models~\cite{} - particularly GLEE~\cite{} - but introduces architectural improvements targeting efficiency. % Our proposed method consists of an image encoder, a text encoder, a visual prompter and an object decoder. 
Our architecture~(\cref{fig:method}) follows established multi-modal DETR-based generalists~\cite{glee,liu2024grounding} and consists of an image encoder, a text encoder, a visual prompter, a pixel decoder and a transformer decoder. Our technical contributions lie in the architectural improvements that substantially reduce the FLOPs of the pixel decoder and transformer decoder.

\myparagraph{Image encoder.} Given an input image, the image encoder extracts a set of multi-scale feature maps $\{\mathbf{S}_2, \mathbf{S}_3, \mathbf{S}_4, \mathbf{S}_5\}$, corresponding to the last four feature scales in the image backbone. Following DINO~\cite{dino}, we further downscale $\mathbf{S}_5$ with stride 2 and obtain $\mathbf{S}_6$.

\myparagraph{Text encoder.} Given a list of categories or free-form text prompts, the text encoder extracts a list of text token embeddings $\mathbf{E}_\text{text}$ which, after category-wise pooling, results in the final text embeddings $\mathbf{z}_\text{text}$.

% \myparagraph{Visual prompter.} Our model can optionally take user inputs in the scenario of interactive segmentation. This is achieved with a visual prompter which is capable of encoding box, scribble or point. 

% \myparagraph{Object decoder.} The feature maps and embeddings obtained above are then fed into an object decoder. Following the GLEE~\cite{wu2024general}, our object decoder consists of a pixel decoder that fuses multi-scale feature maps and text embeddings, and a transformer decoder that predicts the final instance-level bounding box or segmentation mask. As shown in \cref{tab:flops_per_component}, the two decoders are the most computationally expensive components. The key contributions of our work are the proposed pixel decoder and transformer decoder that reduce the complexity significantly.

% The model receives as inputs an image, a set of object categories or a free-form text prompt, and optionally a visual prompt in the form of a box, scribble or point.
%
% The image encoder extracts multi-scale feature maps from the input image, where $\{\mathbf{S}_2, \mathbf{S}_3, \mathbf{S}_4, \mathbf{S}_5\}$ are the last four features scales. Following DINO~\cite{dino}, we further downscale $\mathbf{S}_5$ with stride 2 and obtain $\mathbf{S}_6$.
%
% A text encoder receives a list of categories or free-form text prompts and encodes to a list of text tokens $\mathbf{E}_\text{text}$ which, after category-wise pooling, results in the final text embeddings $\mathbf{z}_\text{text}$.
%
\myparagraph{Pixel decoder.} The feature maps and embeddings obtained above are then fed into a pixel decoder that fuses multi-scale feature maps and text embeddings. Generalist models~\cite{glee,liu2024grounding} typically adopt the DINO~\cite{dino} or MaskDINO~\cite{maskdino} transformer encoder as pixel decoder, consisting of a stack of self-attention layers to fuse multi-scale information, where the input sequence is the concatenation of all multi-scale feature maps. Modality fusion is achieved by bidirectional cross-attention between text tokens and multi-scale feature tokens. These scale and modality fusion operations are extremely expensive due to the long sequence lengths and the quadratic complexity of the self-attention. In contrast, we select only one feature scale $\mathbf{B} = \mathbf{S}_i$ and use it as a representational bottleneck (\cref{ssec:method_pixel_decoder}). Our pixel decoder is then a mixture of deformable self- and cross-attention layers, progressively fusing the multi-scale features $\{\mathbf{S}_3, \mathbf{S}_4, \mathbf{S}_5, \mathbf{S}_6\}$ and the text tokens $\mathbf{E}_\text{text}$ into the single bottleneck $\mathbf{B}$. 
% The resulting enhanced bottleneck $\hat{\mathbf{B}}$ effectively condense multi-scale and multi-modal information in a single scale. 
% \yongqin{add how much we reduce the FLOP here.}
%

\myparagraph{Transformer decoder.} 
% \yongqin{we should emphasize a bit more about our novelty in the transformer decoder and FLOP reduction in this part.} 
The refined feature maps are then fed into a transformer decoder that predicts the final instance-level bounding box or segmentation mask.
Typically, DETR-based transformer decoders suffer from major inefficiencies due to processing multi-scale feature maps.
Our single-scale bottleneck eliminates the need for the inefficient multi-scale processing. 
% resulting in a 50\% FLOPs reduction (\cref{fig:relative_flops}, Decoder).
%
To further improve the efficiency of the transformer decoder, we propose a language-guided query selection strategy. We select from the enhanced bottleneck $\hat{\mathbf{B}}$ the top-K queries $\mathbf{Q}$ by cosine similarity with the text embeddings. 
Such queries $\mathbf{Q}$ are fed to the transformer decoder, where they are refined and optionally pruned (\cref{ssec:method_decoder}) through interactions with the single-scale enhanced bottleneck $\hat{\mathbf{B}}$. 
% Optionally, a visual prompter~\cite{seem} encodes the visual prompts and attends to the enhanced bottleneck representation to propose visual prompting queries for the transformer decoder.
%
The resulting set of refined queries $\hat{\mathbf{Q}}$ is a set of image-specific object representations that can be used for downstream tasks. % is dot-producted with the pixel embedding map to produce instance segmentation masks $\mathbf{I}$.
Following MaskDINO~\cite{maskdino}, we upscale the enhanced bottleneck $\hat{\mathbf{B}}$ and sum it to $\mathbf{S}_2$ to produce an embedding map $\mathbf{M}$, which we dot-product with each refined query to produce the set of instance segmentation masks $\mathbf{I} = \{ \hat{\mathbf{q}} \otimes \mathbf{M} \; \forall \; \hat{\mathbf{q}} \in \hat{\mathbf{Q}}\}$.
%

% \todo{add motivation saying that GLEE's pixel decoder has a lot of FLOPs in the self attn}

\subsection{Efficient Bottleneck Encoder for Multi-scale and Multi-modal Fusion} \label{ssec:method_pixel_decoder}
We design our pixel decoder based on the intuition that, with the proper multi-scale and multi-modal fusion design, a bottleneck representation can optimally condense the fused information and trade off expressivity for model size by varying the bottleneck size.
We propose to select one feature scale $\mathbf{S}_i$  as representational bottleneck $\mathbf{B}$, accompanied by its position embeddings $\mathbf{P}_i$.
Using a specific feature scale instead of a fixed set of learnable embeddings comes with desirable properties: (i) the number of bottleneck tokens $|\mathbf{B}|$ is proportional to the input resolution, (ii) the bottleneck representation inherits the positional embeddings and geometric organization from the corresponding feature map, enabling the use of efficient attention operations such as deformable attention~\cite{zhu2020deformable}. 

\paragraph{Bottleneck Encoder.} A bottleneck encoder block~\Eqref{eq:bottleneck_operations} receives as input the chosen representational bottleneck $\mathbf{B}$, its position embeddings $\mathbf{P}_i$, the multi-scale feature maps $\{\mathbf{S}_3, \mathbf{S}_4, \mathbf{S}_5, \mathbf{S}_6\}$ and the text tokens $\mathbf{E}_\text{text}$.
First, it efficiently fuses the bottleneck representation ${\mathbf{B}}$ with the text tokens $\mathbf{E}_\text{text}$ through bidirectional cross-attention~\EqrefRange{eq:bottleneck_operations_1}{eq:bottleneck_operations_2}~\cite{li2022grounded}, \ie image-to-text cross-attention and text-to-image cross-attention.
Then, we enhance the bottleneck through intra-scale deformable self-attention~\Eqref{eq:bottleneck_operations_3} and multi-scale deformable cross attention~\Eqref{eq:bottleneck_operations_4} with the  multi-scale feature maps $\{\mathbf{S}_3, \mathbf{S}_4, \mathbf{S}_5, \mathbf{S}_6\}$, before feeding it to a feed-forward network (FFN)~\Eqref{eq:bottleneck_operations_5}.
%
% Our bottleneck definition, which preserves the positional embeddings of its original feature scale, enables using deformable attention is competitive with full self-attention while reducing the computational complexity by 20\%. 
Our bottleneck definition preserves the positional embeddings of its original feature scale, enabling the use of deformable attention, which remains competitive with full self-attention while reducing computational complexity by 20\%.
The operations in each bottleneck encoder block $l$ are defined as:
\begingroup
\small
\begin{subequations} \label{eq:bottleneck_operations}
\begin{align}
&\mathbf{B}_{\text{img}\rightarrow \text{text}}^{l} = \text{CA}(\mathbf{B}^{l}, \mathbf{E}_\text{text}) + \mathbf{B}^{l}, \label{eq:bottleneck_operations_1} \\
&\mathbf{B}_{\text{text}\rightarrow \text{img}}^{l} = \text{CA}(\mathbf{E}_\text{text}, \mathbf{B}^{l}) + \mathbf{B}^{l}_{\text{img}\rightarrow \text{text}}, \label{eq:bottleneck_operations_2} \\
&\mathbf{B}_\text{intra}^{l} = \text{DeformSA}(\mathbf{B}_\text{fused}^{l}) + \mathbf{B}_{\text{text}\rightarrow \text{img}}^{l}, \label{eq:bottleneck_operations_3} \\
&\mathbf{B}_\text{multi}^{l} = \text{MSDeformCA}(\mathbf{B}_\text{intra}^{l}, \{\mathbf{S}_3, \mathbf{S}_4, \mathbf{S}_5, \mathbf{S}_6\}) + \mathbf{B}_\text{intra}^{l}, \label{eq:bottleneck_operations_4} \\
&\hat{\mathbf{B}}^{l} = \text{FFN}(\mathbf{B}_\text{multi}^{l}) + \mathbf{B}_\text{multi}^{l}. \label{eq:bottleneck_operations_5}
\end{align}
\end{subequations}
\endgroup
where SA and CA are respectively self and cross attention. GroupNorm is used to normalize the output of each layer. We repeat $M$ such blocks to produce a bottleneck encoder and output the enhanced bottleneck $\hat{\mathbf{B}}$.
The resulting bottleneck encoder efficiently fuses multi-scale and multi-modal information by performing all attention operations at the reduced bottleneck dimensionality. Compared to a MaskDINO-based pixel decoder, our bottleneck encoder reduces the multi-scale fusion cost by 55.5\%, and the modality fusion cost by 79.6\% (\cref{fig:relative_flops}).

\subsection{Efficient Transformer Decoder via Single Scale Decoding and Calibrated Decoder Pruning} \label{ssec:method_decoder}
While the pixel decoder uses more FLOPs, the transformer decoder requires more latency due to being less parallelizable, taking roughly 20\% of the total latency assuming a reference R50~\cite{he2016deep} vision encoder. 
Owing to our bottleneck encoder, our transformer decoder can process the resulting single bottleneck scale with half the FLOPs and without loss of performance wrt. the tradition multi-scale DeformableDETR transformer decoder. 
Nevertheless, we make an additional step to ensure further downscaling under constrained resources.
In particular, we propose to better calibrate the predictive scores of each query during training such that irrelevant queries can be pruned at inference time. 

\paragraph{Single-scale Decoding.} By efficiently condensing multi-scale and multi-modal information into an expressive single-scale representational bottleneck, our model can feed a single scale to the transformer decoder and break free from the multi-scale processing introduced in Deformable-DETR's~\cite{zhu2020deformable} transformer decoder for improved performance. This results in a 50\% FLOPs reduction (\cref{fig:relative_flops}, Decoder) without loss of performance (\cref{tab:bottleneck}).

\paragraph{Language-guided Query Selection.}
Given the enhanced bottleneck $\hat{\mathbf{B}}$ and text embeddings $\mathbf{z}_\text{text}$, we select from the bottleneck $\hat{\mathbf{B}}$ the top-K bottleneck tokens $\mathbf{Q}_K$ ranked by the cosine similarity with the text embeddings and feed them as queries $\mathbf{Q}$ to the transformer decoder:
\begin{subequations} \label{eq:language_guided_selection}
\small
\begin{align}
\sigma^{\text{cls}}_i &= \max_{j} \text{cos}(\mathbf{q}_i, \mathbf{z}_\text{text}^j), \quad \mathbf{q}_i \in \hat{\mathbf{B}}, \mathbf{z}_\text{text}^j \in \mathbf{z}_\text{text}, \label{eq:relevance_score} \\
b_i &= \text{MLP}(\mathbf{q}_i), \label{eq:bounding_box_prediction} \\
% \mathbf{Q_K} &= \text{topK}_{\sigma^{\text{cls}}_i \in [i, M]} (\mathbf{q}_i), \label{eq:top_k_queries} \\
\mathbf{Q}_K &= \{\mathbf{q}_i \mid i \in \text{topK}(\{\sigma^{\text{cls}}_i | \forall \mathbf{q}_i \in \hat{\mathbf{B}}\})\}, \label{eq:top_k_queries} 
\end{align}
\end{subequations}
% where $\sigma^{\text{cls}}_i$ is the confidence score of the feature $\mathbf{q}_i$ based on the cosine similarity {\small $\text{cos}(\mathbf{q}_i, \mathbf{z}_\text{text}^j) = \mathbf{q}_i \cdot \mathbf{z}_\text{text}^j / (\|\mathbf{q}_i\| \|\mathbf{z}_\text{text}^j\|)$} between $\mathbf{q}_i$ and $\mathbf{z}_\text{text}^j$, $b_i$ is the predicted bounding box at each bottleneck feature $\mathbf{q}_i$, and $\mathbf{Q}$ is the set of top-K queries selected based on the confidence scores $\sigma^{\text{cls}}_i$.
% %
% Unlike GLEE, we use a cosine similarity with learnable scaling instead of a simple dot-product to avoid training instabilities~\cref{ssec:method_unified_training}.
where $\sigma^{\text{cls}}_i$ is the confidence score of the feature $\mathbf{q}_i$ based on the scaled cosine similarity {\small$\text{cos}_s(\mathbf{q}_i, \mathbf{z}_\text{text}^j) = exp(s) \cdot \mathbf{q}_i \cdot \mathbf{z}_\text{text}^j/(\|\mathbf{q}_i\| \|\mathbf{z}_\text{text}^j\|)$}.

\begin{equation}
    \text{cos}_s(\mathbf{q}_i, \mathbf{z}_\text{text}^j) = exp(s) \cdot \frac{\mathbf{q}_i \cdot \mathbf{z}_\text{text}^j}{(\|\mathbf{q}_i\| \|\mathbf{z}_\text{text}^j\|)}
\end{equation}

s is a learnable scaling factor. Here, $b_i$ is the predicted bounding box at each bottleneck feature $\mathbf{q}_i$, and $\mathbf{Q}=\mathbf{Q_K}$ is the set of top-K queries selected based on the confidence scores $\sigma^{\text{cls}}_i$.
Unlike GLEE, we replace the simple dot-product with a scaled cosine similarity to avoid training instabilities (\cref{ssec:method_unified_training}).

\paragraph{Language-guided Uncertainty Calibration.}
We propose an uncertainty minimization scheme to improve the calibration of confidence scores for the decoder queries.
We aim to align the predictive distribution $\Sigma$ of the localization error to the one of the classification uncertainty $\mathcal{C}$.
In practice, we define a measure of the localization confidence $\sigma^{\text{loc}}_i=IoU(b_i,y_i)$ as the IoU between a predicted box $b_i$ and its matched ground-truth box $y_i$ and align it to the language-guided classification confidence score $\sigma^{\text{cls}}_{i,j} = \max_{j} \text{cos}(\mathbf{q}_i, \mathbf{z}_\text{text}^j)$ by minimizing a focal loss~\cite{lin2017focal} between the two, where $\sigma^{\text{loc}}_i$ is the target.
{\small
\begin{align}
\mathcal{L}_{cal}(\sigma^{\text{cls}}_{i,j},\sigma^{\text{loc}}_i) = -\alpha_i (\sigma^{\text{loc}}_i - \phi_t(\sigma^{\text{cls}}_{i,j}))^\gamma \log(\phi_t(\sigma^{\text{cls}}_{i,j})), \label{eq:calibration_loss}
\end{align}
}
where $\phi_t(\sigma^{\text{cls}}_{i,j}) = \sigma^{\text{cls}}_{i,j} \cdot \mathbb{I}[j = t] + (1 - \sigma^{\text{cls}}_{i,j}) \cdot \mathbb{I}[j \neq t]$.
$\alpha$ and $\gamma$ are parameters of the focal loss, {\small $\mathbb{I}[\cdot]$} is the indicator function, $\sigma^{\text{cls}}_{i,j} = \text{cos}(\mathbf{q}_i, \mathbf{z}_\text{text}^j)$ is the language-guided classification score corresponding to the text prompt $\mathbf{z}_\text{text}^t$. We replace the standard focal loss for classification in object detection with our language-guided calibration loss.

\paragraph{Uncertainty-guided Query Pruning.}
The number of decoder queries is typically far greater than the number of objects in an image. While this is important during training to learn multiple object prototypes, it results in increased inference time due to the quadratic computational complexity of self attention. To this end, we propose to leverage the predictive scores calibrated through our uncertainty calibration loss to identify irrelevant queries for a given test image, and progressively prune them across layers to reduce the computational complexity.

Given a decoder with $L$ layers, we define the relevance threshold for each layer $l$ as a sigmoidal growth function:
{ \small
\begin{align}
\tau(l) = b_{\text{low}} + (b_{\text{high}} - b_{\text{low}}) / \left(1 + e^{-\frac{10\beta}{L} \cdot \left(x - \frac{L}{2}\right)}\right)
\end{align}
}
where \( \beta \) controls the steepness of the transition, and $b_{\text{low}}$ and $b_{\text{high}}$ represent the lower and upper bounds of the threshold. 
After each layer $l$, queries with predictive confidence below the layer-wise relevance threshold are deemed irrelevant and dropped. We find our sigmoidal growth function to provide a smooth transition, allowing for gradual query pruning across layers while retaining high-confidence queries compared to other alternatives (\cref{fig:pruning}).
On average, our approach reduces the transformer decoder FLOPs by an additional 50\% with minimal performance drop.

\input{new_tables/mobile-results}
\input{tables/main_results}
% ablations
\input{tables/downscaling}

\subsection{Towards a Unified Training Stage} \label{ssec:method_unified_training}
While GLEE~\cite{glee} is the first model to unify instance segmentation tasks across datasets, it relies on an inefficient multi-stage curriculum-learning pipeline. Its unimodal MaskDINO pretraining on COCO, multi-modal tuning on Objects365, and final finetuning on all datasets result in an overly complex training process.
In our experiments, we found that a multi-modal GLEE architecture could not even converge on COCO without unimodal MaskDINO pretraining. We traced this instability to their use of a simple dot product for language-guided classification. Since the dot product is unbounded, its values can arbitrarily explode or vanish, causing severe instability.
Replacing the dot product with cosine similarity {\small$\text{cos}_s(\mathbf{q}_i, \mathbf{z}_\text{text}^j) = exp(s) \cdot \mathbf{q}_i \cdot \mathbf{z}_\text{text}^j/(\|\mathbf{q}_i\| \|\mathbf{z}_\text{text}^j\|)$} with learnable scaling $s$ provides a simple yet effective fix, enabling smooth convergence on COCO. However, training across all datasets and tasks in a single stage remained unstable. We found that \emph{combining cosine similarity with learnable scaling and language-guided uncertainty calibration loss} fully stabilizes training.
Without calibration, query confidence scores can fluctuate arbitrarily, leading to gradient instability and poor convergence. The uncertainty calibration loss aligns classification confidence with localization accuracy (IoU), ensuring well-calibrated predictions throughout training. This prevents overconfident misclassifications, improves gradient consistency, and mitigates confidence collapse in early training. 
As a result, our approach enables stable single-stage training from scratch on diverse datasets (\cref{tab:unified_training}), reducing training iterations to just one-third of GLEE’s, improving efficiency, and democratizing foundation model research.

%% file: figures/method.tex
\begin{figure}[t]
    \centering
    \includegraphics[width=\columnwidth,trim=1.27cm 1.22cm 1.91cm 1.04cm, clip]{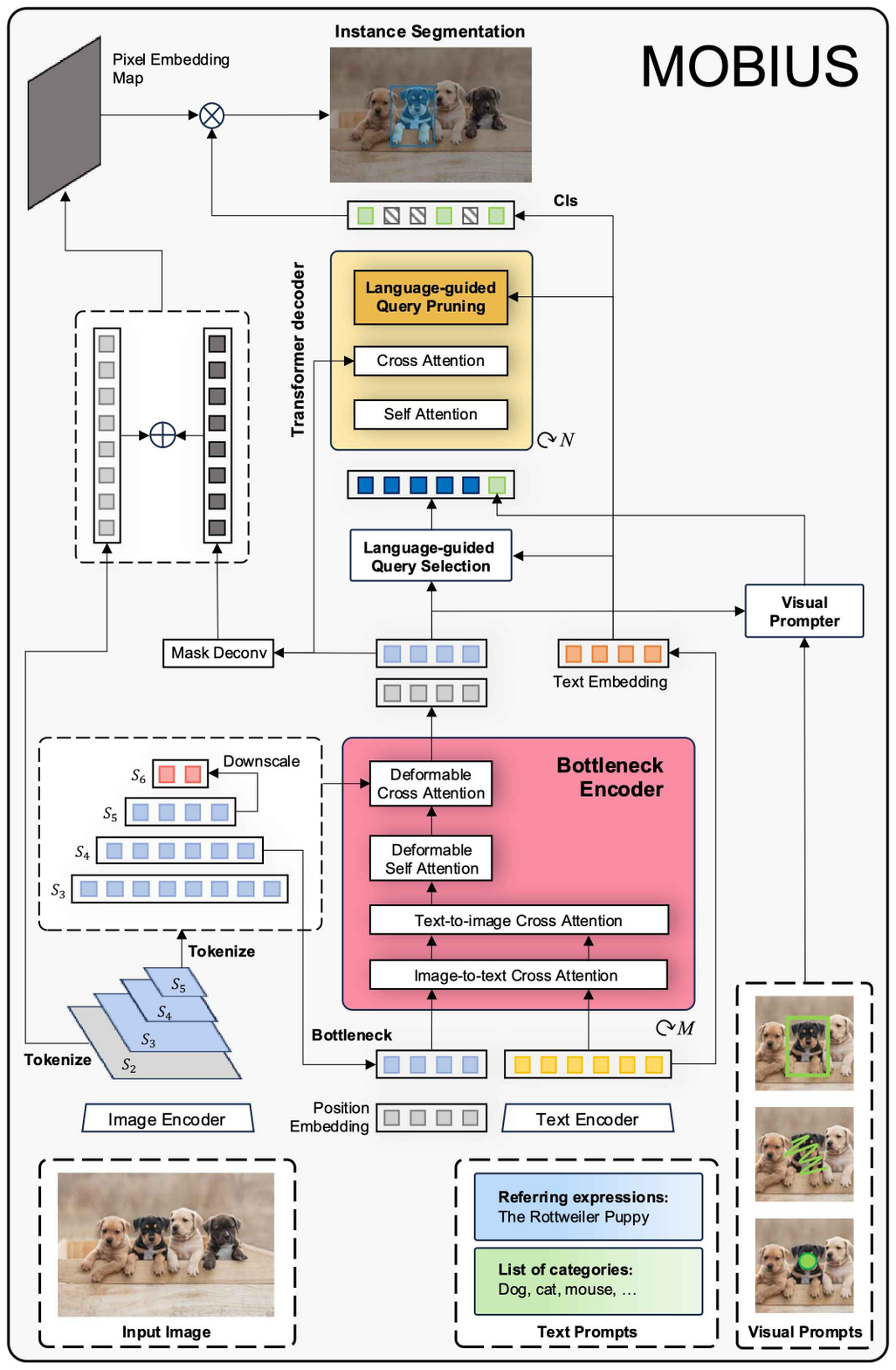} % Adjust path as needed
    % \fbox{\rule[0pt]{1.0\columnwidth}{2.0\columnwidth}} % Placeholder box
    \caption{
        Overview of the \methodNAME{} framework. The figure illustrates the core components: (i) the novel pixel decoder for efficient multi-scale and multi-modal fusion, and (ii) the transformer decoder with pruning strategy. This design enables Pareto-efficient downscaling for universal instance segmentation.
    }
    \label{fig:method}
\end{figure}

%% file: new_tables/mobile-results.tex
\begin{table*}[!t]
\centering
\scriptsize
\setlength{\tabcolsep}{2pt}
\renewcommand{\arraystretch}{1.2} % Adjust row height (optional)
\resizebox{1.0\linewidth}{!}{
\begin{tabular}{l lllccccccccccccc} 
\toprule
\multirow{3}{*}{Tag} & \multirow{3}{*}{Model} & \multirow{3}{*}{Backbone} & \multicolumn{2}{c}{\makecell{Pix. Dec.}} & \multicolumn{2}{c}{COCO} & \multicolumn{2}{c}{LVIS} & ODinW & \multicolumn{2}{c}{Efficiency} & \multicolumn{4}{c}{Mobile Latency (ms)}  & \multicolumn{1}{c}{GPU Latency (ms)} \\ 
\cmidrule(lr){4-5} \cmidrule(lr){6-7} \cmidrule(lr){8-9} \cmidrule(lr){10-10} \cmidrule(lr){11-12} \cmidrule(lr){13-16} \cmidrule(lr){17-17}
 &  & & \multicolumn{1}{c}{Type} & Blocks & $\rm AP_{b}$ & $\rm AP_{m}$ & $\rm AP_{b}$ & $\rm AP_{m}$ & $\rm AP_{b}$ & FLOPs (G) & Param. (M) & Samsung S24 & Xiaomi 12 Pro & Snap. X Elite & Snap. 8 Elite & NVIDIA RTX 3090 \\ 
\midrule
\rowcolor{gray!20} a) & GLEE$^{\dagger}$~\cite{glee} & MNv4-CM & MaskDINO~\cite{li2023mask} & 3 & 41.8 & 37.1 & 28.9 & 26.2 & 37.0 & 30.6 & 29.3 & 436.6 & 728.4 & 579.2 & 505.1 & 48.4 \\
\midrule
b) & GLEE$^{\dagger}$~\cite{glee} & \multirow{2}{*}{\makecell{MNv4-CM}} & MaskDINO~\cite{li2023mask} & 1 & \underline{39.3} & 34.6 & \underline{27.0} & \underline{24.3} & \underline{32.8} & 25.9 {\color{gray}(-15.4\%)} & \textbf{27.8} {\color{gray}(-5.1\%)} & 259.0 {\color{gray}(-40.7\%)} & 422.5 {\color{gray}(-42.0\%)} & 315.2 {\color{gray}(-45.6\%)} & 292.2 {\color{gray}(-42.1\%)} & 44.2 {\color{gray}(-8.7\%)} \\
c) & GLEE$^{\dagger}$~\cite{glee} &  & RT-DETR~\cite{zhao2024detrs} & 1 & 35.3 & \textbf{36.4} & 22.8 & 23.4 & 30.7 & \underline{25.3} {\color{gray}(-17.3\%)} & 35.4 {\color{gray}(+20.8\%)} & 107.2 {\color{gray}(-75.5\%)} & 206.5 {\color{gray}(-71.6\%)} & 126.7 {\color{gray}(-78.1\%)} & 147.7 {\color{gray}(-70.7\%)} & 40.4 {\color{gray}(-16.5\%)} \\
d) & MOBIUS (Ours) &  & Bottleneck & 3 & \textbf{40.5} & \textbf{36.4} & \textbf{28.1} & \textbf{26.2} & \textbf{38.6} & \textbf{18.2} {\color{gray}(-40.5\%)} & \underline{29.4} {\color{gray}(+0.3\%)} & 127.1 {\color{gray}(-70.9\%)} & 235.5 {\color{gray}(-67.7\%)} & 158.3 {\color{gray}(-72.7\%)} & 137.8 {\color{gray}(-72.7\%)} & 40.6 {\color{gray}(-16.1\%)} \\
\midrule
e) & MOBIUS (Ours) & \makecell{MNv4-CL} & Bottleneck & 3 & 41.5 & 37.2 & 29.4 & 27.2 & 38.3 & 22.8 {\color{gray}(-25.5\%)} & 52.4 {\color{gray}(+78.8\%)} & 136.9 {\color{gray}(-68.6\%)} & 238.9 {\color{gray}(-67.2\%)} & 148.8 {\color{gray}(-74.3\%)} & 137.5 {\color{gray}(-72.8\%)} & 42.0 {\color{gray}(-13.2\%)} \\
\bottomrule
\end{tabular}
}
\caption{\textbf{Mobile Universal Instance Segmentation.} We compare MOBIUS against mobile versions of GLEE~\cite{glee}, using either its original MaskDINO~\cite{li2023mask} decoder or an RT-DETR~\cite{zhao2024detrs}-based decoder. The first GLEE row (highlighted in gray) represents the baseline implementation, directly following the original reference. $\dagger$ denotes GLEE models retrained with mobile backbones, following our unified training approach (\cref{ssec:method_unified_training}). All models share the MobileNetv4 (MNv4)~\cite{qin2024mobilenetv4} backbone and a 1024-dimensional decoder hidden space. We report instance segmentation performance, efficiency metrics, and latency on mobile and GPU devices, together with the relative percentage change wrt. the reference GLEE baseline. Latency is profiled on the Qualcomm AI Hub at 384$\times$384 resolution with float32 precision. The text encoder is excluded from efficiency and latency measurements. Parentheses indicate the relative percentage change wrt. the baseline.}
\label{tab:mobile_results}
\end{table*}

%% file: tables/main_results.tex
\begin{table}[!ht]
\centering
\resizebox{1.0\linewidth}{!}{
\scriptsize
\setlength{\tabcolsep}{2pt}
\begin{tabular}{llccccccccccc} 
\toprule
  & \multirow{4}{*}{Method} & \multirow{4}{*}{\makecell{FLOPs\\(G)}} & \multicolumn{6}{c}{ {\it{Generic Detection \& Segmentation}}}     & \multicolumn{1}{c}{ {\it{Zero-shot}}}         \\
 \cmidrule(lr){4-9} \cmidrule(lr){10-10} 

& & & \multicolumn{2}{c}{COCO-val}  & \multicolumn{4}{c}{LVIS}  & \multicolumn{1}{c}{ODinW}  \\
 \cmidrule(lr){4-5}   \cmidrule(lr){6-9}  \cmidrule(lr){10-10} 
\footnotesize & & &$\rm AP_{b}$  & $\rm AP_{m}$ & $\rm AP_{b}$ & $\rm AP^r_{b}$  & $\rm AP_{m}$  &$\rm AP^r_{m}$ & $\rm AP_{b} $       \\ 
\midrule
\scriptsize
\multirow{5}{*}{\rotatebox{90}{\textbf{Specialist}}} 
% & MDETR~\cite{kamath2021mdetr} & - & -   & -   & -   & -   & -   & -   & -\\
% & SeqTR~\cite{zhu2022seqtr} & - & -   & -   & -   & -   & -   & -    & -    \\
% & PolyFormer (L)~\cite{liu2023polyformer} & - &-   & -   & -   & -   & -   & -  & -  \\
& ViTDet-L ~\cite{li2022exploring} & - &57.6  &49.8   &51.2  &-  &46.0  &34.3  & -\\
& ViTDet-H ~\cite{li2022exploring} & - &58.7  &50.9   &53.4  &-  &48.1   &36.9  & - \\
& EVA-02-L~\cite{fang2024eva} & - &64.2  &55.0   &65.2  &-   &57.3  &-    & -\\
% & ODISE~\cite{xu2023open} & - & -   & -   & -   & -   & -  & -   & -  \\
& Mask2Former (L)~\cite{cheng2022masked} & - & -   & 50.1   & -   & -   & -   & -   & - \\
& MaskDINO (L)~\cite{li2023mask} & - & -   & 54.5   & -   & -   & -   & -   & -  \\
\midrule
\multirow{15}{*}{\rotatebox{90}{\textbf{Generalist}}} 
% & UniTAB (B)~\cite{yang2022unitab} & -           &-  & -  & -   & -    & -   & -   & - \\ 
% & OFA (L)~\cite{wang2022ofa} & -  &-  & -   & -   & -   & -   & -   & - \\
& Pix2Seq v2 ~\cite{chen2021pix2seq} & - &46.5   &38.2   & -    & -   & -   & -   & -  \\ 
% & Uni-Perceiver-v2 (B)~\cite{li2023uni} & - &58.6   &50.6   & -    & -   & -   & -   & -  \\
% & Uni-Perceiver-v2 (L)~\cite{li2023uni} & - &61.9  & 53.6  & -    & -   & -   & -   & -  \\
& UNINEXT (R50)~\cite{lin2023uninext} & - &51.3   &44.9    &36.4   & -   & -   & -   & -  \\
& UNINEXT (L)~\cite{lin2023uninext} & - &58.1   &49.6    & -   & -   & -   & -   & -\\
% & UNINEXT (H)~\cite{lin2023uninext} & - &60.6   &51.8    & -   & -   & -   & -   & - \\
% & GLIPv2 (B)~\cite{zhang2022glipv2} & - &-   & -   & -    & -   & -   & -   & - \\
% & GLIPv2 (H)~\cite{zhang2022glipv2} & - &-   & -   & -    & -   & -   & -   & - \\
& X-Decoder (B)~\cite{zou2023generalized} & - &- &45.8 & -   &45.8  & -   & -   & -    \\
& X-Decoder (L)~\cite{zou2023generalized} & - &- &46.7 & -   &47.1   & -   & -   & -   \\
& Florence-2 (L)~\cite{xiao2024florence} & - &43.4   & -   & -   & -   & -   & -  & - \\
\cmidrule{2-10}
& GLEE-Plus~\cite{wu2024general} & 704 &60.4   &53.0   &52.7  &44.5  &47.4  &40.4     & 48.3 \\
\cmidrule{2-10}
& GLEE-Lite~\cite{wu2024general} & 239 &55.0   &48.4   &44.2  &36.7  &40.2  &33.7     & 43.2 \\
& \methodNAME-3 & 354 &\textbf{ 57.7} & \textbf{51.0}   & \textbf{50.3} & \textbf{43.9} & \textbf{46.8}  & \textbf{41.2}  & \textbf{45.5} \\
& \methodNAME-2 & \textbf{206} & 56.4  & 49.5   & 47.5 & 37.5  & 44.3  & 35.6  & 43.8 \\
\cmidrule{2-10}
& \methodNAME-1 & 155 & \textbf{55.7} & \textbf{49.2}   & \textbf{46.3} & 36.5 & \textbf{43.0}  & 34.2 & \textbf{42.0} \\
& \methodNAME-0 & \textbf{123} & 54.3  & 48.2   & 45.0  & \textbf{37.6} & 41.8  & \textbf{35.0}   & 41.2 \\
% \midrule
% \multirow{5}{*}{\rotatebox{90}{\textbf{Low-res }}} & \methodNAME-3 & 116  &  \textbf{50.1}   & \textbf{45.4}   &   \textbf{40.1} &  \textbf{36.8 }& \textbf{37.3 }  &  \textbf{34.7}   & \textbf{40.5} \\
% & \methodNAME-2 & 69 &   48.7  & 43.4   &     36.1  &  31.1 & 33.8   & 29.8  & 39.1 \\
% \cmidrule{2-10}
% & \methodNAME-1 & 52 &  \textbf{47.4}   &  \textbf{42.3}  &    \textbf{35.2} &   \textbf{32.6}   & \textbf{32.8 } &  \textbf{32.4}   & \textbf{40.5} \\
% & \methodNAME-0 & \textbf{41} &   46.2  &  41.5  &      33.7 &  28.0   &  31.4 &  27.3  & 43.8 \\
\bottomrule
\end{tabular}
}
\caption{\textbf{Big Universal Instance Segmentation.} We compare \methodNAME to recent specialist and generalist models on object-level image tasks. Comparable models are ranked by descending FLOPs and divided into groups with similar FLOPs count. FLOPs are computed at 800×800 resolution, omitting the text encoder.
}
\label{tab:main_results}
\end{table}

%% file: tables/downscaling.tex
\begin{table}[!ht]
\centering
\scriptsize
\setlength{\tabcolsep}{2pt}
\renewcommand{\arraystretch}{1.2} % Adjusts row height (optional)
\resizebox{1.0\linewidth}{!}{
\begin{tabular}{llccccccccc} 
\toprule
\addlinespace[1em]
\multirow{3}{*}{Tag} & \multirow{3}{*}{\makecell[l]{Pixel\\Decoder}}
& \multirow{3}{*}[0.7em]{\rotatebox{90}{\makecell{Bottleneck}}}
& \multirow{3}{*}[0.2em]{\rotatebox{90}{\makecell{Decoder\\Scales}}} 
& \multirow{3}{*}[0.2em]{\rotatebox{90}{\makecell{Layers}}} 
& \multicolumn{2}{c}{FLOPs (G)} 
& \multicolumn{2}{c}{COCO-val} 
& \multicolumn{2}{c}{LVIS-minival} \\
\cmidrule(lr){6-7} \cmidrule(lr){8-9} \cmidrule(lr){10-11} 
& & & & & Pix. Dec. & Decoder & $\rm AP_{b}$  & $\rm AP_{m}$ & $\rm AP_{b}$ & $\rm AP_{m}$ \\ 
\addlinespace[0.2em]
\midrule
\rowcolor{gray!20} 
a) & MaskDINO & - & Multi & 6 & 222 & 20 & 49.2  & 43.8  & 42.1 & 38.7 \\
b) & MaskDINO & - & Single & 6 & 222 \textcolor{gray}{(0.0\%)} & 10 \textcolor{gray}{(-50.0\%)}  & 47.9  & 42.9  & 40.7 & 38.2 \\
c) & MaskDINO & - & Single & 1 & 114 \textcolor{gray}{(-48.6\%)} & 10 \textcolor{gray}{(-50.0\%)}  & 43.4  & 38.6  & 36.0 & 33.7 \\
\midrule
d) & RT-DETR & - & Multi & 1 & 102 \textcolor{gray}{(-54.1\%)} & 20 \textcolor{gray}{(0.0\%)} & 47.4  & 42.1  & 38.1 & 35.1 \\
e) & RT-DETR & - & Single & 1 & 95 \textcolor{gray}{(-57.2\%)} & 10 \textcolor{gray}{(-50.0\%)} & 46.8  & 42.2  & 36.7 & 35.3 \\
\midrule
f) & Ours & - & Multi & 6 & 222 \textcolor{gray}{(0.0\%)} & 20 \textcolor{gray}{(0.0\%)} & 49.2  & 43.9  & 42.0 & 38.7 \\
g) & Ours & 1/16 & Multi & 6 & 101 \textcolor{gray}{(-54.5\%)} & 20 \textcolor{gray}{(0.0\%)} & 47.9  & 42.5  & 40.8 & 37.7 \\
h) & Ours & 1/8 & Single & 6 & 200 \textcolor{gray}{(-9.9\%)} & 20 \textcolor{gray}{(0.0\%)} & 47.5  & 42.3  & 40.3 & 37.4 \\
i) & Ours & 1/16 & Single & 6 & 91 \textcolor{gray}{(-59.0\%)} & 10 \textcolor{gray}{(-50.0\%)} & 47.5  & 42.2  & 40.3 & 37.8 \\
\bottomrule
\end{tabular}
}
\caption{\textbf{Ablation on bottleneck encoder and single-scale decoding.} We analyze the downscalability of different pixel decoders by comparing their impact on computational efficiency (FLOPs), performance on COCO-val and open-set performance on LVIS-minival. We ablate on bottleneck size (reported as a ratio of the input image size), number of scales processed by the transformer decoder, and number of pixel decoder layers. All ablations are conducted under the 100k iterations setting. Parentheses indicate the relative percentage change wrt. the baseline.}
\label{tab:bottleneck}
\end{table}

%% file: sec/4_experiments.tex
\section{Experiments}

First, we provide implementation details in \cref{ssec:exp_implementation_details} and conduct a preliminary investigation to identify the pitfalls of existing architecture designs in \cref{ssec:exp_preliminary} and how MOBIUS addresses them.
We then compare to the state of the art using both mobile and large backbones, validating how MOBIUS trades off efficiency and performance in a Pareto-efficient fashion (\cref{ssec:exp_sota}).
We perform ablation studies in \cref{ssec:exp_ablation}, where (i) we validate the design of our bottleneck encoder and single-stage decoding, (ii) we demonstrate the effectiveness of our inference-time pruning strategy, and (iii) we show the importance of our training recipe to enable training across all datasets and tasks in a single unified training stage.
More in the supplement.

\subsection{Implementation Details} \label{ssec:exp_implementation_details}
\paragraph{Datasets.}  We follow GLEE~\cite{glee} and train our models on the object detection datasets Objects365~\cite{shao2019objects365} and OpenImages~\cite{krylov2021open} and on the instance segmentation datasets COCO~\cite{lin2014microsoft}, LVIS~\cite{gupta2019lvis} and BDD~\cite{seita2018bdd100k}, We further train on three video instance segmentation datasets (YTVIS19~\cite{yang2019video}, YTVIS21~\cite{yang2019video}, OVIS~\cite{qi2022occluded}) treating them as image datasets. We further employ datasets including referring descriptions (RefCOCO~\cite{nagaraja2016modeling}, RefCOCO+~\cite{nagaraja2016modeling}, RefCOCOg~\cite{nagaraja2016modeling}, VisualGenome~\cite{krishna2017visual}, RVOS~\cite{seo2020urvos}). Finally, we use the open-world segmentation datasets UVO~\cite{wang2021unidentified} and SA-1B~\cite{kirillov2023segany}, for which we set the category name to `object' and train according to the multi-modal instance segmentation pipeline. A comprehensive list of our training datasets and their details is in the supplement.

\input{tables/calibration-cosine}

\paragraph{Training Details.}
Unlike GLEE~\cite{glee}, we perform a single training stage across all datasets and tasks.
We use CLIP-B~\cite{radford2021learning} as text encoder.
In the spirit of providing practitioners model sizes for all needs, we train MOBIUS with mobile backbones (MobileNetv4~\cite{qin2024mobilenetv4}-Conv-M and -Conv-L) and with efficient big backbones (FasterViT~\cite{hatamizadeh2023fastervit}-0, -1, -2, -3), corresponding respectively to \methodNAME-Mini-M, -Mini-L, -0, -1, -2, -3. We initialize the MobileNetv4 models from ImageNet12K-pretrained weights, and the FasterViT from ImageNet1K-pretrained ones.
We use our bottleneck encoder~\cref{eq:bottleneck_operations} as pixel decoder to efficiently merge the vision-language modalities and the multiple features scales. We use 6 (3) layers with hidden dimension 2048 (1024) for big (mobile) backbones, and choose as representational bottleneck the feature map with stride 16.
We use a deformable transformer decoder with 9 layers based on MaskDINO, and use 300 queries. We use query denoising and hybrid matching~\cite{maskdino} to accelerate convergence.
We train our model with multi-scale training on 64 H100 GPUs with a batch size of 128 for 500,000 iterations in a single unified stage. We test on both high-resolution (short side resized to 800) and low-resolution images (short side resized to 384). When conducting ablations we train our model for 100k iterations using ResNet-50 as vision backbone.

\paragraph{Evaluation Details.} We compare MOBIUS to the state of the art on object-level image tasks, including COCO-val, LVIS, and ODinW~\cite{li2022grounded} benchmarks. We choose the established COCO dataset to evaluate the closed-set detection and instance segmentation performance, the LVIS benchmark to assess the open-set capabilities of our model, and the ODinW datasets to assess the zero-shot generalization performance of our models in the wild. We report the average score across 13 ODinW benchmarks. Alongside key performance metrics, we compare the computational efficiency in terms of FLOPs. $\rm AP_{b}$ ($\rm AP_{m}$) is short for $\rm AP_{box}$ ($\rm AP_{mask}$).

\paragraph{Baselines.} We compare MOBIUS against GLEE~\cite{glee} models leveraging different pixel decoders. Specifically, we compare two widely adopted pixel decoder designs: MaskDINO’s~\cite{maskdino} transformer encoder, commonly chosen for performance~\cite{liu2024grounding,glee}, and RT-DETR’s~\cite{zhao2024detrs} hybrid pixel decoder, preferred for efficiency~\cite{zhao2024real,ren2024grounding}. We further compare against the naive efficient baseline represented by reducing the number of MaskDINO pixel decoder blocks to 1.

\subsection{Efficiency Analysis} \label{ssec:exp_preliminary}

\paragraph{Component-wise FLOPs Comparison.} In \cref{fig:relative_flops}, we analyze the FLOPs of different model components as a percentage of a fixed R50 vision encoder (52.4 GFLOPs). We find that the MaskDINO pixel decoder requires up to 263\% the FLOPs of the vision backbone. Moreover, modality fusion alone consumes as much as 54\% of the vision encoder FLOPs. Finally, the transformer decoder is equivalent to 38\% of the vision encoder. Replacing the MaskDINO pixel decoder with our bottleneck encoder (\cref{ssec:method_pixel_decoder}) significantly lightens the model, with an overall FLOPs reduction of -45.6\%. By acting on a lower-dimensional representation, our bottleneck encoder reduces the pixel decoder cost by -55.5\%, and the modality fusion by -79.6\%. Our single scale decoding additionally halves the decoder FLOPs. 

\paragraph{Performance-efficiency Trade-off.}
While MaskDINO excels in in-domain and open-vocabulary settings, it comes at a high computational cost (\cref{tab:mobile_results}, a). Both the naive baseline consisting of leveraging only 1 MaskDINO decoder layer (\cref{tab:mobile_results}, b) and RT-DETR's pixel decoder (\cref{tab:mobile_results}, c) result in a $\sim$15\% FLOPs reduction, while MOBIUS in $\sim$40.5\%.
While the RT-DETR pixel decoder would result in a similar latency reduction as our bottleneck encoder, it compromises the open-vocabulary performance (\cref{tab:mobile_results}, c). In particular, MOBIUS's bottleneck encoder (\cref{tab:mobile_results}, d-e) results in a 28.1 $\rm AP_{b}$ on LVIS and 38.6 $\rm AP_{b}$ on ODinW, far higher than RT-DETR's 22.8 and 30.7.  

\paragraph{Latency Evaluation.}
In \cref{tab:mobile_results} we evaluate the latency of all models on mobile and GPU devices at 384x384 resolution. 
As mentioned above, RT-DETR's latency reduction comes at significant open-vocabulary performance costs.
Crucially, we find that MOBIUS reduces the mobile latency by  $\sim$70\% across all edge devices compared to the GLEE-MaskDINO baseline, while retaining competitive performance and outscoring all efficient baselines. Unlike GLEE - which takes 0.8s to process one image on a Xiaomi 12 Pro - MOBIUS runs real-time on a variety of edge devices, achieving 127ms on the flagship Samsung Galaxy S24 and 235ms on the older Xiaomi 12 Pro.
We use float32 precision everywhere except for the Snapdragon 8, where we apply uint8 quantization to validate the compatibility of MOBIUS with the power-efficient formats. This quantization reduces peak memory consumption from 200MB to just 15MB, further enhancing MOBIUS’s suitability for deployment in resource-constrained environments.

\subsection{State of the Art Comparison} \label{ssec:exp_sota}
\paragraph{Mobile Universal Instance Segmentation.}
In \cref{tab:mobile_results}, we validate the efficiency and performance of mobile MOBIUS models against GLEE~\cite{glee} models leveraging different pixel decoders. All models are trained using our unified training strategy, uncertainty calibration loss, and share the same MobileNetv4 conv-M backbone. For completeness, we train MOBIUS with a MNv4-conv-L backbone (row e).
Of all the efficient pixel decoders (rows b-d), we find that only MOBIUS's bottleneck encoder (row d) remains competitive with the large MaskDINO pixel decoder (row a). Remarkably, MOBIUS performs even better than GLEE-MaskDINO out-of-distribution, reporting an impressive 38.6 $\rm AP_{b}$ on ODinW compared to GLEE-MaskDINO's 37.0 and GLEE-RT-DETR's 30.7.

\paragraph{Big Universal Instance Segmentation.}
In \cref{tab:main_results}, we provide a detailed comparison of big \methodNAME models against state-of-the-art specialist and generalist models. 
We evaluate the Pareto-efficiency of our big models and rank them in descending order by FLOPs.
\methodNAME models demonstrate a remarkable balance between computational efficiency and task performance. For instance, {MOBIUS-3} achieves a COCO-val $\rm AP_{b}$ of 57.7 and LVIS $\rm AP_{b}$ of 50.3 while operating at 354G FLOPs, a significant reduction compared to {GLEE-Plus}, which requires 704G FLOPs to achieve only slightly higher $\rm AP_{b}$ scores of 60.4 and 52.7, respectively. Among our smaller models, {MOBIUS-1} notably outperforms GLEE-Lite with 35\% less FLOPs.

\input{tables/pruning}

\subsection{Ablation Study} \label{ssec:exp_ablation}

% We conducted a comprehensive ablation study to validate the key design choices and contributions of \methodNAME. The study focuses on three main aspects: (i) the design of the bottleneck encoder and single-scale decoding, (ii) the effectiveness of our inference-time pruning strategy, and (iii) the unified training approach.

\paragraph{Bottleneck Encoder and Single-Scale Decoding.}
\cref{tab:bottleneck} compares baseline pixel decoders to various configurations of our bottleneck encoder, analyzing the effect of different bottleneck strides and the use of multi-scale decoding. 
We find that: (i) using a bottleneck stride of 16 (row i) performs competitive with the 4$\times$ larger bottleneck obtained with stride 8 (row h), but with 55\% less FLOPs. Similarly, single-scale decoding (row i) performs similar to multi-scale decoding (row g) for MOBIUS, but with 10G FLOPs less. This demonstrates the effectiveness of condensing multi-scale information into a single expressive representation, while competitors' performance drops significantly when decoding only a single scale (rows a-b and d-e).

\paragraph{Inference-Time Pruning Strategy.}
\cref{fig:pruning} evaluates the impact of different query pruning strategies on performance and computational efficiency. Our language-guided uncertainty calibration enables progressive query pruning, reducing transformer decoder FLOPs by an additional 50\%. For instance, our pruning strategy based on sigmoidal growth achieves an $\rm AP_{m}$ of 44.0 on COCO-val with minimal performance loss compared to the full set of queries.

\paragraph{Unified Training Approach.}
Table~\ref{tab:unified_training} highlights the advantages of our unified training paradigm (\cref{ssec:method_unified_training}), comparing the convergence of a GLEE model to a MOBIUS without bottleneck (-H) for fair comparison. Unlike GLEE, which requires a multi-stage training process, \methodNAME achieves stable convergence in a single stage. Convergence on COCO is facilitated by our scaled cosine similarity (row c), which does not suffice for joint training stability (row g). Its combination with our uncertainty calibration loss (row h) improves model stability and enables MOBIUS convergence in a third of GLEE's training iterations.

%% file: tables/calibration-cosine.tex
\begin{table}[!t]
\centering
\scriptsize
\setlength{\tabcolsep}{2pt}
\renewcommand{\arraystretch}{1.2} % Adjusts row height (optional)
\resizebox{1.0\linewidth}{!}{
\begin{tabular}{llccccccc} 
\toprule
& \multirow{3}{*}{\makecell{Method}} & \multirow{3}{*}{\makecell{Scaled\\Cosine}}
& \multirow{3}{*}{\makecell{Calibration}}
& \multirow{3}{*}{\makecell{Training\\Stages}}
& \multicolumn{2}{c}{COCO-val} 
& \multicolumn{2}{c}{LVIS-minival} \\
\cmidrule(lr){6-7} \cmidrule(lr){8-9} 
& & & & & $\rm AP_{box}$  & $\rm AP_{mask}$ & $\rm AP_{box}$ & $\rm AP_{mask}$ \\ 
\midrule
\multirow{4}{*}{\rotatebox{90}{\textbf{COCO}}} 
& (a) MaskDINO~\cite{maskdino} & - & - & Single & 45.9 & 41.3 & -  & -  \\
& (b) GLEE-Lite~\cite{glee} & - & - & Single & \multicolumn{4}{c}{D.N.C.} \\
& (c) MOBIUS-H-R50 & \checkmark & - & Single & 45.9 & 41.3 & -  & - \\
& (d) MOBIUS-H-R50 & \checkmark & \checkmark & Single & \textbf{46.5} & \textbf{41.9} & -  & - \\
\midrule
\multirow{4}{*}{\rotatebox{90}{\textbf{Joint}}} 
& (e) GLEE-Lite~\cite{glee} & - & - & Single & \multicolumn{4}{c}{D.N.C.} \\
& (f) GLEE-Lite~\cite{glee} & - & - & Multi & \textbf{50.0} & \textbf{48.4} & 50.5  & 45.9 \\
& (g) MOBIUS-H-R50 & \checkmark & - & Single & \multicolumn{4}{c}{D.N.C.} \\
& (h) MOBIUS-H-R50 & \checkmark & \checkmark & \textbf{Single} & \textbf{50.0} & \textbf{48.4} & \textbf{50.7}  & \textbf{46.0} \\
\bottomrule
\end{tabular}
}
\caption{\textbf{Ablation on the unification of training stages.} We ablate on the importance of our simple yet necessary tricks to improve the model stability and enable training across all datasets and tasks in a single unified stage. We ablate on the application of scaled cosine similarity and uncertainty calibration loss, and report the Average Precision (AP) for box and mask predictions on COCO-val and LVIS-minival. D.N.C. stands for ``did not converge''. For unified training we follow the 1x schedule on COCO and the 100k schedule on joint. All models use R50.
}
\label{tab:unified_training}
\end{table}

%% file: tables/pruning.tex
\definecolor{darkgreen}{RGB}{0,120,0} % Adjusted dark green color

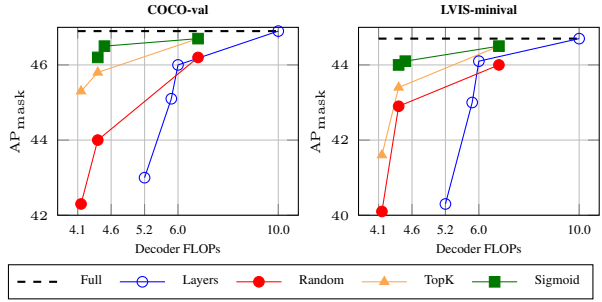
\begin{figure}[!t]
\centering
\begin{tikzpicture}
\begin{groupplot}[
    group style={
        group size=2 by 1,
        horizontal sep=0.8cm, % Reduced space between plots
    },
    width=3.2cm,
    height=2.5cm,
    scale only axis=true, % Ensures width and height apply only to the axis area
    title style={font=\tiny, yshift=-1ex}, % Adjusts title position
    xlabel={Decoder FLOPs},
    ylabel={$\rm AP_{mask}$},
    xlabel style={font=\tiny, yshift=0.7ex}, % Shift x-axis labels closer
    ylabel style={font=\tiny, yshift=-1ex}, % Shift y-axis labels closer
    tick label style={font=\tiny},
    label style={font=\tiny},
    grid=major,
    xtick={4, 5, 6, 7, 10},
    xticklabels={4.1, 4.6, 5.2, 6.0, 10.0},
    legend columns=5,
    legend style={
        font=\tiny,
        cells={anchor=west},
        column sep=1ex,
    },
    % Removed cycle list since we're specifying colors manually
]

% First plot
\nextgroupplot[
    title={\textbf{COCO-val}},
    legend to name=sharedlegend,
    ymin=42.0, ymax=47,
]

% Baseline horizontal line
\addplot[thick, dashed, black] coordinates {
    (4, 46.9) (10, 46.9)
};
\addlegendentry{Full}

% Layers
\addplot[mark=o, blue] coordinates {
    (10, 46.9) (7, 46.0) (6.8, 45.1) (6.0, 43.0)
};
\addlegendentry{Layers}

% Random
\addplot[mark=*, red] coordinates {
    (7.6, 46.2) (4.6, 44.0) (4.1, 42.3)
};
\addlegendentry{Random}

% TopK
\addplot[mark=triangle*, color=orange!70!white] coordinates {
    (7.6, 46.7) (4.6, 45.8) (4.1, 45.3)
};
\addlegendentry{TopK}

% Sigmoid
\addplot[mark=square*, color=darkgreen] coordinates {
    (7.6, 46.7) (4.8, 46.5) (4.6, 46.2)
};
\addlegendentry{Sigmoid}

% Second plot
\nextgroupplot[
    title={\textbf{LVIS-minival}},
    ymin=40, ymax=45,
]

% Baseline horizontal line
\addplot[thick, dashed, black] coordinates {
    (4, 44.7) (10, 44.7)
};

% Layers
\addplot[mark=o, blue] coordinates {
    (10, 44.7) (7, 44.1) (6.8, 43.0) (6.0, 40.3)
};

% Random
\addplot[mark=*, red] coordinates {
    (7.6, 44.0) (4.6, 42.9) (4.1, 40.1)
};

% TopK
\addplot[mark=triangle*, color=orange!70!white] coordinates {
    (7.6, 44.5) (4.6, 43.4) (4.1, 41.6)
};

% Sigmoid
\addplot[mark=square*, color=darkgreen] coordinates {
    (7.6, 44.5) (4.8, 44.1) (4.6, 44.0)
};

\end{groupplot}
\end{tikzpicture}

% Place the shared legend
\centering
\pgfplotslegendfromname{sharedlegend}

\caption{\textbf{Ablation on pruning strategies.} We compare the effect of different pruning on the number of decoder FLOPs and the $\rm AP_{mask}$ on COCO-val and LVIS-minival datasets.} \label{fig:pruning}
\end{figure}

%% file: sec/5_conclusion.tex
\section{Conclusion}
We introduced MOBIUS, a Pareto-efficient family of big-to-mobile universal instance segmentation models, balancing scalability, efficiency, and performance. MOBIUS enables real-time deployment across high-end accelerators and edge devices without compromising accuracy.
At its core, our bottleneck pixel decoder compresses multi-scale, multi-modal information, reducing pixel decoder FLOPs by 55\% while preserving open-vocabulary performance. Our single-scale transformer decoder eliminates redundant multi-scale processing, cutting FLOPs by 50\%, while language-guided uncertainty calibration enables adaptive decoder pruning, further halving transformer decoder computational cost.
Additionally, our unified single-stage training removes the need for multi-stage curriculum learning, reducing training iterations to one-third of GLEE’s. Experiments validate state-of-the-art efficiency and performance trade-offs, with real-time inference at 10 FPS on mobile devices and 25 FPS on GPUs. MOBIUS sets a new benchmark for scalable, generalist perception models, paving the way for broader real-world adoption in both high-performance and resource-constrained environments.

%% file: sec/X_suppl.tex
\clearpage
\setcounter{page}{1}
\maketitlesupplementary

We here report additional implementation details (\cref{sec:supp:implementation}) and state-of-the-art comparison on additional datasets (\cref{sec:supp:sota}).
Moreover, we extend our ablation study and include analysis on the component-wise efficiency (\cref{sec:supp:abl:componentwise_efficiency}), the different mobile encoders and the relative computational complexity of our decoders (\cref{sec:supp:abl:mobile_encoders}), the FLOPs at low image resolution (\cref{sec:supp:abl:low_res_flops}), decoder design choices (\cref{sec:supp:abl:decoder_design}), the effect of calibration on decoder pruning (\cref{sec:supp:abl:calibration_pruning}), and different confidence trajectory functions (\cref{sec:supp:abl:conf_trajectory}).
Finally, we provide qualitative results for the different tasks supported by our foundational universal instance segmentation model (\cref{sec:supp:qualitative}).

\section{Implementation Details} \label{sec:supp:implementation}
\paragraph{Datasets.} In \cref{ssec:exp_implementation_details}, we have described the datasets that we used for training our model. We here report additional details in table \cref{table:dataset_list}. Notice that, unlike GLEE~\cite{wu2024general}, \methodNAME is trained in a single stage across all listed datasets. The table also reports the sampling ratio for each dataset. Following GLEE, to ensure that objects from SA1B are at the object-level rather than the part-level, we apply mask IoU based NMS and use area as NMS score to eliminate part-level object annotations.

\paragraph{Additional Training Details.} To ensure full reproducibility of our approach, we here report additional training details to the ones reported in \cref{ssec:exp_implementation_details}. In particular, we train our model for 500,000 iterations on the joint set of datasets listed in \cref{table:dataset_list}. We use the AdamW~\cite{loshchilov2017decoupled} optimizer with learning rate $10^{-4}$ and weight decay of 0.05. We decay the learning rate twice by a factor of 0.1 after 400k and 500k iterations respectively. The learning rates of the image encoder and text encoder are multiplied by a factor of 0.1. We use multi-scale augmentation, and resize the input images such that the shortest side is at least 384 and at most 800 pixels while the longest at most 1333.

\input{tables/datasets/datasets}
\input{tables/lowres-results}
\input{tables/refcoco}
\input{tables/seginw}

\input{tables/odinw}

\section{Additional State-of-the-art Comparisons} \label{sec:supp:sota}

\paragraph{Low-resolution evaluation.} For completeness, we provide the low-resolution performance of our big models (\cref{tab:lowres_results}), so that they can be fairly compared to our mobile models in \cref{tab:mobile_results}. This analysis further demonstrate the adaptability of \methodNAME models. At 89G FLOPs, \textbf{MOBIUS-3 (low-res)} achieves a COCO-val $\rm AP_{b}$ of 50.8 and LVIS $\rm AP_{b}$ of 40.2, with a modest performance drop compared to its high-resolution counterpart (COCO-val $\rm AP_{b}$ of 57.7 and LVIS $\rm AP_{b}$ of 50.3 at 354G FLOPs). Lower-tier models, such as \textbf{MOBIUS-0 (low-res)}, operate at just 33G FLOPs while maintaining competitive performance (COCO-val $\rm AP_{b}$ of 46.9). Nevertheless, the smallest big model still requires almost twice as many FLOPs as our mobile model based on MNv4-conv-M (\cref{tab:mobile_results}, d). These results highlight the suitability of \methodNAME models for resource-constrained platforms, such as mobile and edge devices.

\paragraph{RefCOCO - Referring Object Detection and Segmentation.}
We report a state-of-the-art comparison on the RefCOCO, RefCOCO+ and RefCOCOg datasets in \cref{tab:refcoco_results}. For each dataset, we report the P@0.5 and the oIoU. We find that, despite the decreased number of FLOPs, our model remains effective in grounding referring expressions. However, we want to highlight that, while switching from ResNet-50 to FasterViT variants allowed us to leverage a more edge-friendly architecture, it seems that FasterViT provides a worse initialization for the referring tasks. We indeed report the performance of a MOBIUS variant trained with R50 and find that, despite having a number of FLOPs comparable to MOBIUS-0, it achieves much higher referring performance. We hope that this insight will guide future researchers towards choosing more suitable vision encoder initializations for referring and grounding.

\paragraph{ODinW - Zero-shot Object Detection.}
We report a state-of-the-art comparison on 13 ODinW~\cite{li2022grounded} datasets in \cref{table:odinw_zero_shot}, benchmarking the zero-shot generalization of our models for the object detection task.
We find that our model remains competitive with GLEE-Lite while achieving better efficiency, with MOBIUS-3 even outperforming GLEE-Lite (45.5 vs 43.2 average box AP)

\paragraph{SegInW - Zero-shot Instance Segmentation.}

We report a state-of-the-art comparison on 22 SegInW~\cite{zou2023generalized} datasets in \cref{table:seginw_results}, benchmarking the zero-shot generalization of our models for the instance segmentation task.
Remarkably, we find that our model outperforms all prior methods (47.3 average mask AP with MOBIUS-3), exhibiting already competitive performance with its smallest size MOBIUS-0.

\section{Additional Ablation Studies} \label{sec:supp:ablation}

\input{new_tables/component-efficiency}
\input{tables/fastervit_mobilenet}
\input{tables/lowres_flops}
\input{tables/bottleneck_decoder}
\input{tables/pruning_coco}
\input{tables/pruning_trajectory}

\subsection{Component-wise Efficiency Analysis} \label{sec:supp:abl:componentwise_efficiency}
In \cref{tab:component_efficiency}, we report the component-wise numerical FLOPs values used to generate \cref{fig:relative_flops}.

\subsection{Mobile Encoders} \label{sec:supp:abl:mobile_encoders}
We show in \cref{tab:mobilenet} that further downscaling can be allowed by switching the vision encoder from FasterViT~\cite{hatamizadeh2023fastervit} to MobileNetv4~\cite{qin2024mobilenetv4}. While FasterViT has been optimized for performance / throughput trade-off on high-end and edge GPUs, different versions of MobileNetv4 have also been optimized for  performance / throughput trade-off on different mobile devices. As can be seen from our comparison, MobileNetv4 variants require significantly less FLOPs. Nevertheless, despite the larger FLOPs count, FasterViT retains good latency and provides significantly better detection performance. For this reason, we prefer leveraging the efficient FasterViT in our experiments in the main paper so to fairly compete with GLEE-Lite. Nevertheless, the results in \cref{tab:mobilenet} show that further downscaling of our model can be enabled by using one of the MobileNetv4 architectures, trading off performance for less compute requirements.

\subsection{Low-resolution FLOPs} \label{sec:supp:abl:low_res_flops}
In \cref{tab:lowres_flops_per_component} we compare the FLOPs requirements of different \methodNAME variants and GLEE under the low-resolution setting, where images are rescaled to 384 on their short side while preserving aspect ratio. The results show that the computational complexity of our pixel decoder and transformer scales down nicely with the input image size, still resulting in less FLOPs than the corresponding vision encoders (except for MOBIUS-0). 
Moreover, even at smaller resolution, using our bottleneck encoder as pixel decoder results in only 41\% of GLEE's pixel decoder FLOPs. Finally, thanks to our single-scale processing, our transformer decoder only takes 50\% on GLEE's.

% \subsection{Latency Analysis} \label{sec:supp:abl:latency}
% \cref{tab:latency_per_component} (reported for convenience in \cref{tab:latency_per_component_supp} in the supplement) provides a breakdown of latency (in milliseconds) across key components of MOBIUS and GLEE models on an NVIDIA A100 GPU. Notice that we benchmark the latency of each component independently and do not enable automatic parallelization of different model components for this analysis; enabling parallelization enables a 20\% speed-up. MOBIUS demonstrates substantial latency improvements over GLEE across all components, particularly in the vision encoder and pixel decoder stages. For example, \textbf{MOBIUS-3} reduces the pixel decoder latency by over 36\% compared to \textbf{GLEE-Plus} (43.7ms vs. 69.1ms), while also reducing the total latency from 350.1ms to 217.5ms. Lower-tier models, such as \textbf{MOBIUS-0}, achieve further latency reductions, with a total latency of just 179.7ms, making them highly suitable for real-time applications on resource-constrained devices. These results highlight the efficiency of MOBIUS’s architectural innovations, including the bottleneck pixel decoder and single-scale decoding, which optimize performance while significantly reducing latency. Ultimately, all of our proposed models result faster on a high-performance computing accelerator than the smallest version of GLEE (Lite).
% %
% Consequently, we expect this gap to be far wider on edge devices, where complex operations such as attention are not nearly as optimized as on modern GPU architectures.

\subsection{Decoder Design} \label{sec:supp:abl:decoder_design}
In \cref{tab:pixel_decoder_ablation} we ablate on different design choices for our pixel decoder. In particular, we ablate on the COCO dataset on the effect on FLOPs and performance of: type of self-attention used, bottleneck size, number of pixel decoder layers, whether to use single or multiple scales in the transformer decoder.
We find that: (i) deformable self-attention - enabled by our smart design of the bottleneck representation as an individual scale from the feature scale pyramid - achieves the same performance as standard self-attention but with a  significantly lower FLOPs count; (ii) the bottleneck size, measured according to the feature stride selected, saturates at stride 16, with the smaller stride 32 resulting in lower performance but better efficiency; (iii) the performance can greatly vary based on the number of pixel decoder layers, and we thus advise practitioners to choose the number of layers based on their computational budget; (iv) thanks to the multi-modal and multi-scale fusion happening within our pixel decoder, leveraging a single scale or multiple scales in the transformer decoder does not result in a significant difference, and we thus advise to use a single scale to improve efficiency.

\subsection{Effect of Uncertainty Calibration on Query Pruning} \label{sec:supp:abl:calibration_pruning}
In \cref{tab:calibration_pruning_ablation_coco}, we investigate the effect of uncertainty calibration on query pruning on the COCO dataset. Importantly, we find that uncertainty calibration enables more meaningful differentiation of relevant vs. irrelevant queries, enabling better performance when applying query pruning at inference time.

\subsection{Confidence Trajectory Functions} \label{sec:supp:abl:conf_trajectory}
In \cref{tab:sigmoid_logarithm_exponential} we investigate the effect of different confidence trajectories for our query pruning strategy. As explained in \cref{ssec:method_pixel_decoder}, our query pruning strategy relies on a threshold that increases layer-by-layer following a sigmoidal trajectory. We here compare to a logarithmic and exponential trajectory. Each strategy results in a different increase steepness for the confidence threshold at different layers. Empirically, we find that the sigmoidal trajectory, which enables slower increase at the beginning and end of the decoder with a steeper increase in the middle layers, works slightly better under its most FLOPs-efficient setting.

\paragraph{Exponential Interpolation}

Exponential interpolation gradually increases the confidence threshold in an exponential manner. This method is particularly useful when you want to retain more queries in the early layers and prune more aggressively in the later layers.

\begin{equation}
\text{thr}(l) = \text{l} + (\text{u} - \text{l}) \times \frac{e^{\alpha \times \frac{l}{L-1}} - 1}{e^{\alpha} - 1}
\end{equation}

Here, \( l \) is the current layer index, \( L \) is the total number of layers, and \( \alpha \) is a parameter that controls the steepness of the curve. The threshold starts at \( \text{l} \) and approaches \( \text{u} \) as \( l \) increases.

\paragraph{Logarithmic Interpolation}

Logarithmic interpolation increases the confidence threshold logarithmically. This method allows for a rapid increase in the threshold in the early layers, which then slows down in the later layers. It is ideal for scenarios where you want to prune more aggressively in the initial layers.

\begin{equation}
\text{thr}(l) = \text{l} + (\text{u} - \text{l}) \times \frac{\log(1 + \alpha \times \frac{l}{L-1})}{\log(1 + \alpha)}
\end{equation}

In this equation, \( \alpha \) is a parameter that controls the curve's steepness. The threshold starts at \( \text{l} \) and grows rapidly at first, then gradually levels off as it approaches \( \text{u} \).

\paragraph{Sigmoid Interpolation}

Sigmoid interpolation provides a smooth, S-shaped curve that starts slowly, increases more rapidly in the middle layers, and slows down again as it approaches the upper layers. This method is useful when a balanced, gradual transition is desired.

\begin{equation}
\text{thr}(l) = \text{l} + (\text{u} - \text{l}) \times \frac{1}{1 + e^{-\beta \times \left(\frac{l - \frac{L}{2}}{L/10}\right)}}
\end{equation}

In this formula, \( \beta \) controls the steepness of the transition. The threshold starts at \( \text{l} \), increases more rapidly around the middle layers, and finally levels off as it approaches \( \text{u} \).

\section{Qualitative Results} \label{sec:supp:qualitative}
\input{figures/qualitative}
In table \cref{fig:qualitative_results} we show results for the following supported tasks for a variety of input images: (1) category-guided instance segmentation using COCO categories, (2) category-agnostic instance segmentation, (3) referring detection and segmentation.

%% file: tables/datasets/datasets.tex
\begin{table}[t]
\renewcommand\arraystretch{1.0}
\newcommand{\band}{\rowcolor{gray!10}}
    \small 
    \centering
    \resizebox{1.0\columnwidth}{!}{
    \begin{tabular}{lcccccc}
    \toprule
    &\multicolumn{2}{c}{Sizes} & \multicolumn{3}{c}{Annotations} &  \multirow{2}{*}{\makecell{Sampling\\ Ratio}} \\
    \cmidrule(lr){2-3} \cmidrule(lr){4-6}
    dataset & images & objects & semantic & box & mask & \\
    \midrule
    \band \textbf{Detection Data} &  &  &  &  &  & \\
    Objects365~\cite{shao2019objects365} & 1817287 & 26563198 & category & \checkmark & - & 1.5 \\
    OpenImages~\cite{krylov2021open} & 1743042 & 14610091 & category & \checkmark & - & 1.5 \\
    LVIS~\cite{gupta2019lvis} & 100170 & 1270141 & category & \checkmark & \checkmark & 1.5 \\
    COCO~\cite{lin2014microsoft} & 118287 & 860001 & category & \checkmark & \checkmark & 1.5 \\
    BDD~\cite{seita2018bdd100k} & 69863 & 1274792 & category & \checkmark & \checkmark & 0.15 \\
    \band \textbf{Grounding Data} &  &  &  &  &  & \\
    RefCOCO~\cite{nagaraja2016modeling} & 16994 & 42404 & description & \checkmark & \checkmark & \multirow{3}{*}{2.5$^\dagger$} \\
    RefCOCOg~\cite{nagaraja2016modeling} & 21899 & 42226 & description & \checkmark & \checkmark & \\
    RefCOCO+~\cite{nagaraja2016modeling} & 16992 & 42278 & description & \checkmark & \checkmark & \\
    VisualGenome~\cite{krishna2017visual} & 77396 & 3596689 & description & \checkmark & - & 2 \\
    \band \textbf{OpenWorld Data} &  &  &  &  &  & \\
    UVO~\cite{wang2021unidentified} & 16923 & 157624 & - & \checkmark & \checkmark & 0.2 \\
    SA1B~\cite{kirillov2023segany} & 2147712$^\ddagger$ & 99427126 & - & \checkmark & \checkmark & 2.5 \\
    \band \textbf{Video Data} &  &  &  &  &  & \\
    YTVIS19~\cite{yang2019video} & 61845 & 97110 & category & \checkmark & \checkmark & 0.3 \\
    YTVIS21~\cite{yang2019video} & 90160 & 175384 & category & \checkmark & \checkmark & 0.3 \\
    OVIS~\cite{qi2022occluded} & 42149 & 206092 & category & \checkmark & \checkmark & 0.3 \\
    % UVO-dense~\cite{UVO} & 45270 & 657990 & - & \checkmark & \checkmark & 0.3 \\
    % VOS~\cite{youtubevos} & 94588 & 156310 & - & \checkmark & \checkmark & - \\
    RefVOS~\cite{seo2020urvos} & 93857 & 159961 & description & \checkmark & \checkmark & 0.3 \\
    
    \bottomrule
    \end{tabular}}
    \caption{\textbf{Training Datasets.} The datasets used to train \methodNAME and the corresponding sampling ratio. We here process each frame in video datasets independently. $\dagger$: sampling ratio of the joint set including all RefCOCO datasets; $\ddagger$: we train on a subset of 500k images from the SA1B dataset.}
    \label{table:dataset_list}
    % \vspace{-3mm}
\end{table}

%% file: tables/lowres-results.tex
\begin{table}[!ht]
\centering
\resizebox{1.0\linewidth}{!}{
\scriptsize
\setlength{\tabcolsep}{2pt}
\begin{tabular}{llccccccccccc} 
\toprule
  & \multirow{4}{*}{Method} & \multirow{4}{*}{\makecell{FLOPs\\(G)}} & \multicolumn{6}{c}{ {\it{Generic Detection \& Segmentation}}}     & \multicolumn{1}{c}{ {\it{Zero-shot}}}         \\
 \cmidrule(lr){4-9} \cmidrule(lr){10-10} 

& & & \multicolumn{2}{c}{COCO-val}  & \multicolumn{4}{c}{LVIS}  & \multicolumn{1}{c}{ODinW}  \\
 \cmidrule(lr){4-5}   \cmidrule(lr){6-9}  \cmidrule(lr){10-10} 
\footnotesize & & &$\rm AP_{box}$  & $\rm AP_{mask}$ & $\rm AP_{box}$ & $\rm AP^r_{box}$  & $\rm AP_{mask}$  &$\rm AP^r_{mask}$ & $\rm AP_{box} $       \\ 
\midrule
\scriptsize
\multirow{5}{*}{\rotatebox{90}{\textbf{Low-res }}} 
& GLEE-Lite~\cite{wu2024general} & 59 & 47.2    &  42.1  & 35.0  & 31.9  & 31.2  &  23.0   & 40.5 \\
& \methodNAME-3 & 89  &  \textbf{50.8}   & \textbf{45.8}   &   \textbf{40.2} &  \textbf{37.7}& \textbf{37.9}  &  \textbf{35.3}   & \textbf{43.7} \\
& \methodNAME-2 & 53 &   49.6  & 44.2   & 37.8  &  32.0 & 35.4   & 30.7  & 43.1 \\
\cmidrule{2-10}
& \methodNAME-1 & 41 &  \textbf{48.0}   &  \textbf{43.0}  &    \textbf{36.3} &   \textbf{31.8}   & \textbf{34.0} &  \textbf{30.3}   & \textbf{43.2} \\
& \methodNAME-0 & \textbf{33} &   46.9  &  42.1  &      34.9 &  28.3   &  32.8 &  27.0  & 40.6 \\
\bottomrule
\end{tabular}
}
\caption{\textbf{Comparison of big models at low-res.} We compare \methodNAME to GLEE~\cite{glee} on object-level image tasks at low-resolution, rescaling the images to 384 on their short side while preserving aspect ratio. The models are ranked by descending FLOPs and divided into groups with similar FLOPs count. FLOPs are computed at 384x384 resolution, omitting the text encoder.
}
\label{tab:lowres_results}
\end{table}

%% file: tables/refcoco.tex
\begin{table}[!ht]
\centering
\resizebox{\linewidth}{!}{
\scriptsize
\setlength{\tabcolsep}{2pt}
\begin{tabular}{llcccccc} 
\toprule
  & Method & \multicolumn{2}{c}{RefCOCO} & \multicolumn{2}{c}{RefCOCO+} & \multicolumn{2}{c}{RefCOCOg} \\ 
\cmidrule(lr){3-4} \cmidrule(lr){5-6} \cmidrule(lr){7-8}
  &        & $\rm P@0.5$ & $\rm oIoU$ & $\rm P@0.5$ & $\rm oIoU$ & $\rm P@0.5$ & $\rm oIoU$ \\ 
\midrule
\multirow{3}{*}{Specialist} 
& MDETR~\cite{kamath2021mdetr} & 87.5 & - & 81.1 & - & 83.4 & - \\
& SeqTR~\cite{zhu2022seqtr} & 87.0 & 71.7 & 78.7 & 63.0 & 82.7 & 64.7 \\
& PolyFormer (L)~\cite{liu2023polyformer} & 90.4 & 76.9 & 85.0 & 72.2 & 85.8 & 71.2 \\
\midrule
\multirow{4}{*}{Generalist}
& UniTAB (B)~\cite{yang2022unitab} & 88.6 & - & 81.0 & - & 84.6 & - \\ 
& OFA (L)~\cite{wang2022ofa} & 90.1 & - & 85.8 & - & 85.9 & - \\
& UNINEXT (L)~\cite{lin2023uninext} & 91.4 & 80.3 & 83.1 & 70.0 & 86.9 & 73.4 \\
& UNINEXT (H)~\cite{lin2023uninext} & 92.6 & 82.2 & 85.2 & 72.5 & 88.7 & 74.7 \\
\midrule
\multirow{2}{*}{Foundation}
& GLEE-Plus~\cite{wu2024general} & 90.6 & 79.5 & 81.6 & 68.3 & 85.0 & 70.6 \\
\cmidrule{2-8}
& GLEE-Lite~\cite{wu2024general} & 88.5 & 77.4 & 78.3 & 64.8 & 82.9 & 68.8 \\
% & \methodNAME-3 & 87.5 & 75.6 & 77.2 & 63.1 & 80.4 & 65.6 \\
% & \methodNAME-2 & 86.1 & 73.6 & 75.7 & 61.0 & 79.2 & 63.5 \\
% \cmidrule{2-8}
% & \methodNAME-1 & 85.8 & 73.2 & 73.1 & 59.0 & 77.4 & 62.1 \\
% & \methodNAME-0 & 85.1 & 71.9 & 73.4 & 58.5 & 77.4 & 61.5 \\
% & \methodNAME-R50 & 86.6 & 74.3 & 76.0 & 61.5 & 79.5 & 64.2 \\
& \methodNAME-3 & 87.5                                          & 75.4                                          & 76.8                                              & 62.8                                              & 80.1                                           & 65.5                                           \\
& \methodNAME-2 & 86.6                                          & 74.2                                          & 74.9                                              & 60.3                                              & 78.3                                           & 63.0                                           \\
\cmidrule{2-8}
& \methodNAME-1 & 86.3                                          & 73.9                                          & 74.4                                              & 59.7                                              & 77.5                                           & 61.4                                           \\
& \methodNAME-0 & 85.7                                          & 72.7                                          & 73.5                                              & 59.1                                              & 77.3                                           & 61.3                                           \\
& \methodNAME-R50 & 86.9                                          & 74.8                                          & 75.2                                              & 61.6                                              & 79.2                                           & 64.0            \\ 
\bottomrule
\end{tabular}
}
\caption{Comparison of methods on RefCOCO, RefCOCO+, and RefCOCOg datasets.}
\label{tab:refcoco_results}
\end{table}

%% file: tables/seginw.tex
\setlength{\tabcolsep}{2pt}
\begin{table*}[t]
\begin{center}
\scriptsize
\resizebox{\linewidth}{!}{
\begin{tabular}{llccccccccccccccccccccccc}
\toprule
Method & \rotatebox{90}{Brain Tumor} & \rotatebox{90}{Chicken} & \rotatebox{90}{Cows} & \rotatebox{90}{Electric Shaver} & \rotatebox{90}{Elephants} & \rotatebox{90}{Fruits} & \rotatebox{90}{Garbage} & \rotatebox{90}{Ginger Garlic} & \rotatebox{90}{Hand} & \rotatebox{90}{Hand Metal} & \rotatebox{90}{HouseHold Items} & \rotatebox{90}{NutterflySquirrel} & \rotatebox{90}{Phones} & \rotatebox{90}{Poles} & \rotatebox{90}{Puppies} & \rotatebox{90}{Rail} &  \rotatebox{90}{Salmon Fillet} & \rotatebox{90}{Strawberry} & \rotatebox{90}{Tablets} & \rotatebox{90}{Toolkits} & \rotatebox{90}{Trash} & \rotatebox{90}{Watermelon} & Avg \\
\midrule
X-Decoder(L)~\cite{zou2023generalized} & 2.2 & 8.6 & 44.9 & 7.5 & 66.0 & 79.2 & 33.0 & 11.6 & 75.9 & 42.1 & 53.0 & 68.4 & 15.6 & 20.1 & 59.0 & 2.3 & 19.0 & 67.1 & 22.5 & 9.9 & 22.3 & 13.8 & 32.3 \\
OpenSEED(L)~\cite{zhang2023simple} & 2.1 & 82.9 & 40.9 & 4.7 & 72.9 & 76.4 & 16.9 & 13.6 & 92.7 & 38.7 & 50.0 & 40.0 & 7.6 & 4.6 & 74.6 & 1.8 & 15.6 & 82.8 & 47.4 & 15.4 & 15.3 & 52.3 & 36.1 \\
ODISE(L)~\cite{xu2023open} & 2.9 & 84.1 & 41.6 & 18.3 & 74.9 & 81.3 & 39.8 & 23.0 & 41.4 & 51.4 & 60.4 & 71.9 & 43.8 & 0.4 & 65.4 & 2.8 & 30.2 & 79.9 & 9.1 & 15.0 & 28.6 & 37.5 & 38.7 \\
SAN(L)~\cite{xu2023side} & 2.6 & 69.2 & 44.0 & 11.4 & 67.4 & 77.4 & 46.5 & 23.3 & 88.8 & 62.9 & 60.1 & 82.2 & 10.4 & 1.8 & 60.1 & 2.9 & 20.0 & 81.8 & 35.1 & 31.2 & 41.4 & 43.5 & 41.4 \\
HIPIE(H)~\cite{wang2024hierarchical} & 1.9 & 46.5 & 50.1 & 76.1 & 68.6 & 61.1 & 31.2 & 24.3 & 94.2 & 64.0 & 53.4 & 79.7 & 7.0 & 6.7 & 64.6 & 2.2 & 41.8 & 81.5 & 8.8 & 17.9 & 31.2 & 50.6 & 41.2 \\
UNINEXT(L)~\cite{UNINEXT} & 2.6 & 75.2 & 52.1 & 71.2 & 72.1 & 81.1 & 16.9 & 23.7 & 93.7 & 57.0 & 54.0 & 84.1 & 6.1 & 13.4 & 64.6 & 0.0 & 44.4 & 80.7 & 21.0 & 10.1 & 10.8 & 56.3 & 42.1 \\
% PartGLEE (R50)~\cite{partglee} & 7.1 & 79.4 & 38.1 & 6.9 & 74.7 & 81.1 & 27.2 & 25.7 & 87.6 & 66.5 & 60.1 & 71.2 & 47.4 & 25.7 & 67.4 & 4.7 & 32.3 & 80.3 & 32.8 & 10.9 & 22.2 & 62.3 & 44.1 \\
% PartGLEE (L)~\cite{partglee} & 20.7 & 77.7 & 48.0 & 18.6 & 77.3 & 82.4 & 31.6 & 23.7 & 82.0 & 55.3 & 52.0 & 84.9 & 17.3 & 23.3 & 63.9 & 20.0 & 37.4 & 80.6 & 6.6 & 6.7 & 24.7 & 68.2 & 44.2 \\
\midrule
\methodNAME-3 & 4.4 & 80.5 & 42.7 & 0.7  & 77.8 & 82.3 & 17.1 & 50.2 & 77.4 & 92.0 & 53.4 & 82.4 & 42.1 & 22.1 & 63.5 & 10.6 & 26.1 & 83.1 & 4.7 & 19.2 & 39.4 & 68.9 & \textbf{47.3} \\
\methodNAME-2 & 5.0 & 79.8 & 29.2 & 35.5 & 76.7 & 80.7 & 22.5 & 48.0 & 80.1 & 47.3 & 25.9 & 79.3 & 20.5 & 23.6 & 63.0 & 13.8 & 16.3 & 85.1 & 0.5 & 14.1 & 27.1 & 61.8 & 42.5 \\
\midrule
\methodNAME-1 & 4.7 & 75.1 & 18.8 & 9.7  & 76.8 & 80.4 & 21.5 & 50.7 & 78.0 & 67.5 & 52.5 & 76.5 & 42.6 & 21.3 & 63.6 & 7.0  & 38.0 & 88.1 & 1.2 & 15.1 & 18.7 & 63.0 & \textbf{44.1} \\
\methodNAME-0 & 6.9 & 80.8 & 18.5 & 0.7  & 75.4 & 82.2 & 13.4 & 48.8 & 79.4 & 77.8 & 27.5 & 73.8 & 27.1 & 10.9 & 65.3 & 9.0  & 29.5 & 88.1 & 0.5 & 10.9 & 30.5 & 66.2 & 42.0 \\
% \methodNAME-3 & 15.4 & 75.5 & 40.5 & 14.1 & 78.1 & 82.2 & 15.9 & 40.5 & 32.5 & 29.6 & 51.8 & 83.0 & 26.2 & 6.2 & 67.1 & 2.2 & 26.5 & 78.9 & 4.3 & 9.6 & 32.0 & 64.9 & 39.9 \\
% \methodNAME-2 & 1.0 & 65.8 & 23.8 & 0.4 & 75.7 & 74.3 & 16.1 & 40.7 & 94.9 & 73.4 & 28.4 & 77.5 & 54.8 & 12.6 & 68.3 & 12.9 & 31.8 & 83.8 & 1.7 & 25.0 & 29.9 & 44.8 & \textbf{42.6} \\
% \midrule
% \methodNAME-1 & 3.1 & 63.2 & 23.3 & 0.9 & 72.6 & 84.0 & 17.0 & 39.9 & 94.3 & 58.7 & 31.7 & 73.9 & 53.5 & 23.7 & 62.8 & 6.4 & 19.0 & 82.8 & 1.5 & 18.7 & 33.9 & 57.6 & 41.9 \\
% \methodNAME-0 & 4.4 & 79.4 & 29.9 & 0.1 & 76.2 & 77.7 & 14.7 & 33.4 & 92.2 & 72.7 & 54.0 & 70.2 & 40.5 & 22.2 & 70.0 & 7.4 & 24.0 & 88.3 & 0.3 & 15.6 & 29.1 & 62.8 & \textbf{43.9} \\

\bottomrule
\end{tabular}}
\end{center}
\caption{Results on SeginW benchmark across 22 datasets. We report the AP mask.}
\label{table:seginw_results}
\end{table*}

%% file: tables/odinw.tex
\begin{table}[t!]

\begin{center}
\resizebox{\linewidth}{!}{
% \begin{tabular}{l@{\hskip9pt}| 
% l@{\hskip9pt}l@{\hskip9pt}l@{\hskip9pt} 
% l@{\hskip9pt}l@{\hskip9pt}l@{\hskip9pt}
% l@{\hskip9pt}l@{\hskip9pt}l@{\hskip9pt}l@{\hskip9pt}
% l@{\hskip9pt}l@{\hskip9pt}l@{\hskip9pt}l@{\hskip9pt}l@{\hskip9pt}l@{\hskip9pt}l}
\centering
\scriptsize
\setlength{\tabcolsep}{2pt}
\newcommand{\band}{\rowcolor{gray!10}}
\begin{tabular}{lcccccccccccccc}
\toprule

Model  & \rotatebox{90}{PascalVOC} &
\rotatebox{90}{AerialDrone} & 
\rotatebox{90}{Aquarium} &
\rotatebox{90}{Rabbits} &
\rotatebox{90}{EgoHands} &
\rotatebox{90}{Mushrooms} &
\rotatebox{90}{Packages} &
\rotatebox{90}{Raccoon} &
\rotatebox{90}{Shellfish} &
\rotatebox{90}{Vehicles} &
\rotatebox{90}{Pistols} &
\rotatebox{90}{Pothole} &
\rotatebox{90}{Thermal} & 
Avg
\\
\midrule

 GLIP-T~\cite{zhang2022glipv2} % 69.4
 & 56.2
& 12.5
& 18.4
& 70.2
& 50.0
& 73.8
& 72.3
& 57.8
& 26.3
& 56.0
& 49.6
& 17.7
& 44.1
  & 46.5
  
\\

 GLIP-L~\cite{zhang2022glipv2}  % 74.0
 & 61.7
& 7.1
& 26.9
& 75.0
& 45.5
& 49.0
& 62.8
& 63.3
& 68.9
& 57.3
& 68.6
& 25.7
& 66.0
& 52.1
\\

GLEE-Plus~\cite{wu2024general}
& 67.8 
& 10.8 
& 38.3 
& 76.1 
& 47.4 
& 19.2 
& 29.4 
& 63.8 
& 66.7 
& 63.8 
& 62.6 
& 15.3 
& 66.5 
& 48.3 
\\
\midrule

GLEE-Lite~\cite{wu2024general}
& 61.7
& 7.9 
&23.2 
& 72.6 
& 41.9 
& 51.6 
& 32.9 
& 51.1 
& 35.0 
& 59.4 
& 45.6 
& 21.8 
& 56.9 
& 43.2
\\

\methodNAME-3 & 67.2 & 18.2 & 31.1 & 76.7 & 13.8 & 41.4 & 66.0 & 48.3 & 46.3 & 61.3 & 67.5 & 13.8 & 40.2 & \textbf{45.5} \\
\methodNAME-2 & 64.8 & 11.8 & 28.1 & 77.5 & 19.2 & 38.9 & 52.3 & 57.7 & 46.3 & 61.3 & 62.2 & 13.2 & 36.6 & 43.8 \\
\midrule
\methodNAME-1 & 64.8 & 13.5 & 29.4 & 76.2 & 16.6 & 19.0 & 59.8 & 50.6 & 43.7 & 59.5 & 60.4 & 14.3 & 38.2 & \textbf{42.0} \\
\methodNAME-0 & 64.5 & 16.0 & 26.5 & 78.7 & 12.5 & 18.8 & 43.8 & 55.4 & 37.0 & 58.0 & 59.3 & 17.2 & 37.0 & 41.2 \\

\bottomrule
\end{tabular}
}
\caption{Zero-shot performance on 13 ODinW datasets.}
\label{table:odinw_zero_shot}
\end{center}
\end{table}

%% file: new_tables/component-efficiency.tex
\begin{table}[!ht]
\centering
\scriptsize
\setlength{\tabcolsep}{2pt}
\renewcommand{\arraystretch}{1.2} % Adjust row height (optional)
\resizebox{1.0\linewidth}{!}{
\begin{tabular}{llccccc} 
\toprule
\multirow{3}{*}{Method}
& \multirow{3}{*}{\makecell{Pix. Dec.\\Type}} & \multicolumn{5}{c}{FLOPs (G)} \\ 
\cmidrule(lr){3-7}
&  & Vis. Enc. & Pix. Dec. & (+Modality Fusion) & Decoder & Total \\ 
\midrule
GLEE$^{\dagger}$~\cite{glee} & MaskDINO~\cite{li2023mask} &  52.4 & 138  & 28.2 & 20.1 & 238.9    \\
GLEE$^{\dagger}$~\cite{glee} & RT-DETR~\cite{zhao2024detrs} &  52.4 &  69.2  & \textbf{1.6} & 20.0 &   143.1  \\
MOBIUS (Ours) & Bottleneck &  52.4 & \textbf{61.4}   & 5.6 & \textbf{10} &  \textbf{129.8}   \\
\bottomrule
\end{tabular}
}
\caption{\textbf{Component-wise Efficiency Analysis.} We compare the computational cost of MOBIUS and GLEE~\cite{glee} variants using MaskDINO~\cite{li2023mask} or RT-DETR~\cite{zhao2024detrs} decoders. FLOPs are reported for the vision encoder, pixel decoder, modality fusion, and decoder. All models use an R50 vision encoder at 800×800 resolution, excluding the text encoder from the total FLOPs count.}
\label{tab:component_efficiency}
\end{table}

%% file: tables/fastervit_mobilenet.tex
\begin{table}[!ht]
\centering
\scriptsize
\setlength{\tabcolsep}{2pt}
% \renewcommand{\arraystretch}{1.2} % Adjusts row height (optional)
% \resizebox{1.0\linewidth}{!}{
\begin{tabular}{llcccc} 
\toprule
& \multirow{2}{*}{Vision Encoder} 
& \multicolumn{2}{c}{\makecell{Vision Encoder \\ Efficiency}} 
& \multicolumn{2}{c}{COCO-val} \\
\cmidrule(lr){3-4} \cmidrule(lr){5-6}
& & FLOPs (G) & Latency (ms) & $\rm AP_{box}$  & $\rm AP_{mask}$ \\ 
\addlinespace[0.2em]
\midrule
\multirow{5}{*}{\rotatebox{90}{\textbf{MobileNetv4}}} 
& MobileNetv4-conv-small & 3 & 25.4 & 39.0 & 35.4 \\
& MobileNetv4-conv-medium & 15 & 39.0 & 43.6 & 39.2 \\
& MobileNetv4-conv-large & 38 & 48.4 & 47.2 & 42.3 \\
& MobileNetv4-hybrid-medium & 17 & 58.5 & 44.6 & 40.2 \\
& MobileNetv4-hybrid-large & 44 & 66.8 & 46.9 & 41.9 \\
\midrule
\multirow{4}{*}{\rotatebox{90}{\textbf{FasterViT}}}
& FasterViT-0 & 66 & 61.5 & 45.2 & 40.9 \\
& FasterViT-1 & 105 & 72.3 & 46.3 & 41.9 \\
& FasterViT-2 & 170 & 85.3 & 48.2 & 43.4 \\
& FasterViT-3 & 358 & 99.8 & 49.3 & 44.5 \\
\bottomrule
\end{tabular}
% }
\caption{\textbf{Mobile encoders comparison.} We compare the latency, FLOPs, and performance on COCO val of MOBIUS models trained on COCO following the 1x schedule using MobileNetv4 and FasterViT image encoders. We report Average Precision (AP) for box and mask predictions. The latency (in ms) is measured on one NVIDIA A100 with the images resized to 800 on their shorter side while preserving aspect ratio.}
\label{tab:mobilenet}
\end{table}

%% file: tables/lowres_flops.tex
\begin{table}[!ht]
\centering
\scriptsize
\setlength{\tabcolsep}{2pt}
\renewcommand{\arraystretch}{1.2} % Adjust row height (optional)
\resizebox{1.0\linewidth}{!}{
\begin{tabular}{lccccccc} 
\toprule
\multirow{3}{*}{Model} 
& \multicolumn{6}{c}{FLOPs (G)} \\ % Subheader for FLOPs
\cmidrule(lr){2-7}
& \textcolor{gray}{Text Encoder} & Vision Encoder & \multicolumn{2}{c}{Pixel Decoder} & Decoder & Total \\ 
\cmidrule(lr){4-5} \cmidrule(lr){7-7}
& & & w/o & w/ & & w/ \\ 
\midrule
GLEE-Plus~\cite{glee} & \textcolor{gray}{239} & 146 & \textcolor{gray}{49.6} & 59.5 & 9.9 & 454.4 \\
\midrule
GLEE-Lite~\cite{glee} & \textcolor{gray}{239} & \textbf{16.1} & \textcolor{gray}{50} & 59.9 & 9.9 & 324.9 \\
MOBIUS-3 & \textcolor{gray}{239} & 90.5 & \textcolor{gray}{19.8} & 24.7 & 4.9 & 354.2 \\
MOBIUS-2 & \textcolor{gray}{239} & 43.1 & \textcolor{gray}{19.7} & \textbf{24.6} & \textbf{4.9} & \textbf{311.6} \\
\midrule
MOBIUS-1 & \textcolor{gray}{239} & 29 & \textcolor{gray}{18.7} & 23.6 & 4.9 & 296.5 \\
MOBIUS-0 & \textcolor{gray}{239} & \textbf{16.7} & \textcolor{gray}{18.6} & \textbf{23.5} & \textbf{4.9} & \textbf{278.1} \\
\bottomrule
\end{tabular}
}
\caption{\textbf{Low-resolution FLOPs comparison.} We compare the FLOPs for each model component in GLEE and \methodNAME. Notice that the text encoder is a fixed cost that can be removed by caching in most applications. We report its cost for processing the $80$ COCO categories. We evaluate all models on low-resolution images rescaled to 384 on their short side while preserving aspect ratio. We compare the pixel decoder w/ and w/o early vision-language fusion.}
\label{tab:lowres_flops_per_component}
\end{table}

%% file: tables/bottleneck_decoder.tex
\begin{table}[!ht]
\centering
\scriptsize
\setlength{\tabcolsep}{4pt}
\renewcommand{\arraystretch}{1.2} % Adjusts row height (optional)
\resizebox{1.0\linewidth}{!}{
\begin{tabular}{lcccccc} 
\toprule
\multirow{2}{*}{\textbf{\makecell{Self-attn\\ Type}}} & \multirow{2}{*}{\textbf{\makecell{Bottleneck\\ Size}}} & \multirow{2}{*}{\textbf{Layers}} & \multirow{2}{*}{\textbf{Scales}} & \multirow{2}{*}{\textbf{FLOPs (G)}} & \multicolumn{2}{c}{\textbf{COCO-val}} \\
\cmidrule(lr){6-7} 
& & & & & $\rm AP_{box}$  & $\rm AP_{mask}$ \\
\midrule

\textbf{No} & 16 & 6 & Single &  410  & 44.0   & 39.8    \\
\textbf{Standard} & 16 & 6 & Single &   432  &  45.4  &  40.8   \\
\textbf{Deformable} & 16 & 6 & Single &   \textbf{413}  &  \textbf{45.5}  &  \textbf{41.1}   \\
\midrule
\multirow{3}{*}{\textbf{Deformable}}
 & 32 & 6 & Multi &  \textbf{399}  &  43.9  &  39.5   \\
 & 16 & 6 & Multi &  434   &  45.5  &  41.0   \\
 & 8 & 6 & Multi &  547   &  \textbf{45.7}  &  \textbf{41.2}   \\
\midrule
\multirow{2}{*}{\textbf{Deformable}}
 & 16 & 3 & Single &  \textbf{395}   &  44.2  & 39.9    \\
 & 16 & 6 & Single &  413  & \textbf{45.5}   & \textbf{41.1}    \\
\bottomrule
\end{tabular}
}
\caption{\textbf{Design Choices for Bottleneck Decoder.} FLOPs and performance (AP) are reported for COCO-val under different configurations: attention mechanisms (self, deformable, or no self-attention), bottleneck size (1/8, 1/16, 1/32), number of layers (3 or 6), scales (single or multi), and comparisons with/without multi-scale decoding.}
\label{tab:pixel_decoder_ablation}
\end{table}

%% file: tables/pruning_coco.tex
\begin{table}[!ht]
\centering
\scriptsize
\setlength{\tabcolsep}{2pt}
\renewcommand{\arraystretch}{1.2} % Adjusts row height (optional)
\resizebox{1.0\linewidth}{!}{
\begin{tabular}{lcllcccccc} 
\toprule
& \multirow{3}{*}{{Cal.}} 
& \multirow{3}{*}{{Strategy}} 
& \multirow{3}{*}{{Rule}} 
& \multirow{3}{*}{\rotatebox{90}{Lower}} 
& \multirow{3}{*}{\rotatebox{90}{Upper}} 
& \multirow{3}{*}{\rotatebox{90}{Min}} 
& \multirow{3}{*}{\rotatebox{90}{Layers}} 
& \multicolumn{2}{c}{COCO-val} \\
\cmidrule(lr){9-10} 
& & & & & & & & $\rm AP_{box}$  & $\rm AP_{mask}$ \\ 
\midrule
\addlinespace[0.3em]
\multirow{2}{*}{\rotatebox{90}{\textbf{COCO}}}
& - & Confidence & Sigmoid & 0.05 & 0.2 & 100 & 6 & 45.1 & 40.0 \\
& \checkmark & Confidence & Sigmoid & 0.05 & 0.2 & 100 & 6 & 46.0 & 41.1 \\
\addlinespace[0.2em]
\bottomrule
\end{tabular}
}
\caption{\textbf{Ablation Study of Query Pruning Strategy on COCO only.} Comparison of different pruning strategies across COCO with variations in calibration, selection strategy, rule type, threshold bounds, minimum kept elements, and decoder layers. We report FLOPs for the decoder and results on COCO-val.}
\label{tab:calibration_pruning_ablation_coco}
\end{table}

%% file: tables/pruning_trajectory.tex
\begin{table}[!ht]
\centering
\scriptsize
\setlength{\tabcolsep}{3pt}
\renewcommand{\arraystretch}{1.2}
\resizebox{\linewidth}{!}{
\begin{tabular}{lcccccc} 
\toprule
\multirow{2}{*}{Strategy} & \multirow{2}{*}{Rule} & \multirow{2}{*}{FLOPs} & \multicolumn{2}{c}{COCO-val} & \multicolumn{2}{c}{LVIS-minival} \\
\cmidrule(lr){4-5} \cmidrule(lr){6-7}
& & & $\rm AP_{box}$ & $\rm AP_{mask}$ & $\rm AP_{box}$ & $\rm AP_{mask}$ \\ 
\midrule
Confidence & Sigmoid & 4.6--7.6 & 52.2--52.7 & 46.2--46.7 & 47.6--47.9 & 44.0--44.5 \\
Confidence & Logarithm & 4.1--7.6 & 51.7--52.7 & 45.8--46.7 & 47.3--47.9 & 44.0--44.5 \\
Confidence & Exponential & 4.2--7.6 & 51.9--52.7 & 45.9--46.7 & 47.4--47.9 & 44.0--44.5 \\
\bottomrule
\end{tabular}
}
\caption{\textbf{Comparison of Sigmoid, Logarithm, and Exponential strategies.} Results show decoder FLOPs, $\rm AP_{box}$, and $\rm AP_{mask}$ on COCO-val and LVIS-minival. We report the range of results for different hyperparameter configurations.}
\label{tab:sigmoid_logarithm_exponential}
\end{table}

%% file: figures/qualitative.tex
\begin{figure*}[!ht]
    \centering
    \scriptsize
    \renewcommand{\arraystretch}{1.5} % Adjust row spacing for better visuals
    
    % Create a table for the images and text
    \begin{tabular}{cccc}
        \textbf{Raw Image} & \textbf{COCO Segmentation} & \textbf{Category-Agnostic} & \textbf{Referring Segmentation} \\
        
        % First row: Images
        \includegraphics[width=0.22\textwidth]{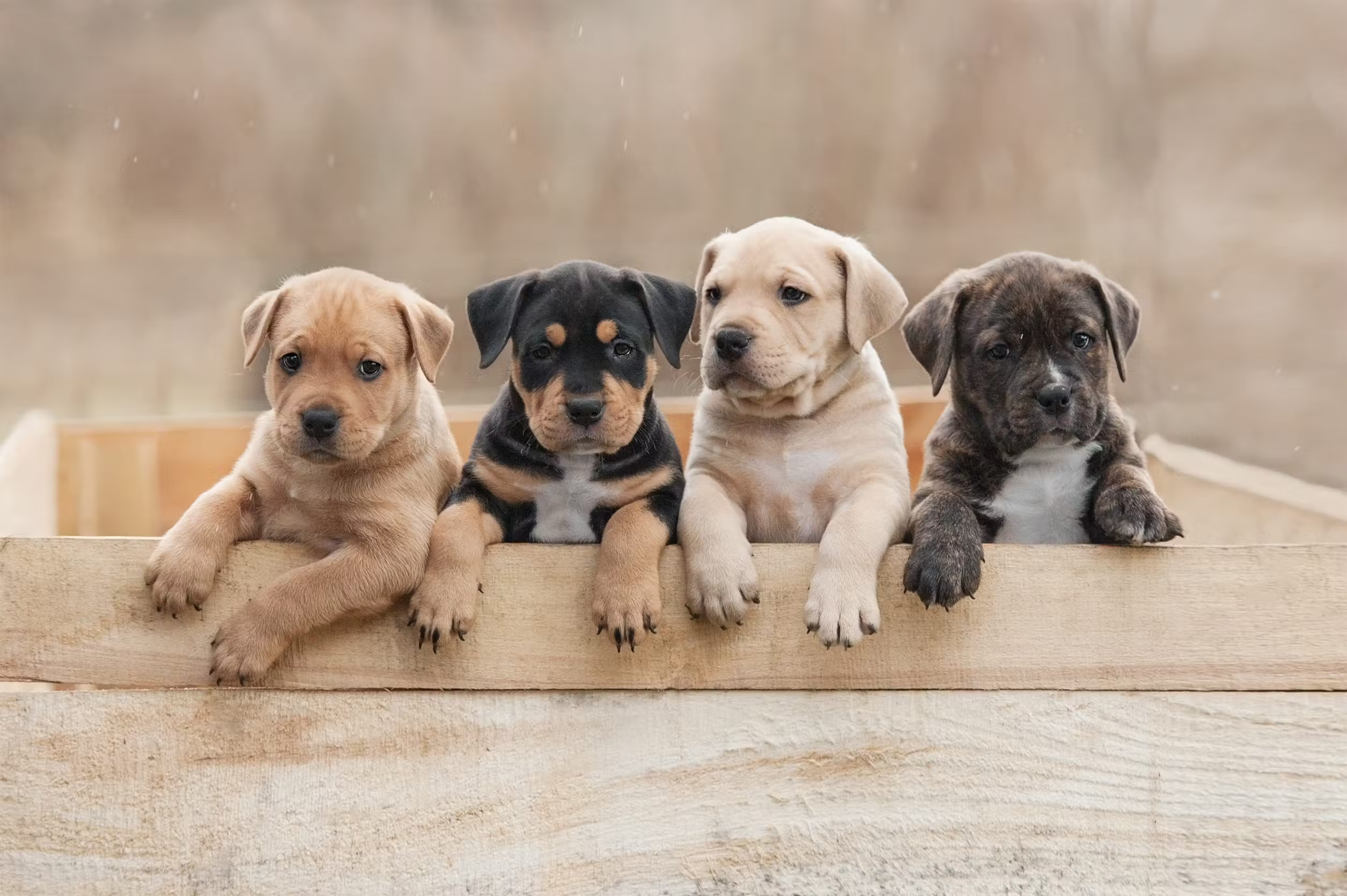} &
        \includegraphics[width=0.22\textwidth]{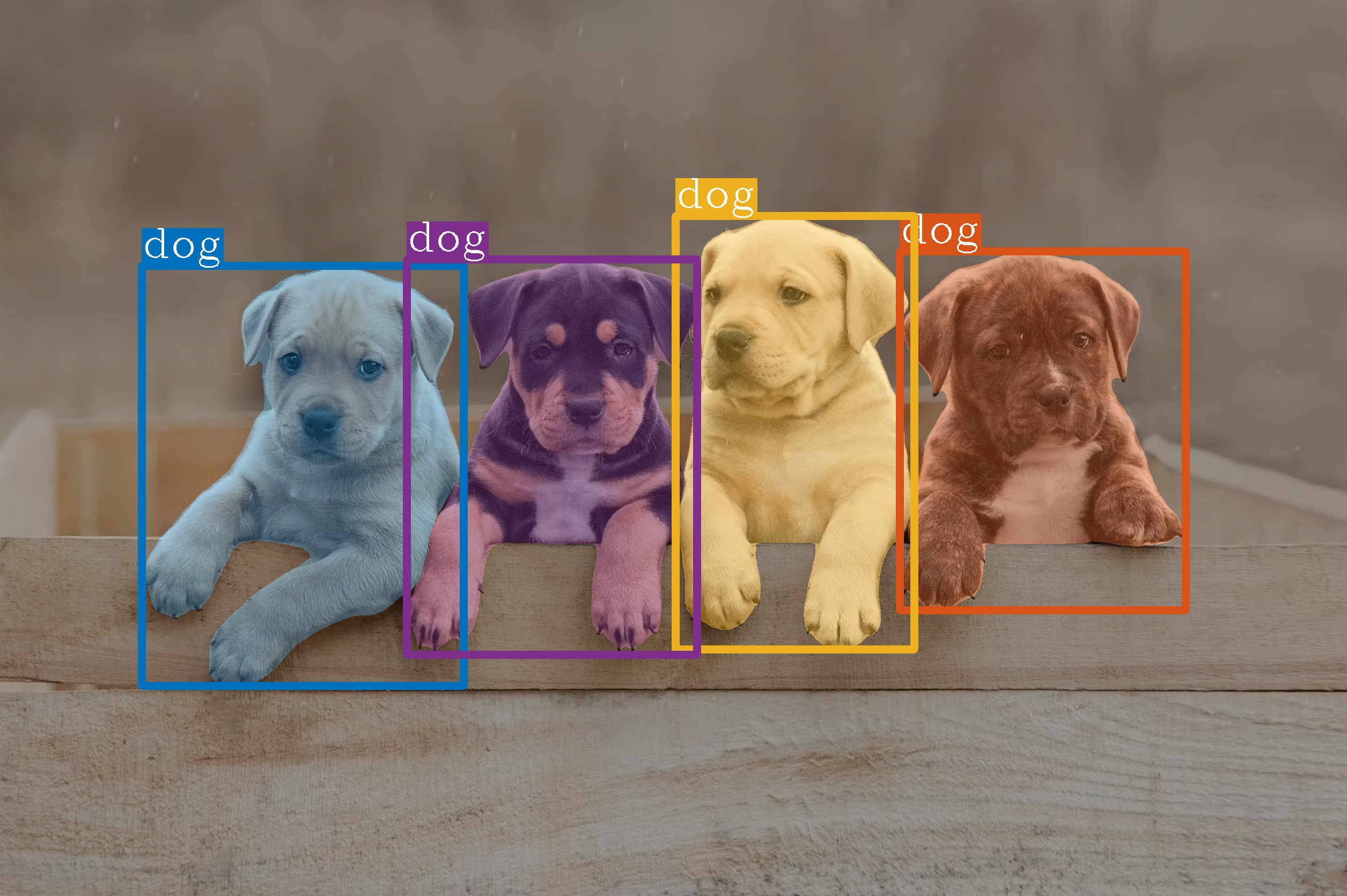} &
        \includegraphics[width=0.22\textwidth]{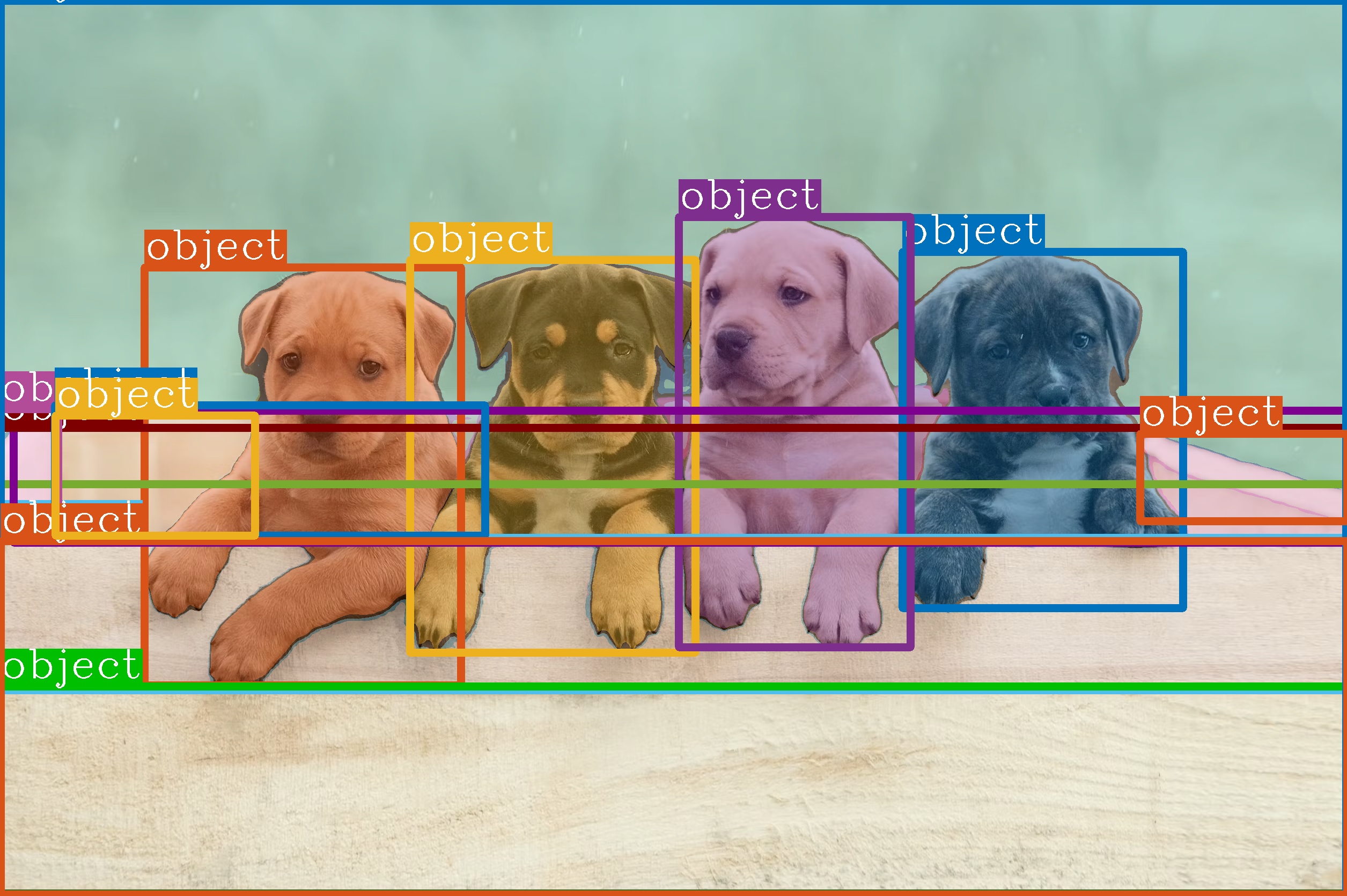} &
        \includegraphics[width=0.22\textwidth]{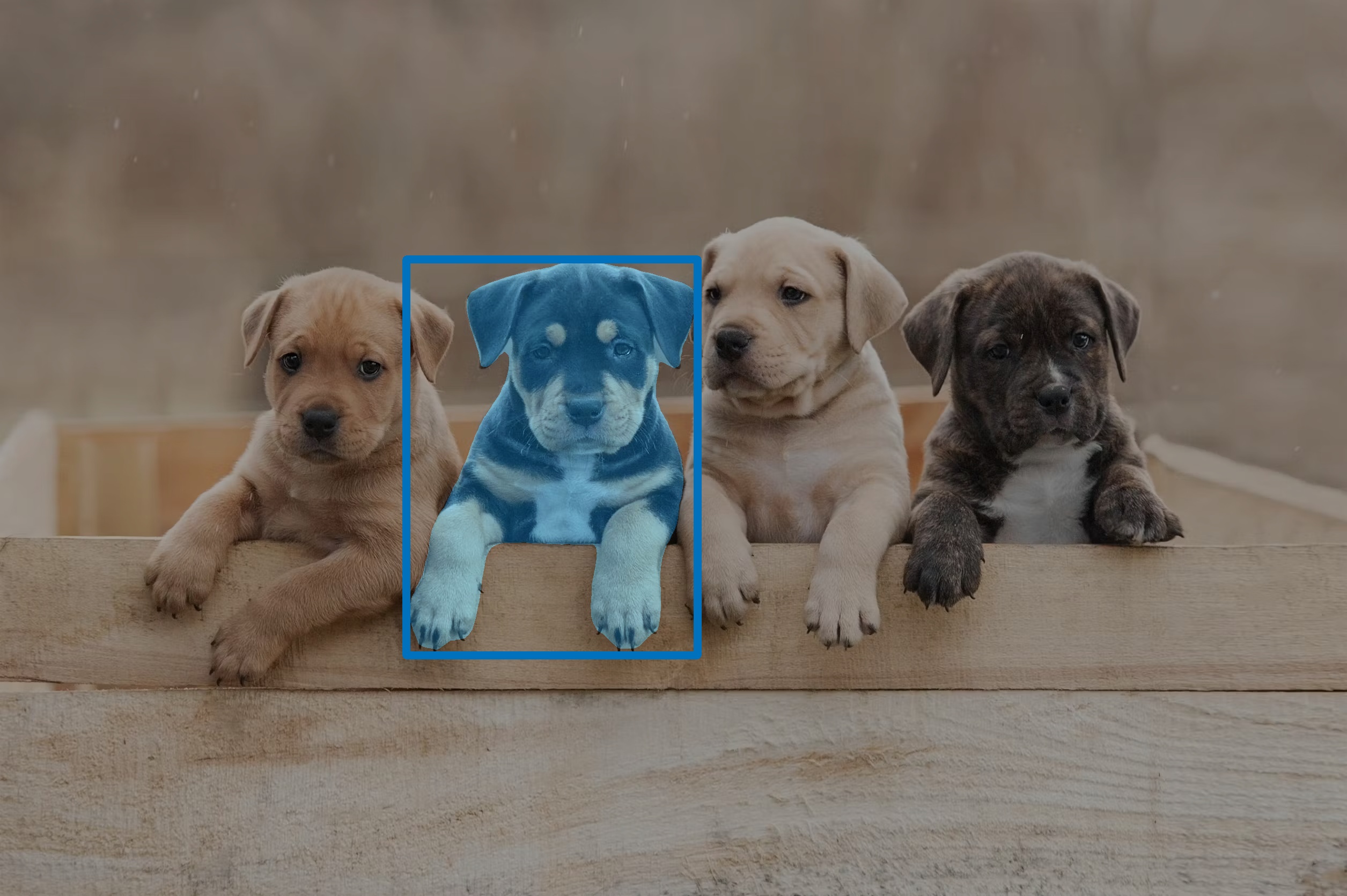} \\
        % Second row: Referring expression (only for the last column)
        & & & \textit{"the Rottweiler puppy"} \\

        % First row: Images
        \includegraphics[width=0.22\textwidth, trim={30px 0 30px 0}, clip]{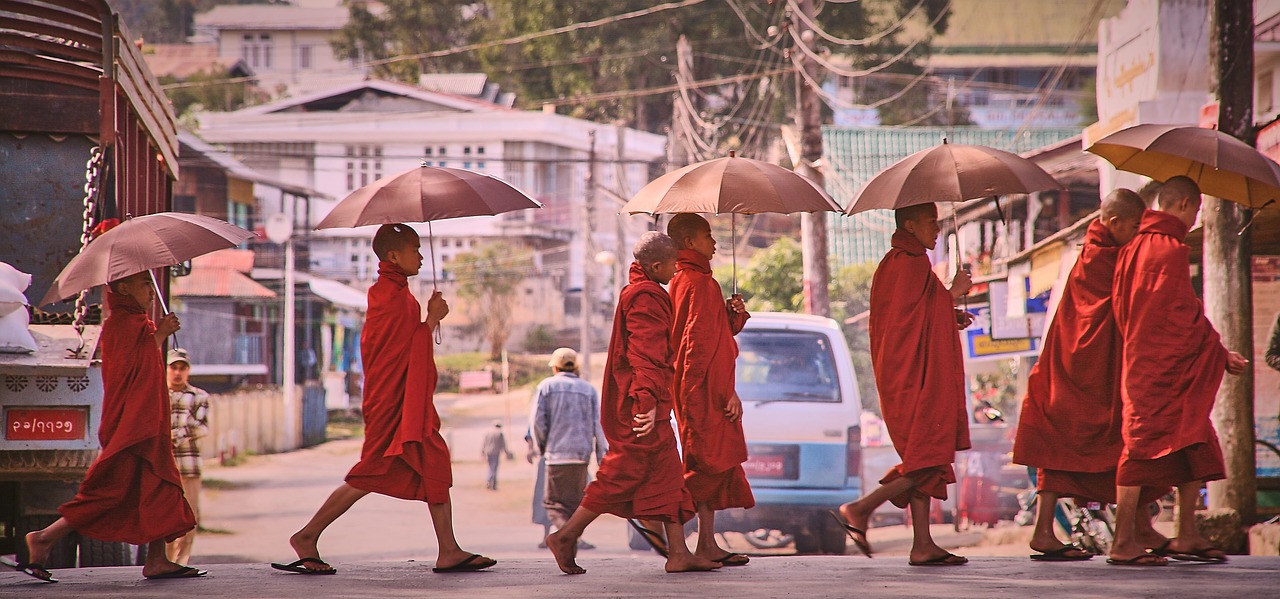} &
        \includegraphics[width=0.22\textwidth, trim={30px 0 30px 0}, clip]{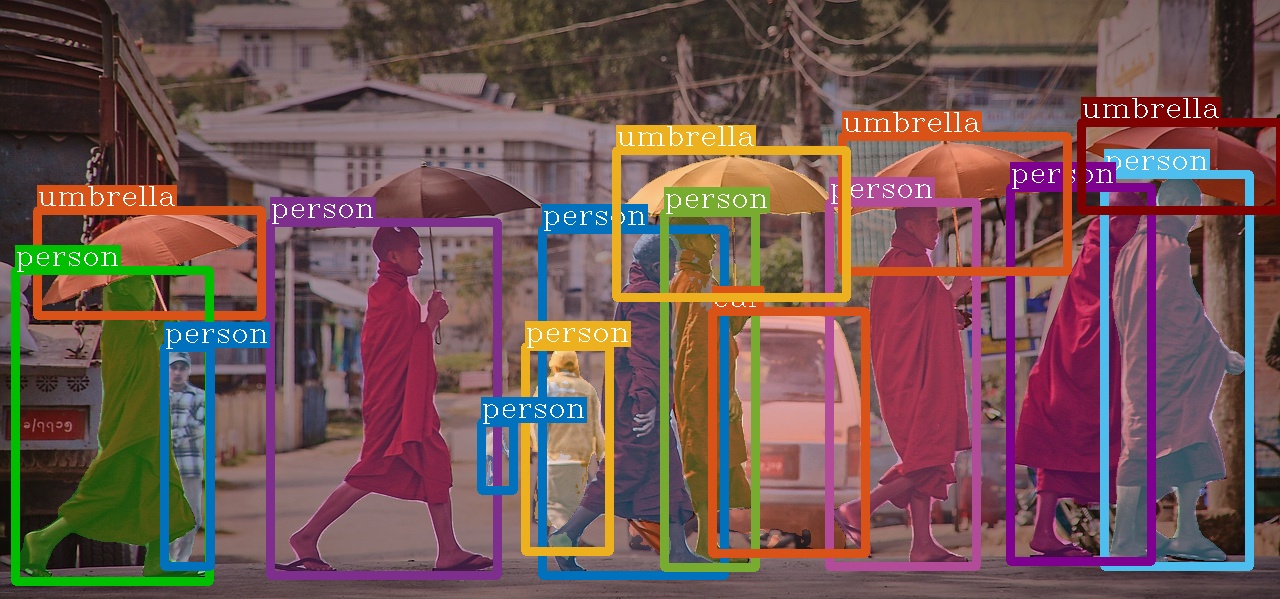} &
        \includegraphics[width=0.22\textwidth, trim={30px 0 30px 0}, clip]{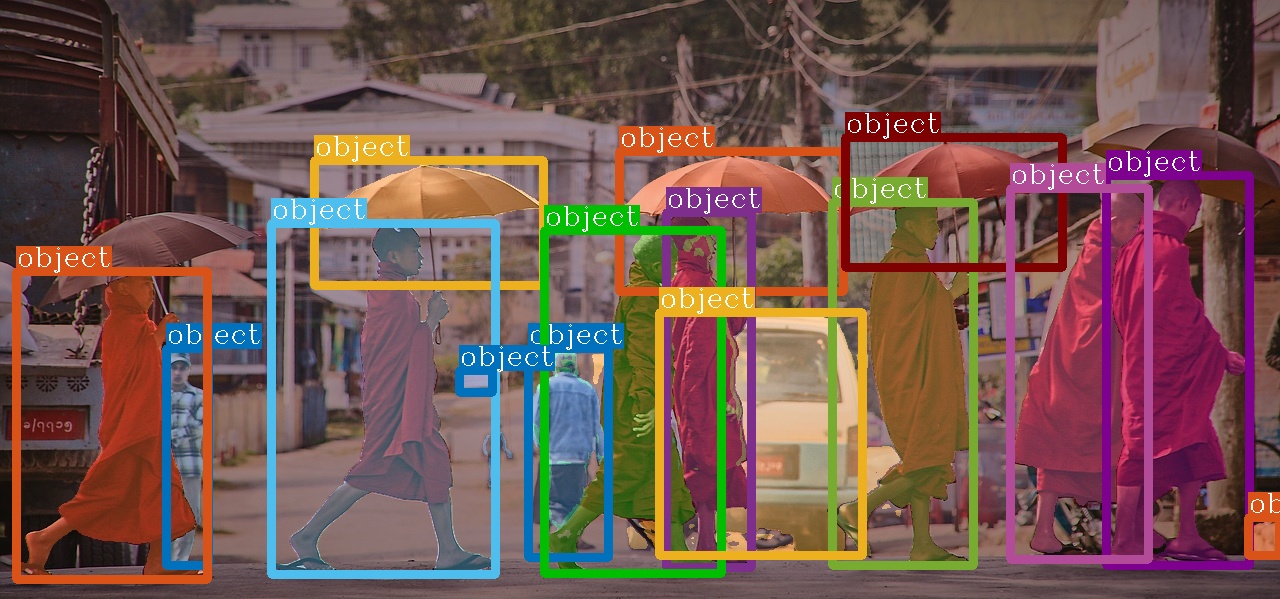} &
        \includegraphics[width=0.22\textwidth, trim={30px 0 80px 0}, clip]{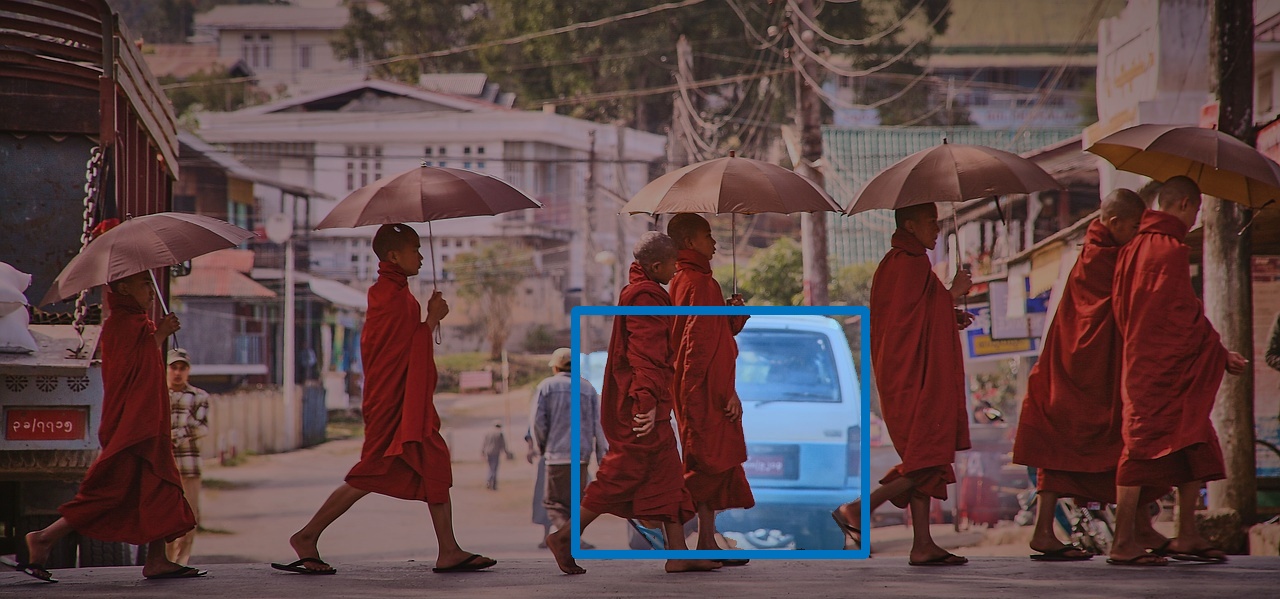} \\
        % Second row: Referring expression (only for the last column)
        & & & \textit{"the white and blue van"} \\
        
        % Row: Racecars (no trimming)
        \includegraphics[width=0.22\textwidth]{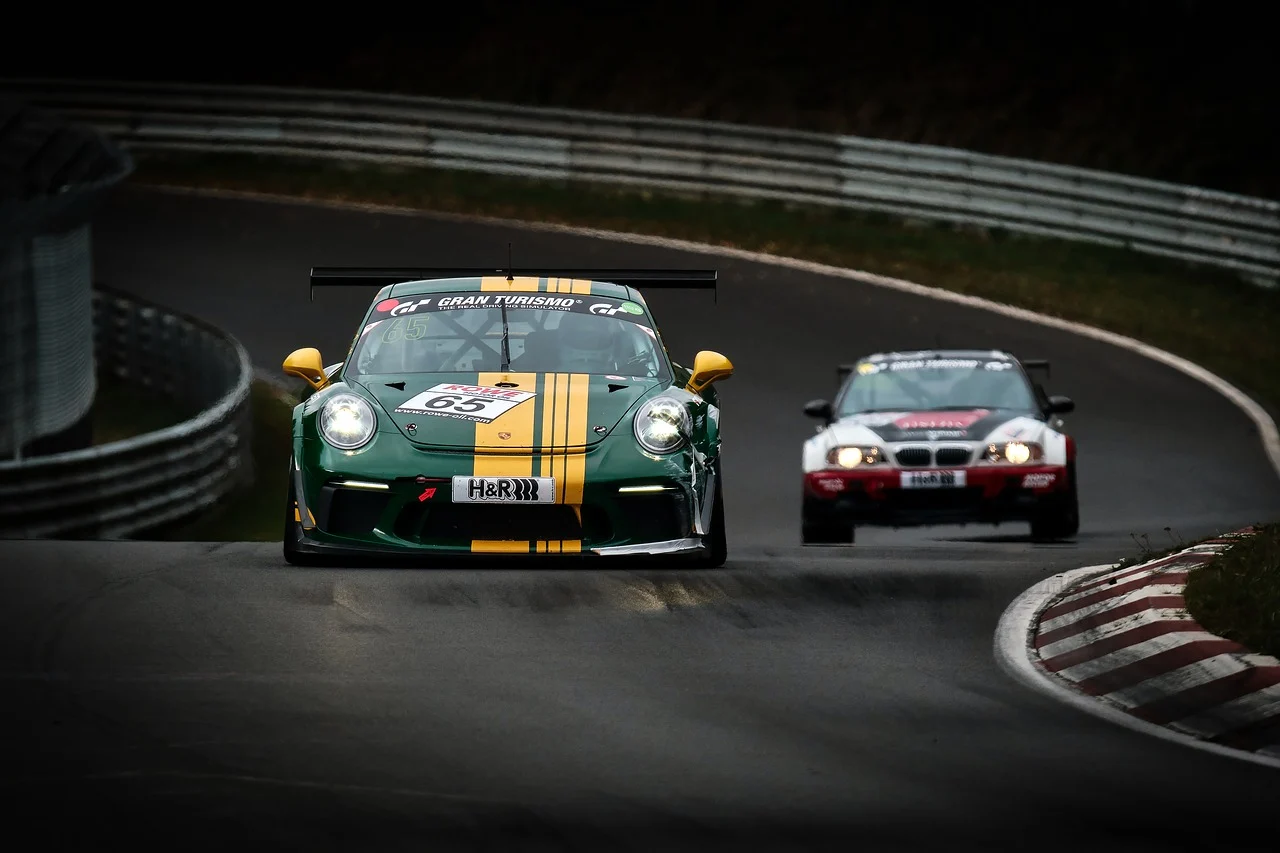} &
        \includegraphics[width=0.22\textwidth]{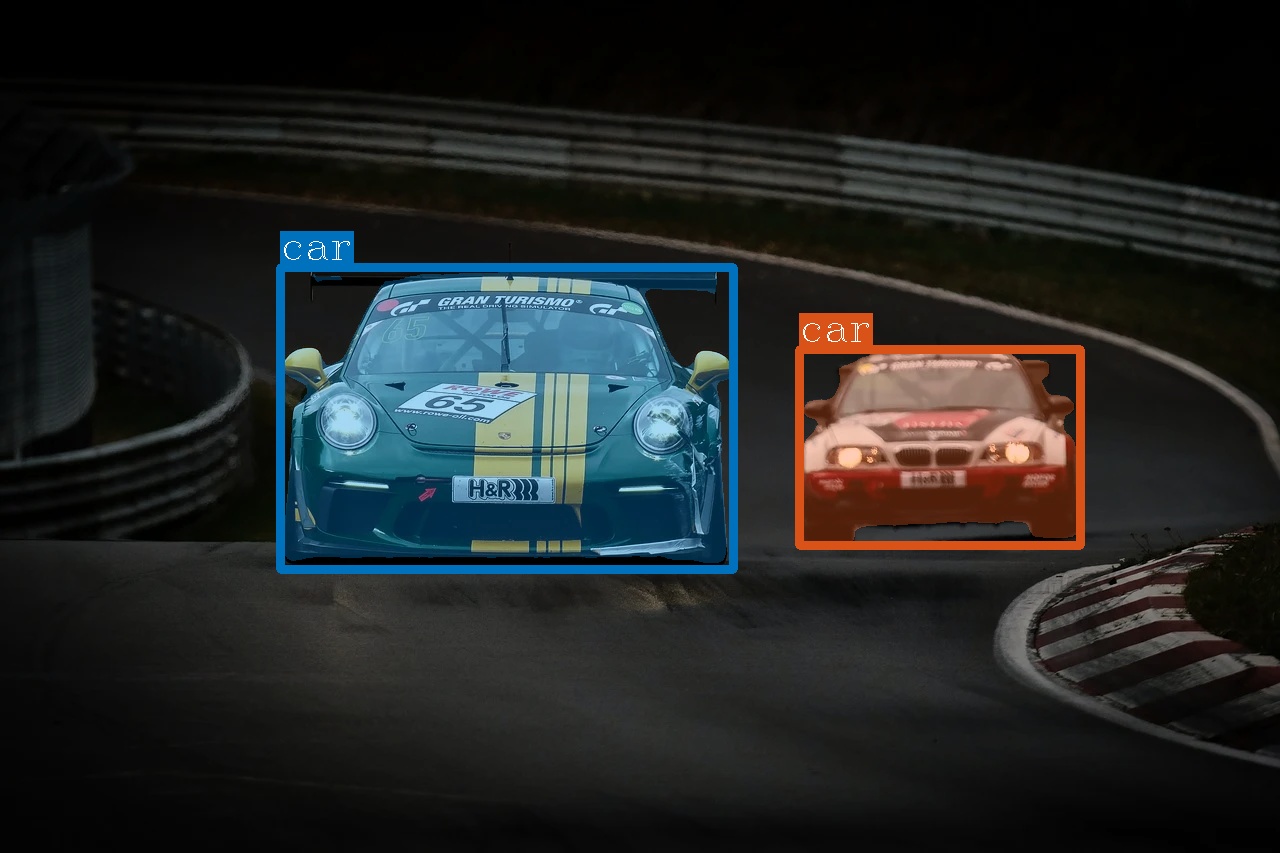} &
        \includegraphics[width=0.22\textwidth]{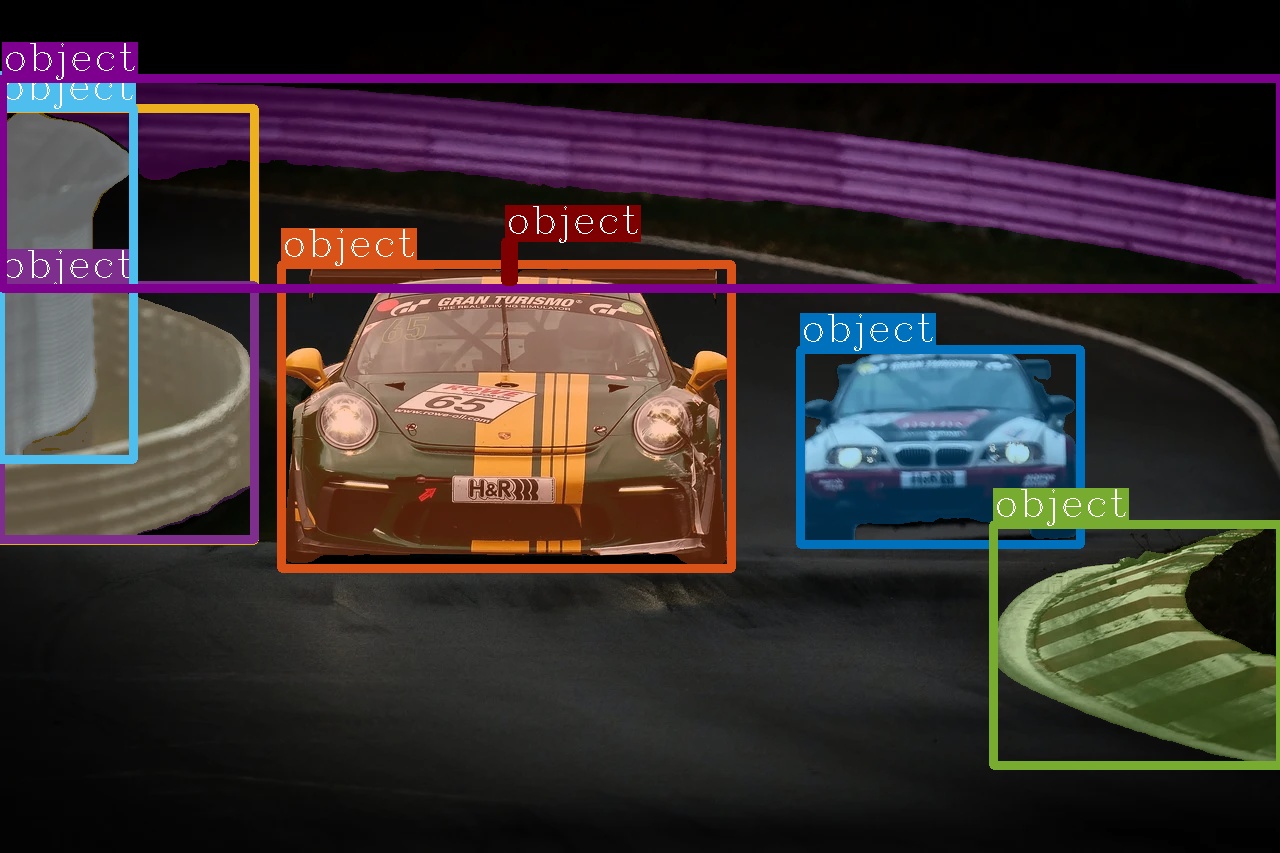} &
        \includegraphics[width=0.22\textwidth]{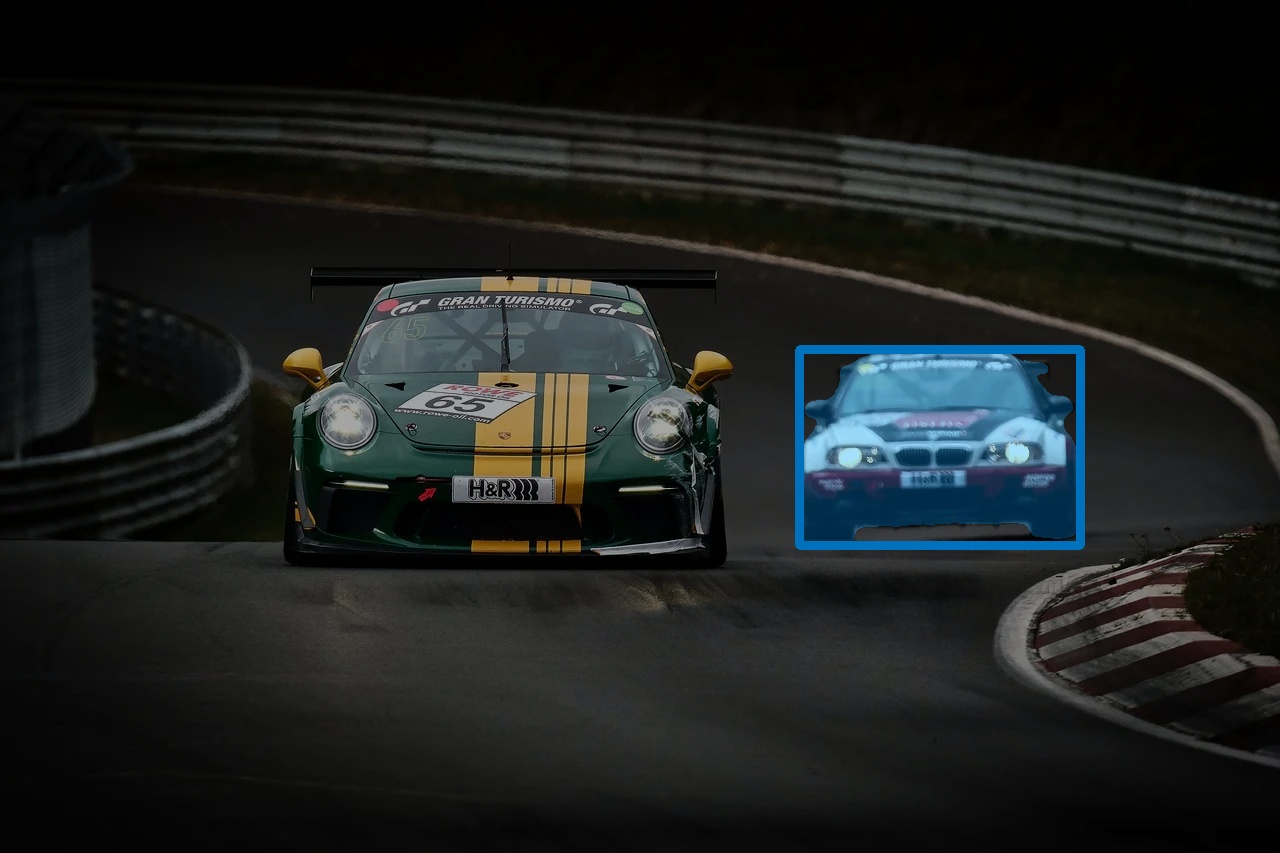} \\
        % Referring expression
        & & & \textit{"the race car behind"} \\
        
        % Row: Lavender (no trimming)
        \includegraphics[width=0.22\textwidth]{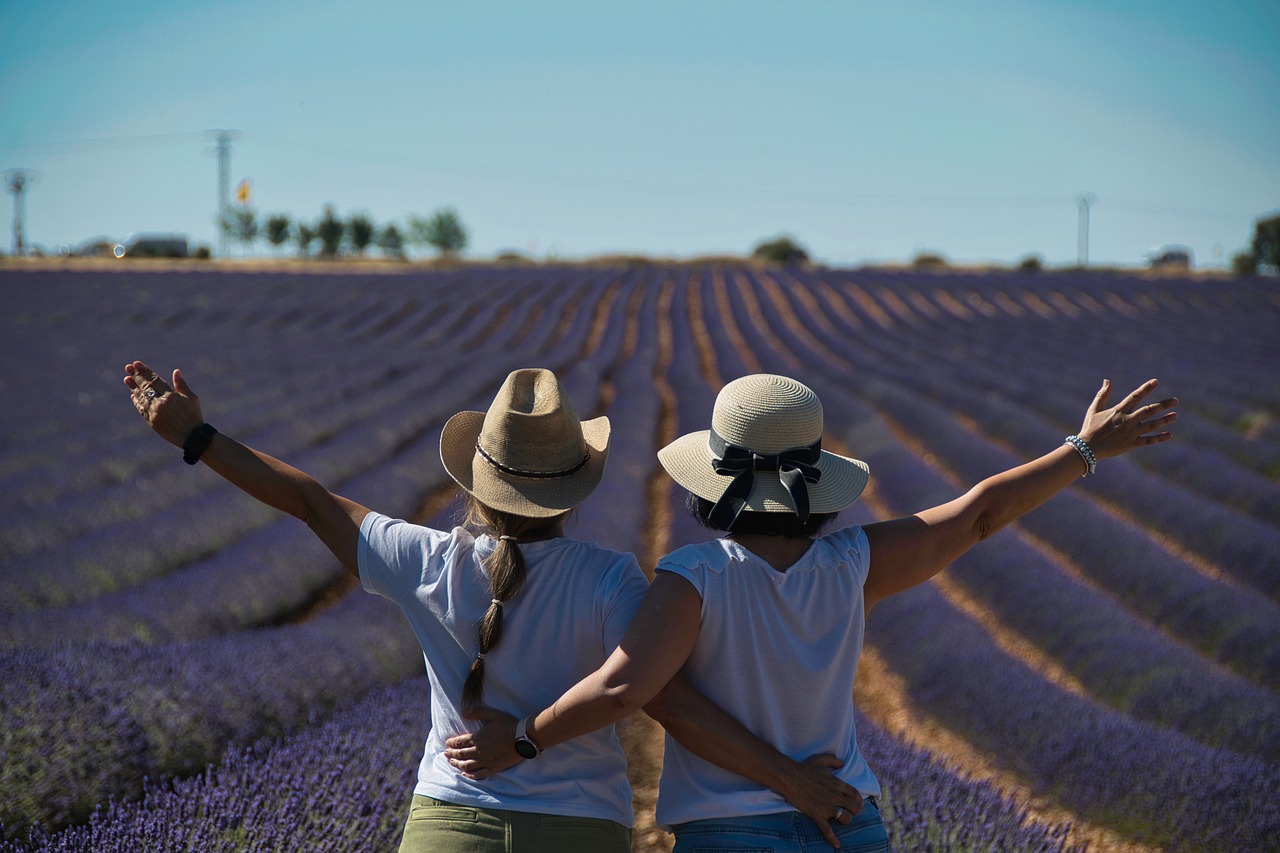} &
        \includegraphics[width=0.22\textwidth]{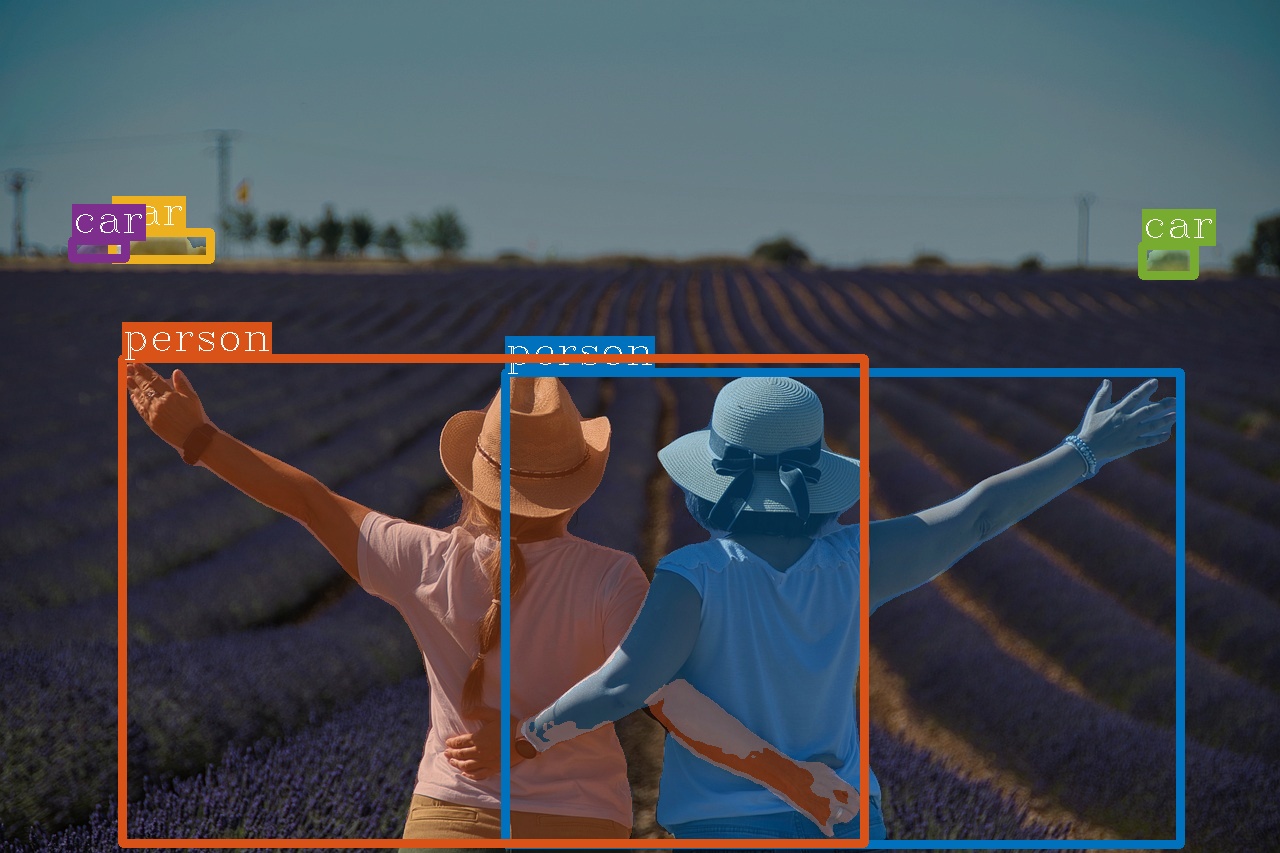} &
        \includegraphics[width=0.22\textwidth]{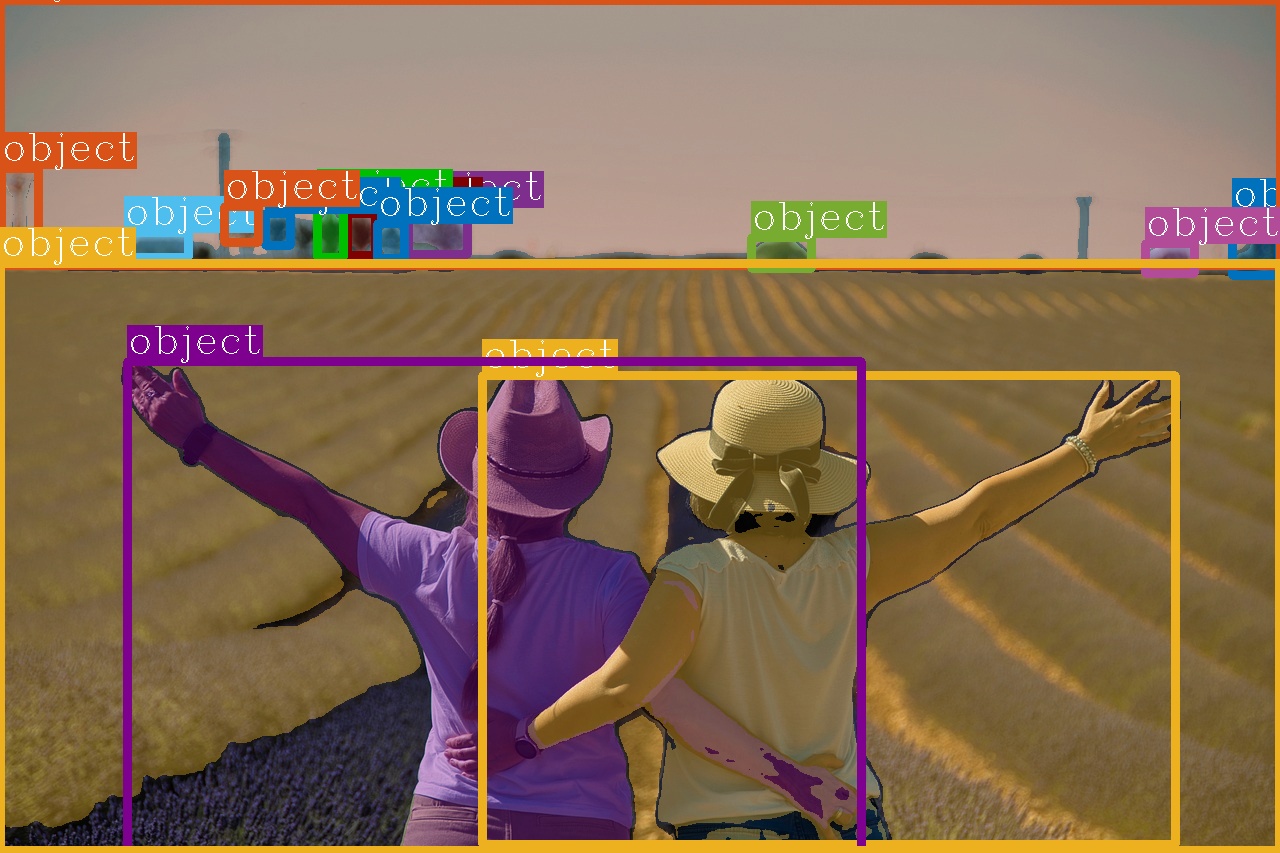} &
        \includegraphics[width=0.22\textwidth]{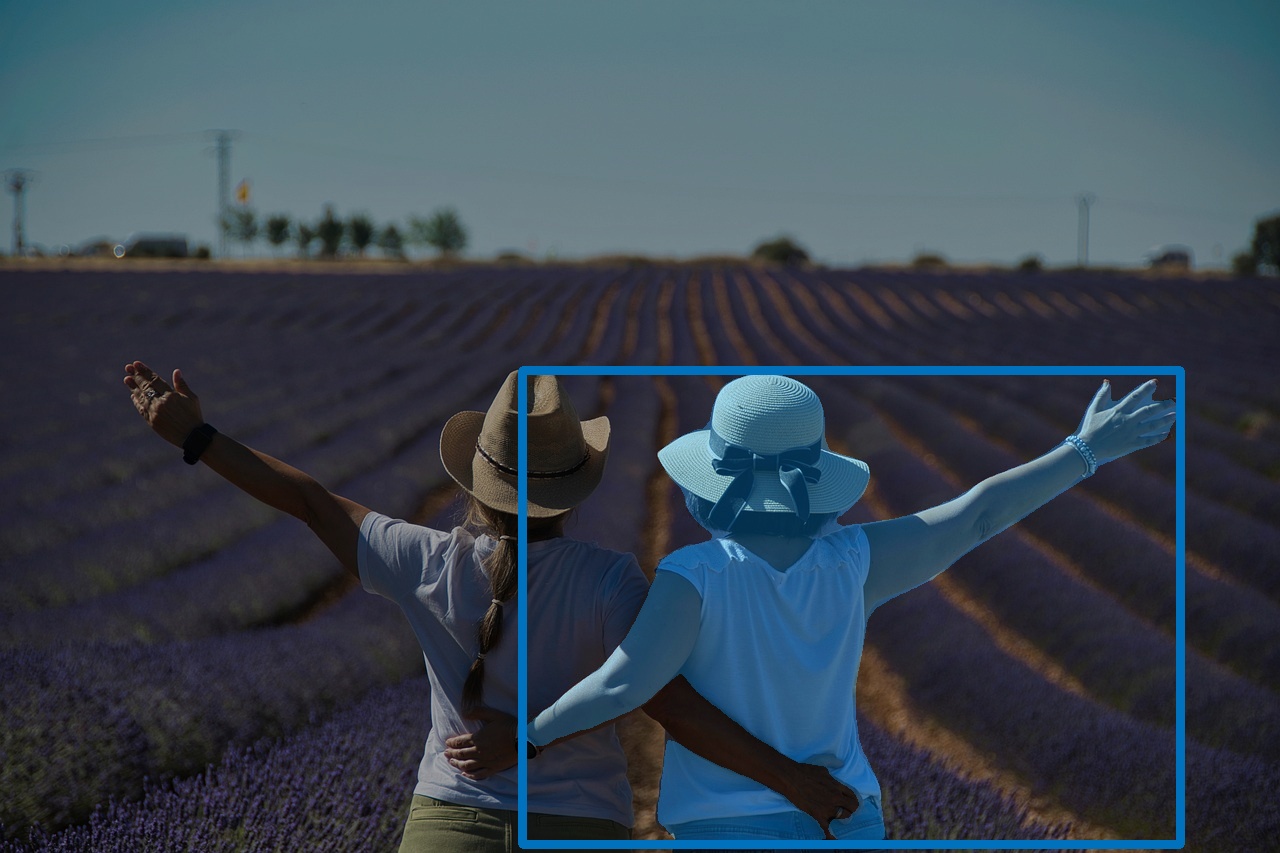} \\
        % Referring expression
        & & & \textit{"the girl wearing a hat with a ribbon"} \\
        
        % Row: Elephants (no trimming)
        \includegraphics[width=0.22\textwidth]{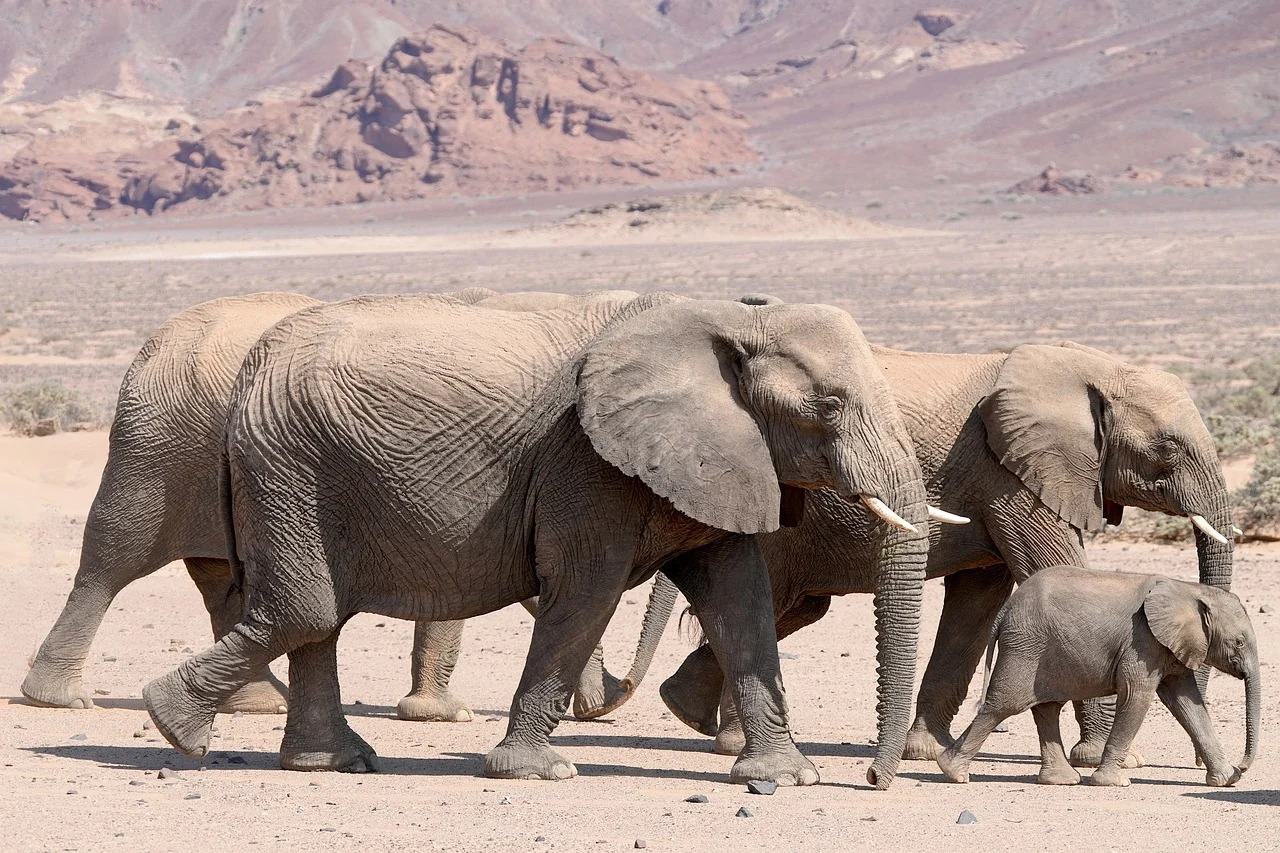} &
        \includegraphics[width=0.22\textwidth]{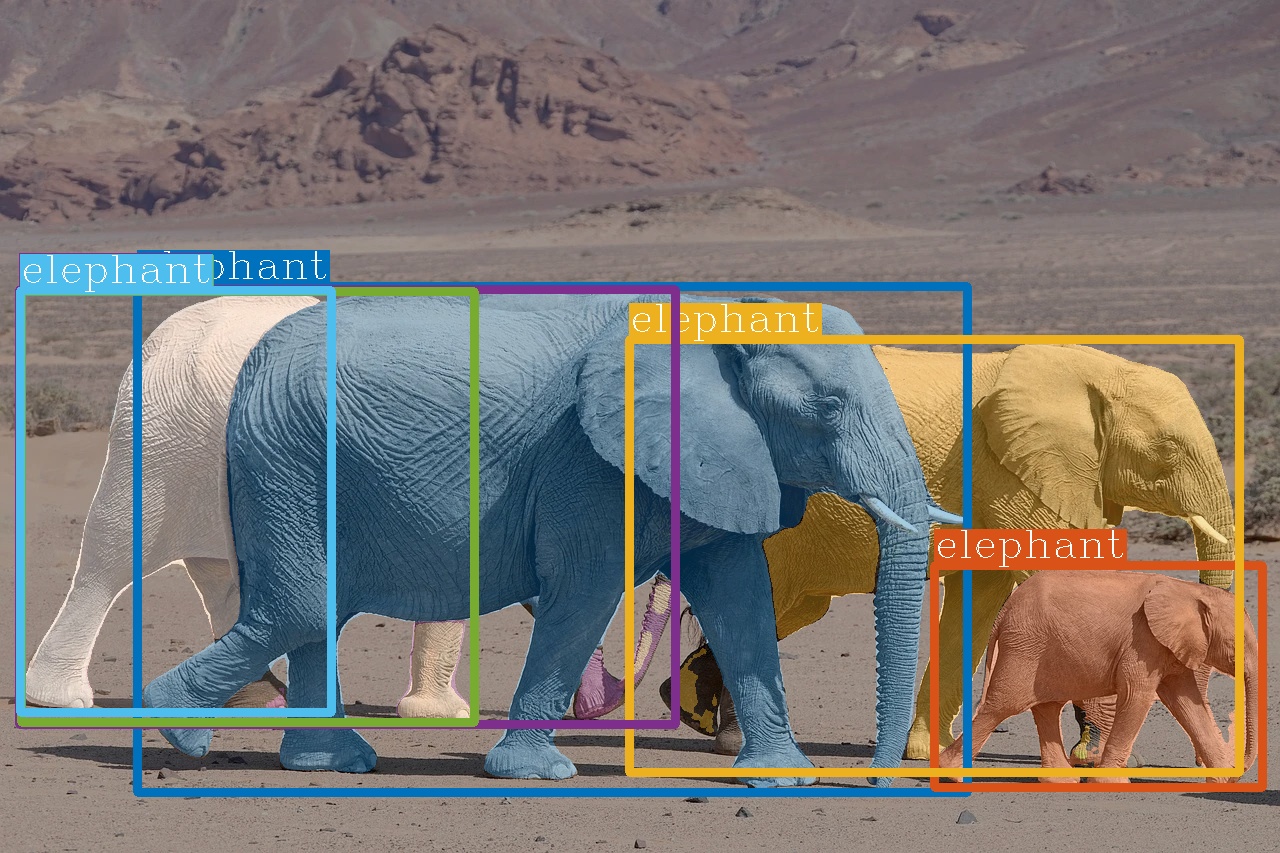} &
        \includegraphics[width=0.22\textwidth]{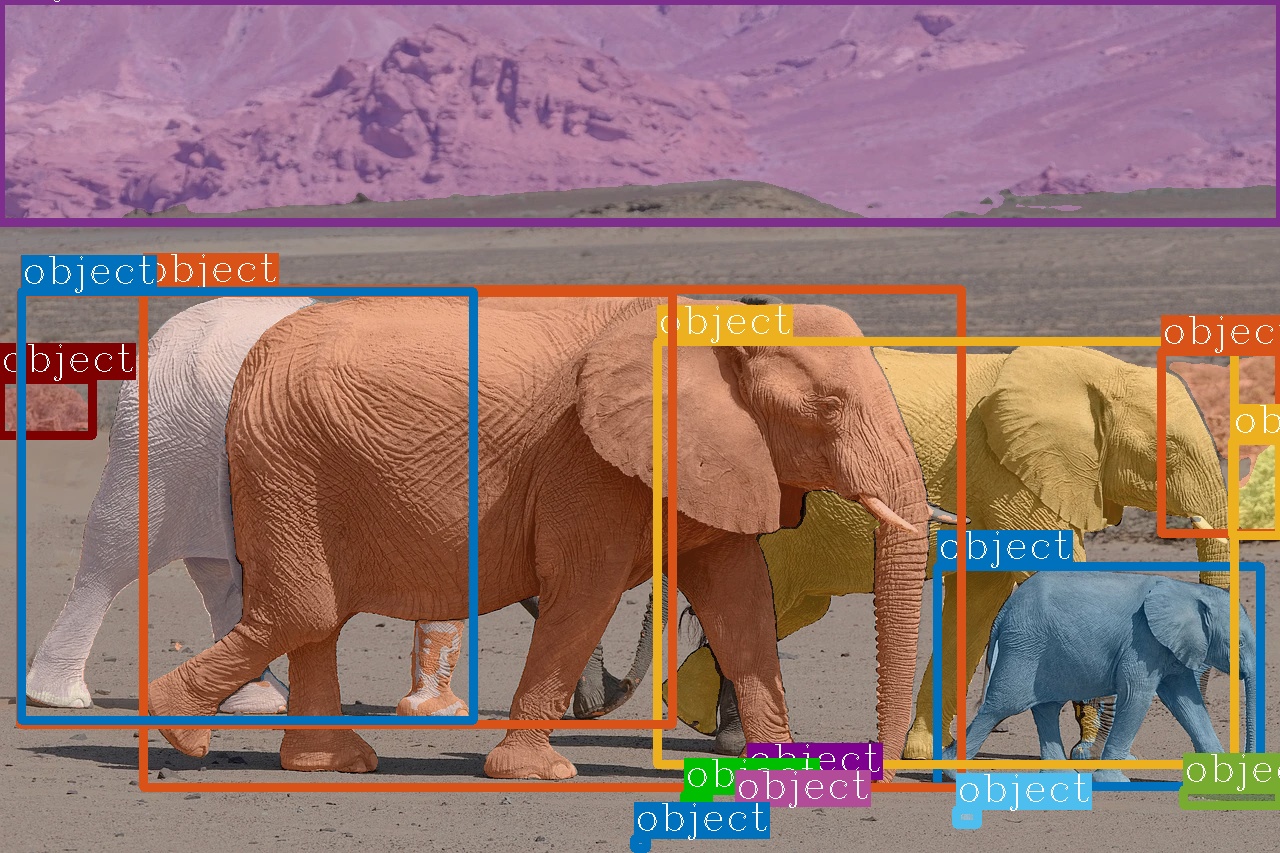} &
        \includegraphics[width=0.22\textwidth]{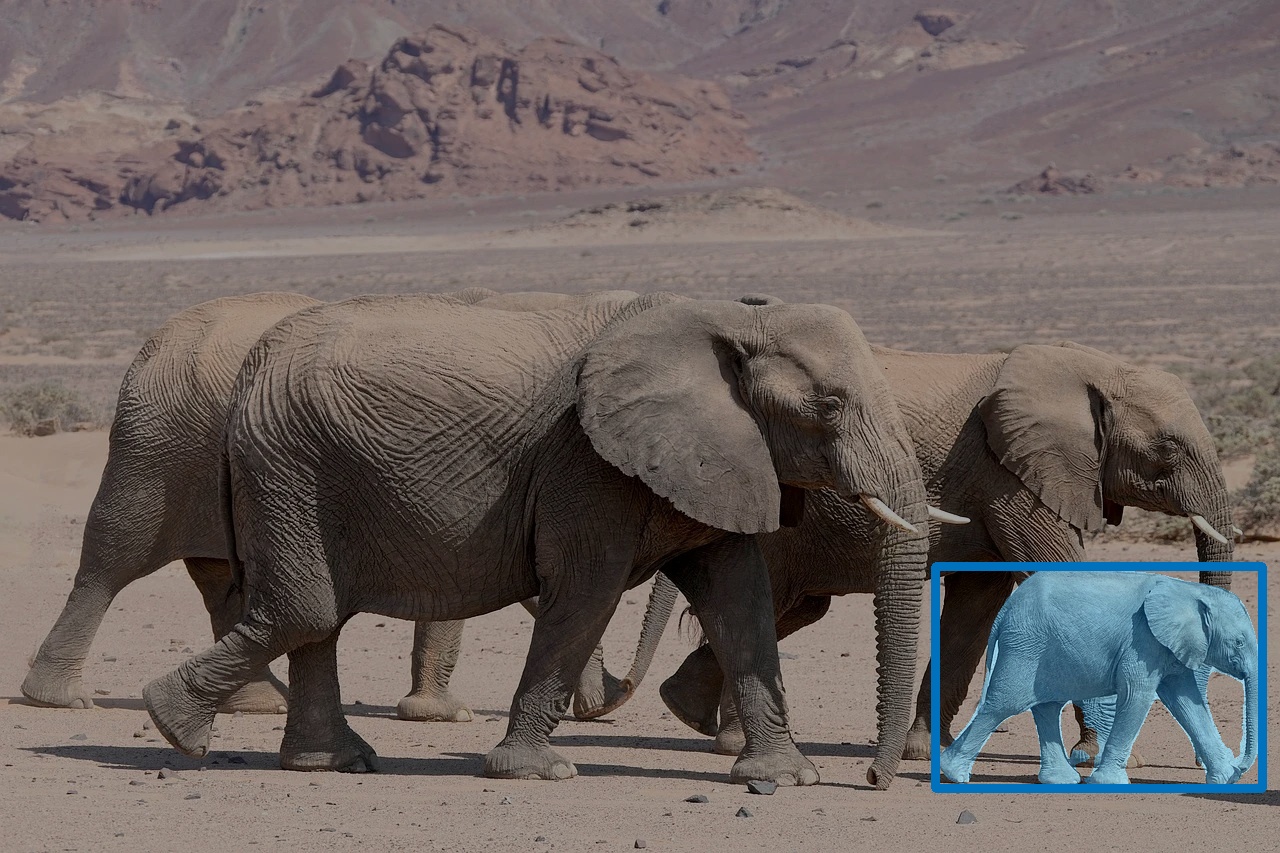} \\
        % Referring expression
        & & & \textit{"the baby elephant"} \\
        
        % Row: Elephants (no trimming)
        \includegraphics[width=0.22\textwidth]{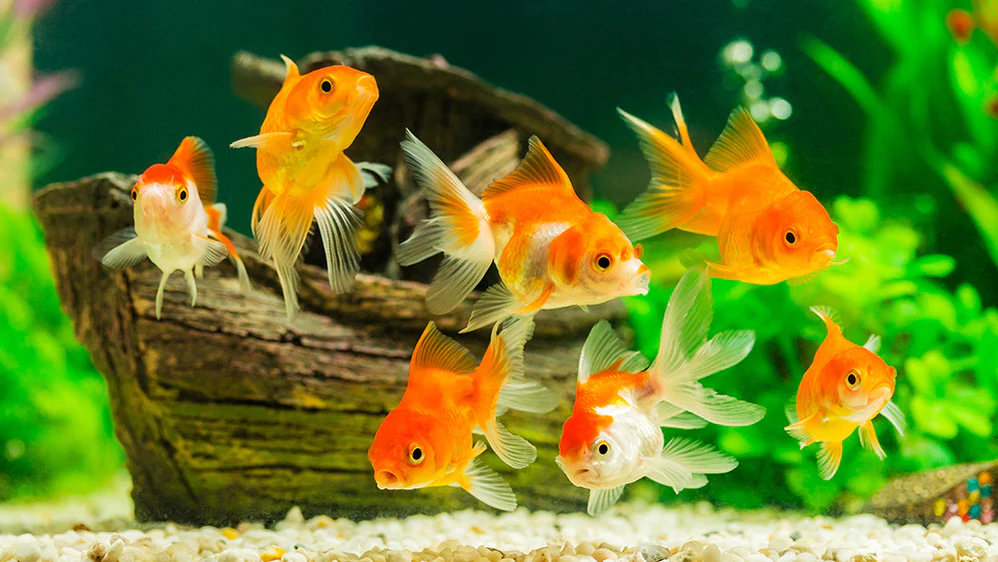} &
        \includegraphics[width=0.22\textwidth]{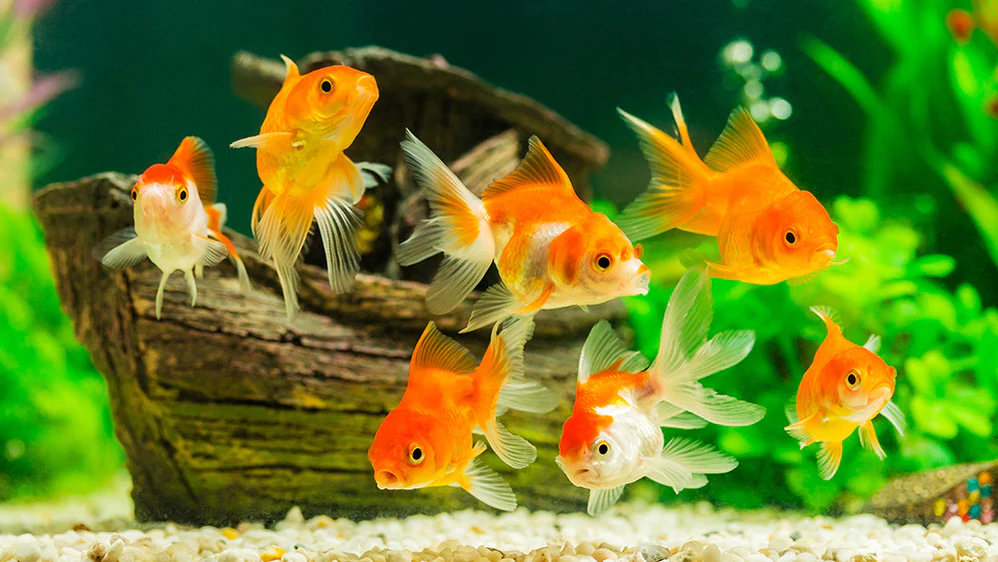} &
        \includegraphics[width=0.22\textwidth]{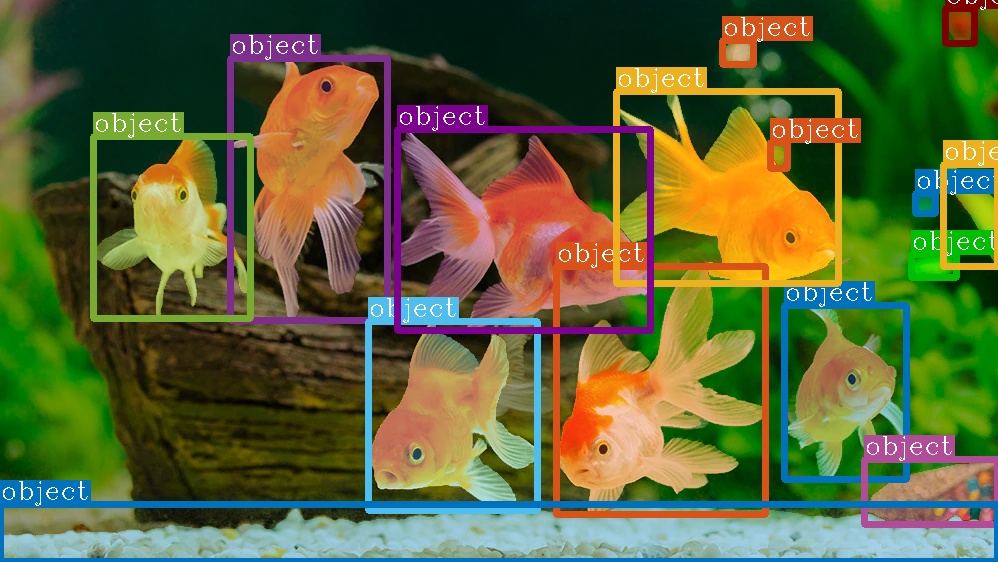} &
        \includegraphics[width=0.22\textwidth]{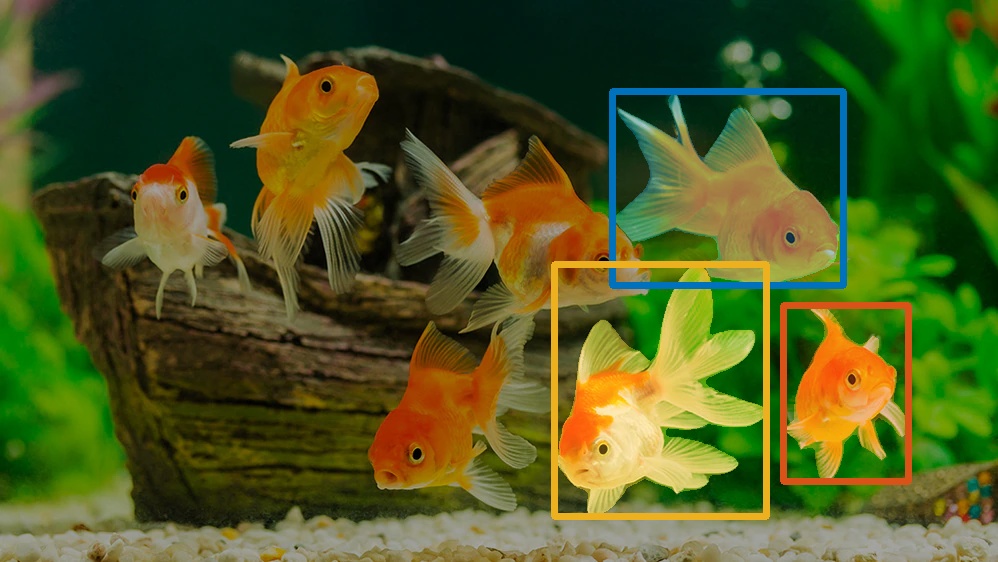} \\
        % Referring expression
        & & & \textit{"the rightmost golfishes"} \\
        
        % Row: Boats (no trimming)
        \includegraphics[width=0.22\textwidth]{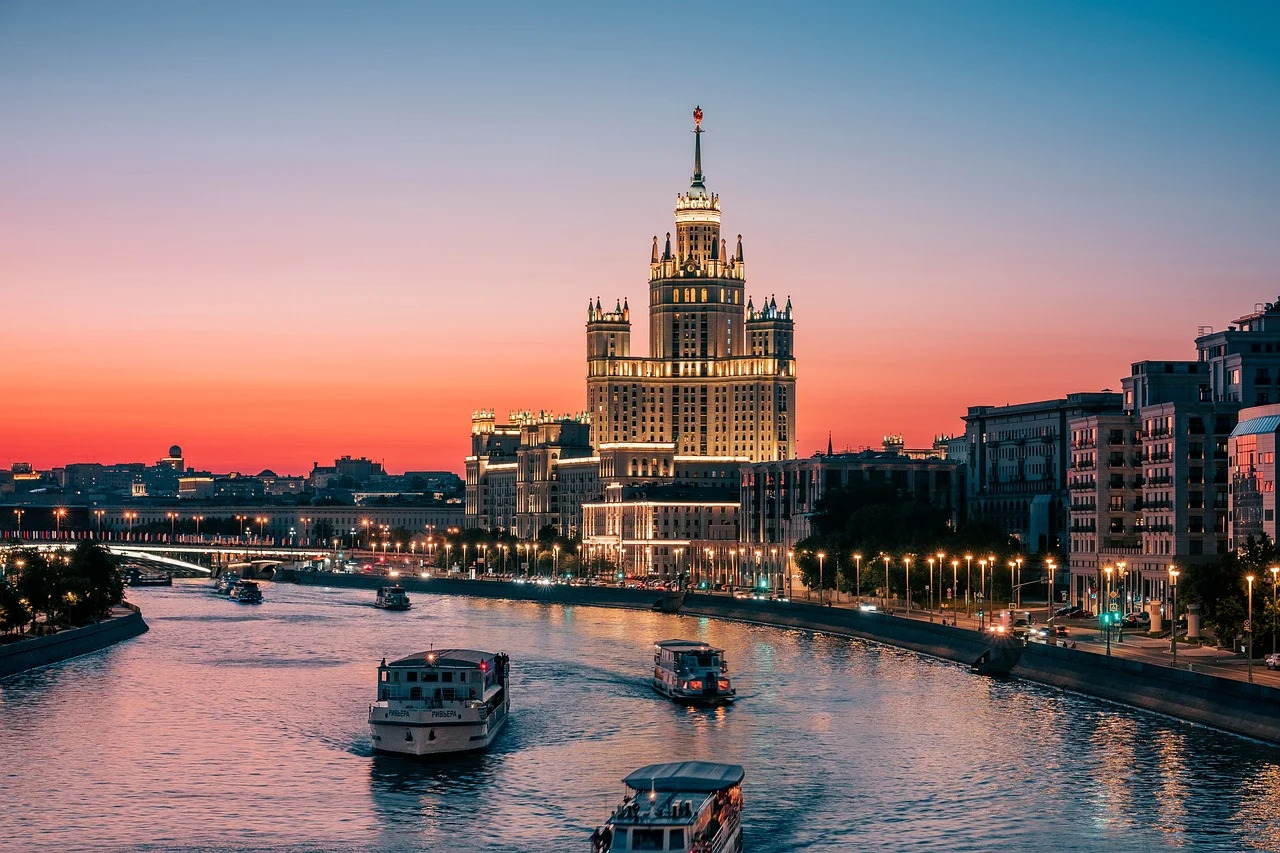} &
        \includegraphics[width=0.22\textwidth]{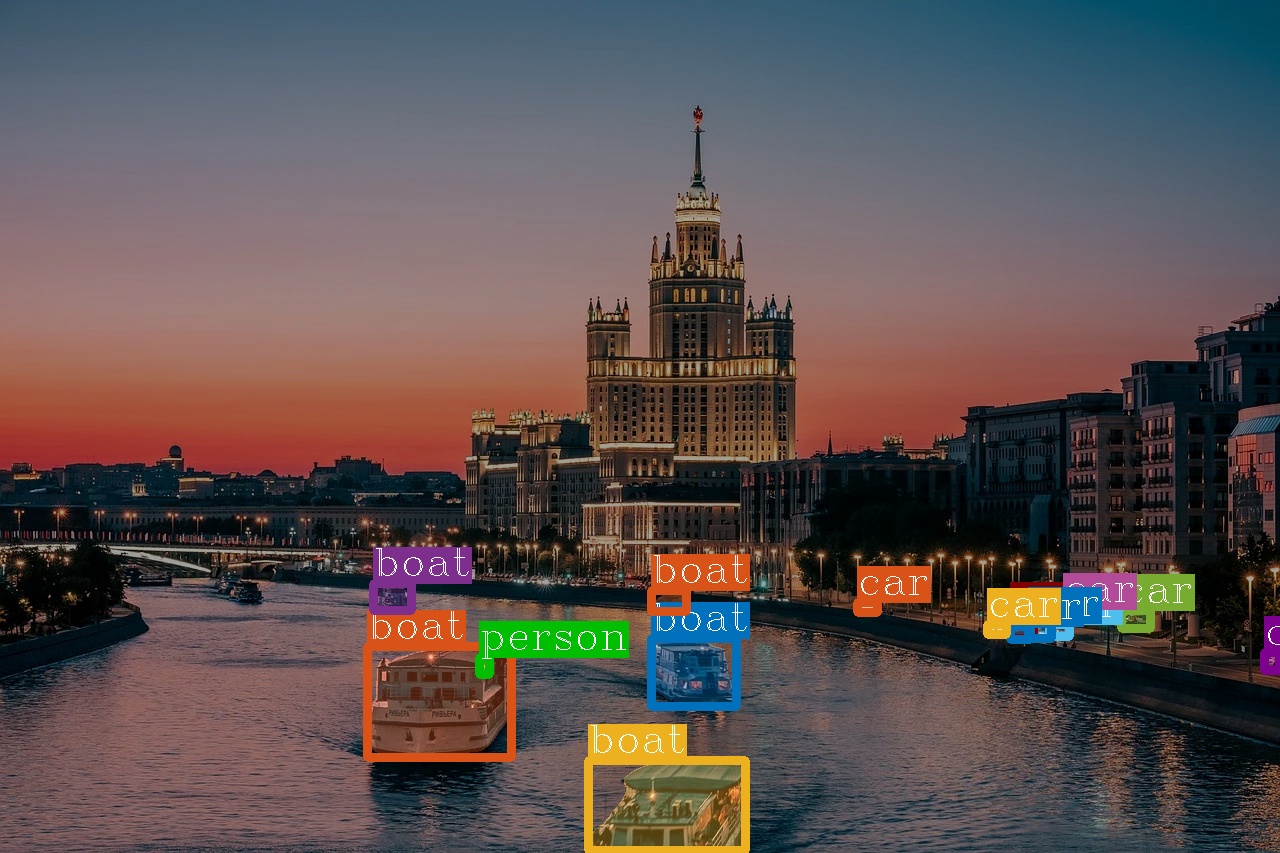} &
        \includegraphics[width=0.22\textwidth]{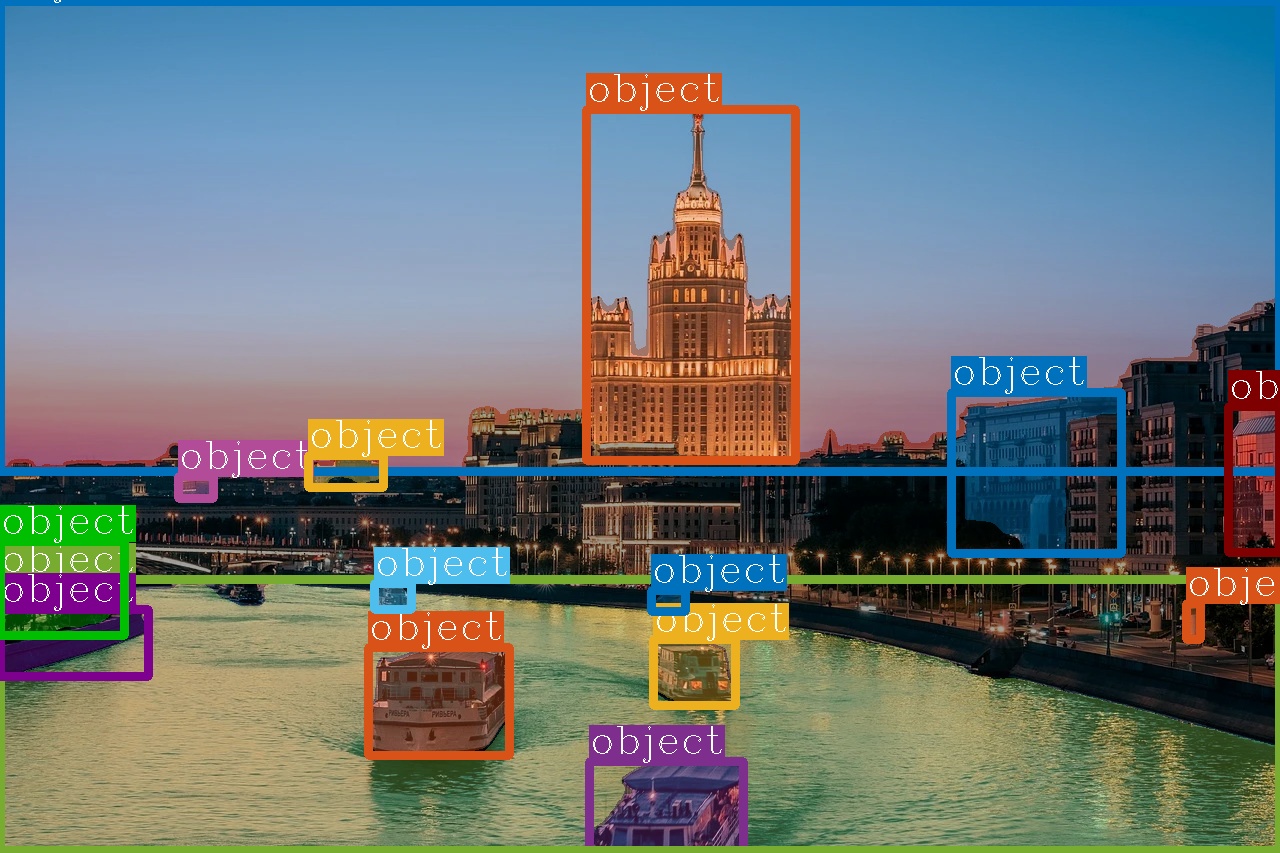} &
        \includegraphics[width=0.22\textwidth]{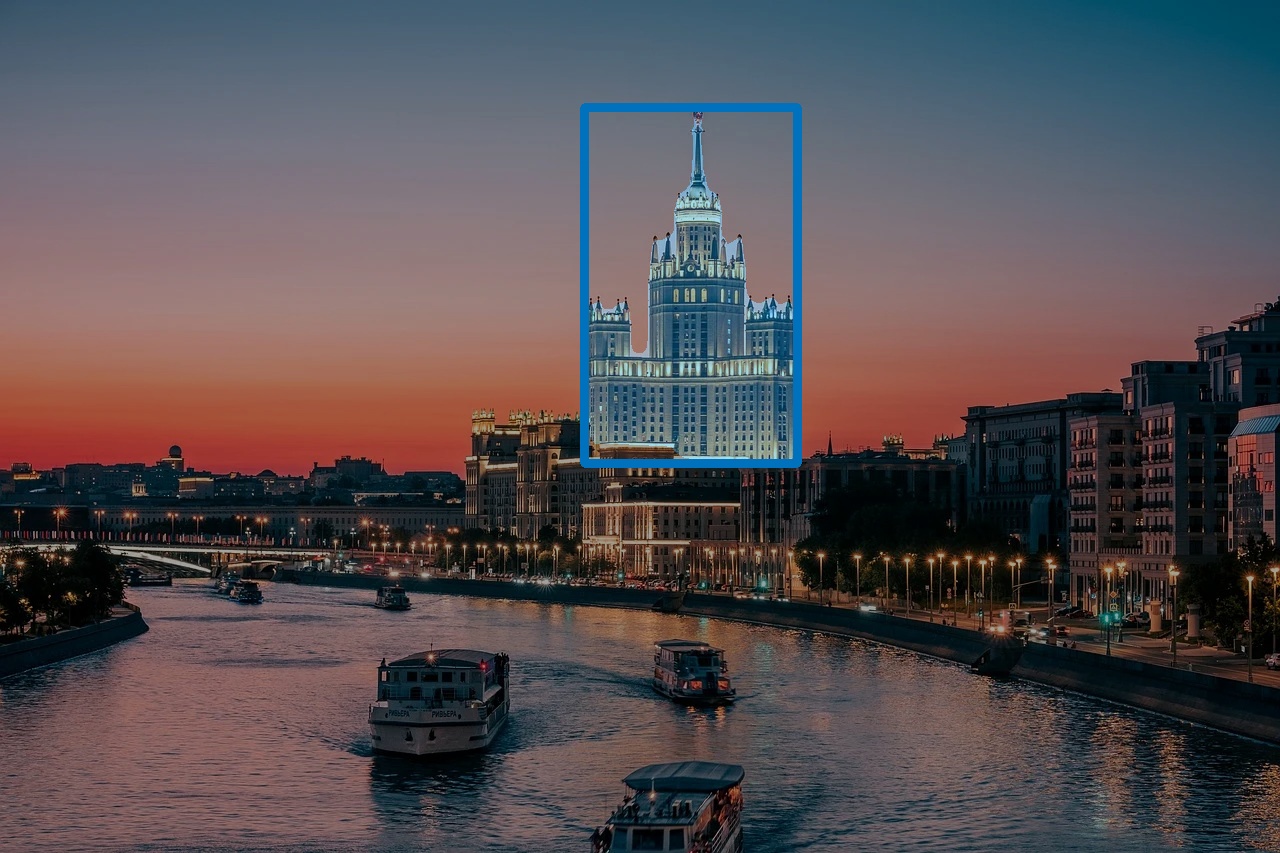} \\
        % Referring expression
        & & & \textit{"the majestic building"} \\
        
    \end{tabular}
    
    \caption{Qualitative results for different instance segmentation supported by our approach. In each row, we show the input image and report the instance segmentation results for (i) category-guided instance segmentation with COCO categories, (ii) category-agnostic instance segmentation, (iii) referring instance segmentation.}
    \label{fig:qualitative_results}
\end{figure*}